\PassOptionsToPackage{table}{xcolor}
\documentclass[11pt]{article}

\usepackage[final]{acl}

\usepackage{times}
\usepackage{latexsym}
\usepackage[T1]{fontenc}
\usepackage[utf8]{inputenc}
\usepackage{microtype}
\usepackage{inconsolata}

\usepackage{booktabs}
\usepackage{amsfonts}
\usepackage{nicefrac}

\usepackage{placeins}
\usepackage{adjustbox}
\usepackage{fancyvrb}
\usepackage{amssymb}
\usepackage{comment}
\usepackage{algorithm}
\usepackage{algorithmic}
\usepackage{newfloat}
\usepackage{listings}
\usepackage{float}
\usepackage{tabularx}
\usepackage{multirow}
\usepackage{xspace} 
\usepackage{todonotes}
\usepackage{pifont}
\usepackage{arydshln}
\usepackage{enumitem}
\usepackage{graphicx}
\usepackage{subcaption}
\usepackage{tikz}
\usetikzlibrary{calc}
\usetikzlibrary{decorations.pathreplacing}
\usetikzlibrary{fit,backgrounds}
\usetikzlibrary{arrows.meta}
\usepackage{colortbl}
\usepackage{siunitx}
\usepackage{amsmath}

\definecolor{heatneutral}{gray}{1.0} 
\definecolor{rowgray}{gray}{0.94}
\definecolor{headergray}{gray}{0.94}
\definecolor{baselinegreen}{RGB}{118,176,83} 
\definecolor{onered}{RGB} {230,61,58}  
\definecolor{neutralblue}{RGB}{68,141,192}
\definecolor{darkblue}{RGB}{30,58,138}

\floatstyle{ruled}
\newfloat{listing}{tb}{lst}{}
\floatname{listing}{Listing}

\usepackage{makecell}   

\usepackage{adjustbox}

\newcommand{\setbase}[1]{\gdef\base{#1}}
\newcommand{\setone}[1]{\gdef\one{#1}}
\newcommand{\base}{} 
\newcommand{\one}{}  

\newcommand{\tcell}[1]{%
  \pgfmathsetmacro{\db}{(#1)-(\base)}
  \pgfmathsetmacro{\do}{(#1)-(\one)}
  \ifdim \do pt < 0pt
    \pgfmathsetmacro{\ad}{abs(\do)}%
    \ifdim \ad pt > 0.50pt {\cellcolor{onered!50}#1}%
    \else\ifdim \ad pt > 0.20pt {\cellcolor{onered!40}#1}%
    \else\ifdim \ad pt > 0.10pt {\cellcolor{onered!30}#1}%
    \else\ifdim \ad pt > 0.01pt {\cellcolor{onered!20}#1}%
    \else {\cellcolor{onered}#1}%
    \fi\fi\fi\fi
  \else
    \ifdim \db pt < 0pt
      \pgfmathsetmacro{\ad}{abs(\do)}%
      \ifdim \ad pt > 1.10pt {\cellcolor{neutralblue!60}#1}%
      \else\ifdim \ad pt > 0.70pt {\cellcolor{neutralblue!40}#1}%
      \else\ifdim \ad pt > 0.50pt {\cellcolor{neutralblue!30}#1}%
      \else\ifdim \ad pt > 0.01pt {\cellcolor{neutralblue!20}#1}%
      \else {\cellcolor{heatneutral}#1}%
      \fi\fi\fi\fi
    \else
      \pgfmathsetmacro{\ad}{abs(\db)}%
      \ifdim \ad pt > 0.50pt {\cellcolor{baselinegreen!50}#1}%
      \else\ifdim \ad pt > 0.20pt {\cellcolor{baselinegreen!40}#1}%
      \else\ifdim \ad pt > 0.10pt {\cellcolor{baselinegreen!30}#1}%
      \else\ifdim \ad pt > 0.01pt {\cellcolor{baselinegreen!20}#1}%
      \else {\cellcolor{baselinegreen}#1}%
      \fi\fi\fi\fi
    \fi
  \fi
}

\newcommand{\ds}[1]{%
  \tikz[baseline=(X.base)]\node[draw,circle,inner sep=0.6pt,minimum size=1em] (X) {\scriptsize #1};%
}

\newcommand{\setlow}[1]{\gdef\low{#1}}   
\newcommand{\setup}[1]{\gdef\up{#1}}     
\newcommand{\low}{0}
\newcommand{\up}{0}

\newcommand{\oom}{\cellcolor{gray!40}{\textbf{OOM}}}

\newcommand{\tcellFO}[1]{%
  \begingroup
  \edef\val{#1}%
  \def\OOM{OOM}%
  \ifx\val\OOM
    \oom
  \else
    \pgfmathsetmacro{\dL}{(#1)-(\low)}%
    \pgfmathsetmacro{\dU}{(#1)-(\up)}%
    \ifdim \dL pt < 0pt
      \pgfmathsetmacro{\ad}{abs(\dL)}%
      \ifdim \ad pt > 2.0pt {\cellcolor{onered!85}#1}%
      \else\ifdim \ad pt > 1.0pt {\cellcolor{onered!70}#1}%
      \else\ifdim \ad pt > 0.5pt {\cellcolor{onered!55}#1}%
      \else {\cellcolor{onered!35}#1}%
      \fi\fi\fi
    \else
      \ifdim \dU pt > 0pt
        \pgfmathsetmacro{\ad}{abs(\dU)}%
        \ifdim \ad pt > 2.0pt {\cellcolor{baselinegreen!85}#1}%
        \else\ifdim \ad pt > 1.0pt {\cellcolor{baselinegreen!65}#1}%
        \else\ifdim \ad pt > 0.5pt {\cellcolor{baselinegreen!45}#1}%
        \else {\cellcolor{baselinegreen!30}#1}%
        \fi\fi\fi
      \else
        \pgfmathsetmacro{\ad}{abs(\dU)}%
        \ifdim \ad pt > 2.0pt {\cellcolor{neutralblue!55}#1}%
        \else\ifdim \ad pt > 1.0pt {\cellcolor{neutralblue!40}#1}%
        \else\ifdim \ad pt > 0.5pt {\cellcolor{neutralblue!25}#1}%
        \else {\cellcolor{neutralblue!15}#1}%
        \fi\fi\fi
      \fi
    \fi
  \fi
  \endgroup
}

\newcommand{\vsep}{\;\textbar\;}

\newcommand{\modelname}{\textit{TileMix}\xspace}

\title{\modelname: Tile-Centric Mixed-Precision Attention\\for LLM Inference Acceleration}
\author{
Hanzhi Zhang\textsuperscript{1},
Qiao Zhang\textsuperscript{2},
Qinglei Cao\textsuperscript{2},
Heng Fan\textsuperscript{1},
Yan Huang\textsuperscript{1},
Kewei Sha\textsuperscript{3},
Yunhe Feng\textsuperscript{1} \\
\textsuperscript{1}LLaVi Lab, Computer Science \& Engineering, University of North Texas \\
\textsuperscript{2}Computer Science, Saint Louis University;
\textsuperscript{3}Data Science, University of North Texas \\
\texttt{\{hanzhi.zhang,heng.fan,yan.huang,kewei.sha,yunhe.feng\}@unt.edu} \\
\texttt{\{qiao.zhang,qinglei.cao\}@slu.edu}
}

\begin{document}

\maketitle

\begin{abstract}
Long-context prefill in large language models (LLMs) incurs substantial computation and memory traffic because dense self-attention computes quadratic query-key scores.
Existing methods either use a uniform low-precision path or select token interactions, leaving spatial precision routing over hardware-aligned score tiles outside fused dense attention. 
We introduce \modelname, a tile-centric precision-routing kernel that makes numerical precision an executable spatial decision over score-tile groups within fused dense attention. \modelname partitions the attention matrix into hardware-aligned score tiles, packs routing decisions into compact bitmasks, and dispatches each tile group through FP16 or INT8 score computation while both paths update a shared online-softmax state. Scalable precision grouping lets each routing bit govern multiple adjacent key tiles, preserving hardware-aligned compute tiles and compact metadata at long contexts. By routing all legal tile groups, \modelname preserves dense token connectivity, requires no training, and supports grouped-query attention, variable-length batches, and INT8 key/value caches. Across LongEval, LV-Eval, and A100 prefill benchmarks on LLaMA, Qwen, and Vicuna, \modelname recovers long-context quality lost under uniform INT8 and improves prefill throughput over FP16, yielding a controllable accuracy-efficiency frontier across model families. The implementation is available at \url{https://github.com/HanzhiZhang-Ulrica/TileMix}.
\end{abstract}


\vspace{-15pt}
\section{Introduction}

Transformer models increasingly rely on long contexts for document summarization~\cite{cohan2018discourse}, multi-page question answering~\cite{tito2023hierarchical}, and retrieval-augmented generation~\cite{lewis2020retrieval}, making efficient long-sequence processing central to practical LLM inference.
During prefill, dense self-attention computes interactions between all query and key tokens, producing $O(L^2)$ score computation for sequence length $L$.
This quadratic computation makes attention a primary execution bottleneck for long documents and other context-intensive workloads.

\begin{figure*}[htbp]
    \centering
    \includegraphics[width=\linewidth]{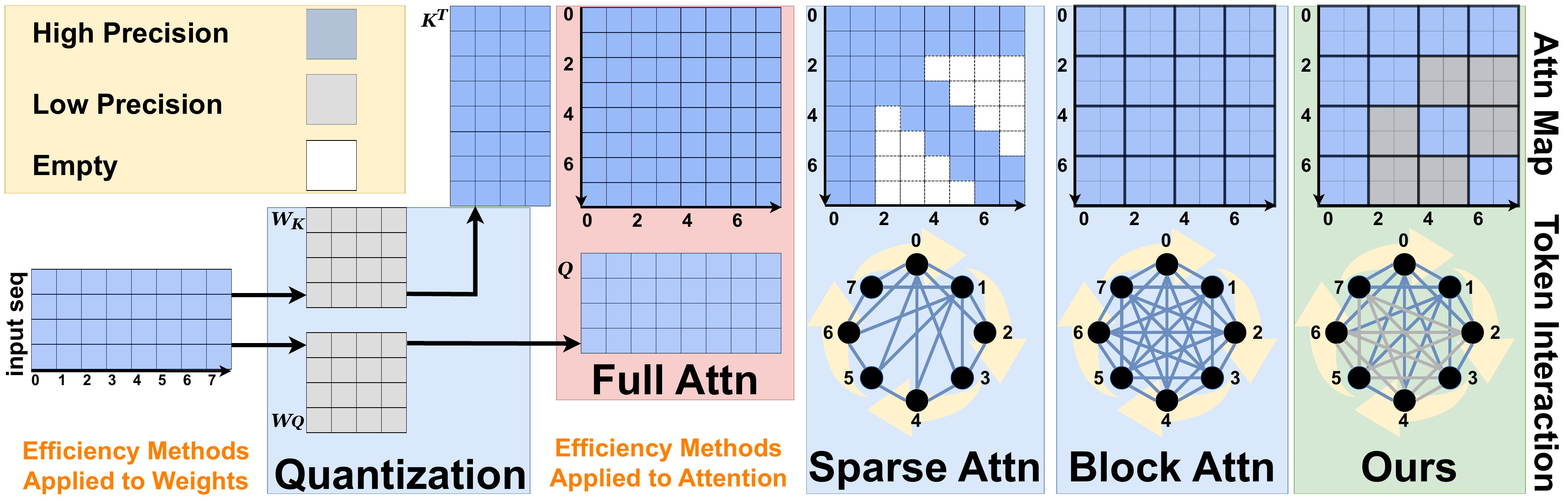}
    \vspace{-20pt}
    \caption{
    Comparison of attention efficiency strategies.
    Colors denote execution states: blue for high precision, gray for low precision, and white for removed interactions.
    \emph{Quantization} reduces weight/activation memory but often keeps attention softmax and accumulation in higher precision.
    \emph{Sparse/block attention} executes a selected subset of token interactions through structured or dynamic patterns.
    \emph{\modelname} preserves all legal token interactions and routes score-tile groups through FP16 or INT8 paths inside one fused attention kernel.}
    \label{fig:intro_compare}
    \vspace{-15pt}
\end{figure*}

Existing acceleration methods mainly optimize numerical format, token connectivity, or IO scheduling, represented in Figure~\ref{fig:intro_compare} by quantization, sparsity, and IO-aware fused attention.
(1) \textit{Low-precision quantization} such as INT8~\cite{xiao2023smoothquant, zhang2025sageattentionaccurate8bitattention} improves arithmetic and memory efficiency across model operators including weight, activation, and attention. Quantized attention kernels commonly use one arithmetic path per invocation or stage, leaving spatial precision routing over the $L\times L$ score tiles outside the streaming loop.
(2) \textit{Sparsity-based methods}~\cite{yuan2026blasst, zaheer2021bigbirdtransformerslonger, beltagy2020longformer} reduce computation by selecting active token interactions.
(3) \textit{IO-aware fused attention}~\cite{dao2022flashattentionfastmemoryefficientexact} partitions attention into hardware-aligned tiles and fuses score computation, online softmax, and value aggregation.
These kernels use tiles for data movement and work partitioning, while retaining a uniform score-computation path.
Together, these directions suggest using hardware-aligned score tiles as the spatial unit of precision.
A fused kernel can preserve the complete attention graph while routing score-tile groups through multiple arithmetic paths in one streaming computation.

Modern fused-attention kernels achieve high utilization through regular Tensor Core tiling and coordinated work partitioning, online softmax, and data movement across the GPU memory hierarchy~\cite{dao2023flashattention, shah2024flashattention}.
Tile-group precision routing must reconcile distinct FP16 and INT8 Tensor Core paths, including INT8 rescaling, before both paths update the shared row-wise maximum, normalizer, and output accumulator~\cite{choquette2021nvidia, chen2024int}.
Precision dispatch therefore falls inside the latency-critical inner loop, making compact tile-aligned routing essential for regular long-context execution and creating a kernel-design problem beyond invocation-level precision selection.

To address these challenges, we introduce \modelname, a tile-centric precision-routing kernel that integrates heterogeneous score arithmetic into a single FlashAttention-style execution.
For each query-tile row, \modelname loads a packed routing word, decodes each key-tile-group decision with constant-time bit operations, and dispatches QK computation to FP16 or INT8 Tensor Core paths. After rescaling, both paths update the shared online-softmax state, preserving dense streaming execution.
Scalable precision grouping lets each routing bit govern adjacent key tiles while retaining the underlying hardware-aligned compute tiles and compact metadata as context length grows.
The resulting kernel combines the dense connectivity of full attention with the arithmetic flexibility of mixed precision under training-free deployment. It also supports grouped-query attention, variable-length batching, and INT8 key/value caches, and exposes a controllable accuracy-efficiency frontier between FP16 and uniform INT8 attention.

Our contributions are as follows:
\begin{enumerate}[nosep,left=0pt]
    \item \textbf{Tile-Centric Precision Routing for Dense Attention:}
    We introduce tile-group precision as a spatial execution abstraction for fused dense attention, enabling fine-grained FP16/INT8 allocation across all legal token interactions.
    \item \textbf{Shared-State Heterogeneous Score Execution:}
    We design a fused kernel that aligns FP16 and INT8 score paths to a common score domain and integrates them through one online-softmax recurrence.
    \item \textbf{Compact and Scalable Kernel-Native Routing:}
    We develop packed bitmask routing with constant-time inner-loop lookup and $\mathcal{O}(H_kT_m)$ metadata, together with precision grouping that scales routing to long contexts while retaining hardware-aligned compute tiles.
    \item \textbf{Practical Long-Context Inference and Evaluation:}
    We implement grouped-query attention, variable-length batching, and INT8 key/value cache support, and validate \modelname through long-context retrieval, question answering, prefill efficiency, and numerical analyses across LLaMA, Qwen, and Vicuna models.
\end{enumerate}


\vspace{-5pt}
\section{Related Work}
\vspace{-5pt}

Transformer acceleration primarily follows three directions: low-precision quantization, IO-aware tiled attention, and structural sparsity. These approaches reduce long-context inference cost through numerical compression, data-movement optimization, or selective token connectivity.

\textit{Low-precision quantization} reduces memory and arithmetic costs by representing weights and activations in formats such as INT8~\cite{van2023fp8,srinivasa2025evaluating} and INT4~\cite{zhao2024atom}. Quantization-aware and post-training methods improve robustness through calibration, activation transformation, outlier handling, and blockwise scaling~\cite{yao2022efficient,xiao2023smoothquant,saxena2024resq}. Recent quantized attention kernels integrate low-precision score computation, value aggregation, and numerical approximation into fused execution~\cite{chen2024int,kang2024turboattentionefficientattentionapproximation,zhang2025sageattentionaccurate8bitattention,sharratt2026thriftattention}. These designs typically assign formats at the tensor, operator, attention-stage, or quantization-block level, while two-dimensional score-tile groups follow one arithmetic path within the streaming loop~\cite{dettmers2022gpt3,van2023fp8,kluska2024qattn}.

\textit{IO-aware attention kernels}~\cite{dao2022flashattentionfastmemoryefficientexact,dao2023flashattention,dege2025flashmla} process attention in SRAM-resident tiles and fuse score computation, online softmax, and value aggregation, avoiding full attention-matrix materialization in HBM. These tiles govern data movement and work partitioning, while arithmetic precision is commonly fixed per kernel invocation or attention stage~\cite{shah2024flashattention,chen2024int}.

\textit{Sparse attention methods} reduce computation by selecting token interactions through sliding-window, strided, dilated, or mixed local-global patterns~\cite{beltagy2020longformer,child2019generatinglongsequencessparse,zaheer2021bigbirdtransformerslonger,ainslie2020etcencodinglongstructured}. Other systems use content-based selection and clustering~\cite{roy2020efficientcontentbasedsparseattention,wang2021clusterformerclusteringbasedsparsetransformer,kitaev2020reformerefficienttransformer}, positional mechanisms~\cite{dai2019transformerxlattentivelanguagemodels,zhang-etal-2023-adaptive}, adaptive architectures~\cite{correia2019adaptivelysparsetransformers}, or dynamically constructed active block sets~\cite{jiang2024minference,lai2025flexprefill,yuan2026blasst}. These methods use spatial structure to select executed token interactions, changing attention connectivity while keeping precision outside the selection decision.


\vspace{-5pt}
\section{Preliminaries}\label{sec:preliminary}
\vspace{-5pt}

We briefly review (i) FlashAttention-style tiled attention with online softmax, and (ii) blockwise quantization for low-precision matrix multiplication.
Their interaction defines the central kernel challenge addressed by \modelname: FP16 and INT8 score tiles follow different arithmetic paths but contribute to one shared streaming softmax state. 
Appendix Table~\ref{tab:notation} summarizes the notation used throughout the paper.

\vspace{-5pt}
\subsection{Tiled Attention with Online Softmax}
\vspace{-5pt}

Consider one self-attention head with $Q,K,V\in\mathbb{R}^{L\times d}$, where $L$ is the sequence length and $d$ is the head dimension.
Attention computes the score matrix $S\in\mathbb{R}^{L\times L}$, attention weights $P\in\mathbb{R}^{L\times L}$, and output $O\in\mathbb{R}^{L\times d}$ as
\(
S = \frac{QK^\top}{\sqrt{d}},
P=\mathrm{softmax}(S),
O=PV.
\)
Materializing $S$ and $P$ in GPU high-bandwidth memory (HBM) requires an $\mathcal{O}(L^2)$ intermediate-memory footprint and substantial \emph{HBM read/write traffic} at long context lengths.

FlashAttention streams key/value tiles while keeping score and probability tiles on chip and writing only normalized outputs to HBM, providing the execution substrate for \modelname's tile-group precision routing. The kernel partitions $Q$ into $T_m=\left\lceil L/b_q\right\rceil$ tiles $\{Q_m\}_{m=1}^{T_m}$ and $(K,V)$ into $T_n=\left\lceil L/b_{kv}\right\rceil$ tiles $\{(K_n,V_n)\}_{n=1}^{T_n}$.
For each $Q_m$, it streams over $n=1,\dots,T_n$ while maintaining the shared online-softmax state $\big(\tilde m_m^{\,n},\tilde z_m^{\,n},\tilde O_m^{\,n}\big)$, comprising the row-wise maximum and normalizer $\tilde m_m^{\,n},\tilde z_m^{\,n}\in\mathbb{R}^{b_q}$ and unnormalized accumulator $\tilde O_m^{\,n}\in\mathbb{R}^{b_q\times d}$. Initialized by $\tilde m_m^{\,0}=-\infty$, $\tilde z_m^{\,0}=0$, and $\tilde O_m^{\,0}=0$, the state updates as
\[
\begin{aligned}
\tilde m_m^{\,n}
&=\max\!\Big\{\tilde m_m^{\,n-1},\mathrm{rowmax}(S_m^{\,n})\Big\},\\
\tilde z_m^{\,n}
&=e^{\tilde m_m^{\,n-1}-\tilde m_m^{\,n}}\tilde z_m^{\,n-1}
+\mathrm{rowsum}\!\left(e^{S_m^{\,n}-\tilde m_m^{\,n}}\right),\\
\tilde O_m^{\,n}
&=e^{\tilde m_m^{\,n-1}-\tilde m_m^{\,n}}\tilde O_m^{\,n-1}
+e^{S_m^{\,n}-\tilde m_m^{\,n}}V_n,\\
S_m^{\,n}
&=\frac{Q_mK_n^\top}{\sqrt{d}}.
\end{aligned}
\]
Here, $\mathrm{rowmax}(\cdot)$ and $\mathrm{rowsum}(\cdot)$ reduce across the $b_{kv}$ columns of $S_m^{\,n}\in\mathbb{R}^{b_q\times b_{kv}}$.
After tile $T_n$, row-wise division of $\tilde O_m^{\,T_n}$ by $\tilde z_m^{\,T_n}$ yields the output for $Q_m$.

\vspace{-5pt}
\subsection{Blockwise Quantization}
\vspace{-5pt}

For a matrix product $C=AB$, blockwise quantization applies $\psi(\cdot)$ independently to operand blocks, producing low-precision representations and their scale factors:
\[
(\hat{A},\delta_A)=\psi(A),\;
(\hat{B},\delta_B)=\psi(B),\;
C\approx\delta_A\delta_B(\hat{A}\hat{B}).
\]
The representations $\hat{A}$ and $\hat{B}$ may use INT8, FP8, or other low-precision formats.
We instantiate $\psi(\cdot)$ with \textbf{INT8} because NVIDIA A100 GPUs provide optimized \textbf{INT8 Tensor Core} primitives for high-throughput matrix multiplication. These Tensor Cores execute \textbf{MMA} (\emph{matrix multiply-accumulate}) instructions that multiply INT8 fragments and accumulate partial sums in INT32 registers. Aligning the quantization blocks with attention tiles allows the corresponding scales $\delta_A$ and $\delta_B$ to be applied efficiently during fused execution.

FP16 and INT8 score tiles exhibit different rounding, accumulation, and rescaling behavior. Because every score tile contributes to the shared state $(\tilde m_{m}^{\,n},\tilde z_{m}^{\,n},\tilde O_m^{\,n})$, path-specific numerical effects propagate through running-maximum tracking, normalization, and output accumulation. Mixed-precision attention is therefore a shared-state kernel problem: heterogeneous score paths must enter a common score domain before updating the same running maximum, normalizer, and output accumulator.

\begin{figure*}[t]
    \centering
    \includegraphics[width=\textwidth]{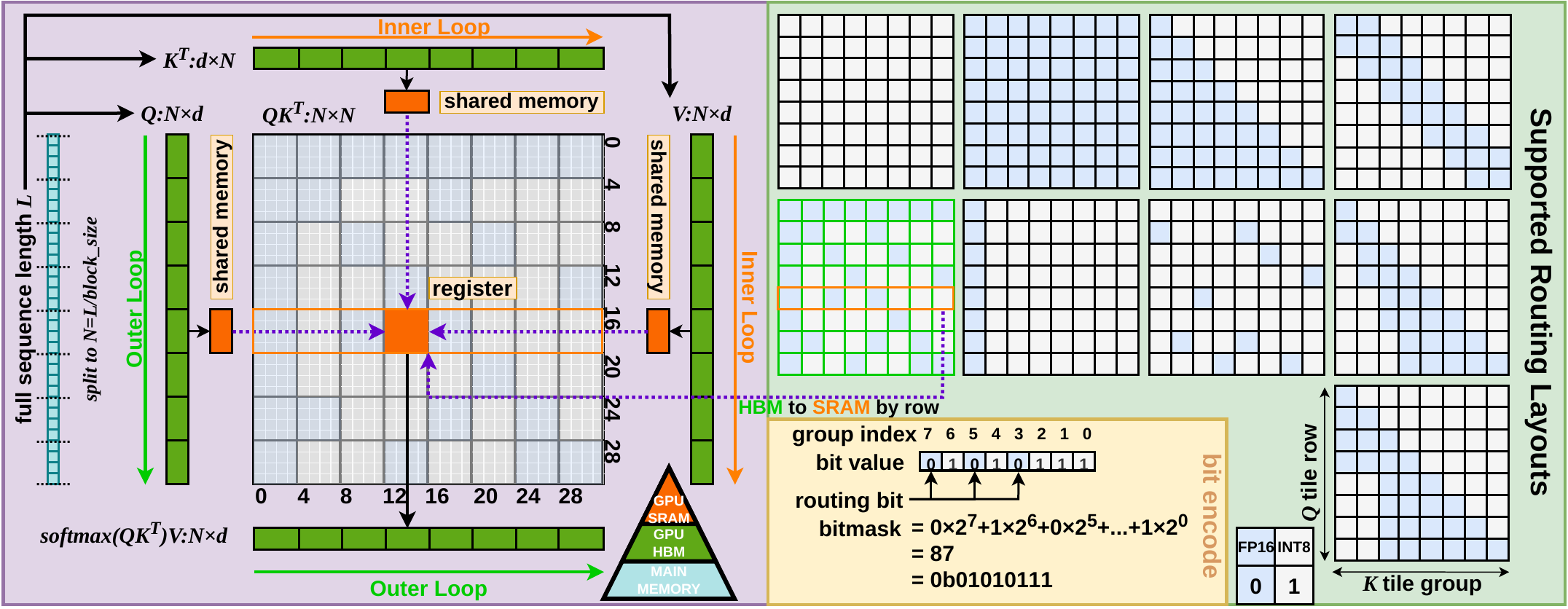}
    \vspace{-20pt}
    \caption{\modelname fused attention with tile-group precision routing. 
    \emph{Left:} The $L_q\times L_k$ attention matrix is processed in $\mathrm{BLOCK}_M\times \mathrm{BLOCK}_N$ compute tiles. For each query-tile row $m$ (outer loop), the kernel streams key/value tiles $n$ (inner loop) from HBM to on-chip SRAM and registers, dispatches the score tile $S_m^{\,n}=Q_mK_n^\top/\sqrt{d}$ to either FP16 matmul or INT8 Tensor Core MMA with INT32 accumulation and rescaling, and updates a shared FP16 online-softmax state to produce $O_m$ without materializing the full score matrix.
    \emph{Right:} A binary routing map is grouped along the key-tile dimension with width $\mathrm{BLOCK}_N^{\text{mask}}=g\,\mathrm{BLOCK}_N$ and packed per $(h_k,m)$ into a 64-bit mask $b_{h_k,m}$. The bit at position $g_j=\lfloor n/g\rfloor$ selects the arithmetic path for the corresponding key-tile group (1: INT8, 0: FP16). Kernel-side legality masks, including causality and boundary conditions, are enforced independently of the routing map.}
    \label{fig:framework}
    \vspace{-15pt}
\end{figure*}

\section{\modelname: Tile-Group Precision Routing}\label{sec:method}
\vspace{-5pt}

\modelname realizes tile-group precision routing \emph{without retraining} by combining:
(i) a precision policy that assigns each legal score-tile group to \textbf{FP16} or \textbf{INT8} score computation,
(ii) packed bitmasks that convert the policy into constant-time inner-loop dispatch, and
(iii) a FlashAttention-style fused kernel in which both score paths update one shared online-softmax state.
For each KV head and query-tile row, the inner loop reads one packed routing word and extracts the arithmetic-path decision for each key-tile group.
The fused kernel natively supports grouped-query attention (GQA) and variable-length batching through prefix-sum metadata, while the implementation provides an INT8 key/value cache interface for decode.

\vspace{-3pt}
\subsection{\modelname Overview}
\vspace{-3pt}

\modelname partitions the $L_q\times L_k$ attention into hardware-aligned two-dimensional score tiles of size $\mathrm{BLOCK}_M \times \mathrm{BLOCK}_N$ (Figure~\ref{fig:framework}), with $\mathrm{BLOCK}_M \equiv b_q$ and $\mathrm{BLOCK}_N \equiv b_{kv}$. 
The compute tile remains the execution unit, while one routing group may govern multiple adjacent compute tiles along the key dimension.
The fused kernel follows the two-level structure of IO-aware attention:

\textbf{Outer loop (over query tiles).}
The kernel iterates over query tiles $\{Q_m\}_{m=1}^{T_m}$, where $T_m=\lceil L_q/\mathrm{BLOCK}_M\rceil$.
Each Triton program instance owns one query-tile row $m$.

\textbf{Inner loop (streaming key/value tiles).}
For each $Q_m$, the kernel streams key/value tiles $\{(K_n,V_n)\}_{n=1}^{T_n}$ from HBM into SRAM and registers, where $T_n=\lceil L_k/\mathrm{BLOCK}_N\rceil$.
Across this inner loop, the kernel updates one online-softmax state and retains the output accumulator on chip until normalization after the final tile.

For each compute tile $(m,n)$, the routing group containing key-tile index $n$ selects the \textbf{FP16} or \textbf{INT8} score-computation path.
Kernel-side causal and boundary masks determine legal interactions, while the routing policy determines the arithmetic path of each legal tile group.

\modelname supports grouped-query attention (GQA) and variable-length batching through one routing interface. Let $H_q$ and $H_k$ denote the numbers of query and KV heads. Each query head $h_q\in\{0,\dots,H_q-1\}$ maps to a KV head
\(
h_k = \left\lfloor \frac{h_q}{H_q/H_k} \right\rfloor,
\)
which indexes routing lookup. Query heads mapped to the same KV head share tile-group routing decisions. For variable-length batching, flattened inputs use $\mathrm{cu\_seqlens}$ and prefix-sum metadata for padding-free routing and execution.

\vspace{-5pt}
\subsection{Routing Policy and Bitmask Encoding}
\vspace{-5pt}

Fused attention kernels derive tile execution from pointers, strides, legality masks, and a shared arithmetic configuration. \modelname introduces binary routing metadata that selects the score-computation path inside the inner loop.

\textbf{Policy definition.}
\modelname accepts a binary tile-group routing map $R$ indexed by KV head $h_k$, query-tile row $m$, and key-tile group $g_j$.
In our evaluation, \emph{static, data-free} structured templates instantiate $R$, distributing FP16-routed groups spatially under configurable INT8 coverage budgets. The decision $R_{h_k,m,g_j}=1$ dispatches the corresponding group to \textbf{INT8}, while $R_{h_k,m,g_j}=0$ dispatches it to \textbf{FP16}.
(See Appendix~\ref{app:precision-layouts} for the full list of precision layouts and their definitions.)

\begin{figure}[ht]
    \centering
    \vspace{-5pt}
    \includegraphics[width=1\linewidth]{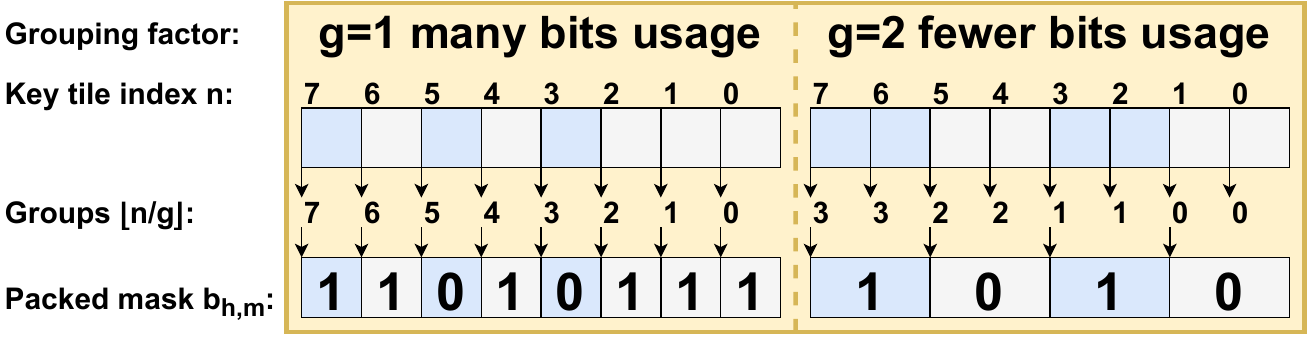}
    \caption{Key-tile grouping and bitmask encoding for one query-tile row $m$. Adjacent compute tiles along the key dimension form routing groups $g_j$, each controlled by one bit in the packed mask $b_{h_k,m}$ (1: INT8, 0: FP16).}
    \label{fig:mask_grouping}
    \vspace{-5pt}
\end{figure}

\textbf{Why grouping is needed.}
\modelname targets constant-time routing lookup inside the attention inner loop. As Figure\ref{fig:mask_grouping} shows, we pack up to 64 key-tile-group decisions for each query-tile row into a single 64-bit word. To extend the same routing word across longer key sequences while retaining hardware-aligned compute tiles, \modelname defines the routing-group width $\mathrm{BLOCK}_N^{\text{mask}}$ through a grouping factor $g\in\mathbb{Z}^+$:

{\small
\[
\mathrm{BLOCK}_{N}^{\text{mask}}=g\cdot\mathrm{BLOCK}_N,\;
T_{\text{mask}}=\left\lceil\frac{L_k}{\mathrm{BLOCK}_{N}^{\text{mask}}}\right\rceil\le 64.
\]
}

When $g>1$, \textbf{one routing bit controls a contiguous group of $g$ adjacent key tiles} (Figure~\ref{fig:mask_grouping}). Each decision therefore spans $g\cdot \mathrm{BLOCK}_N$ key tokens.
This grouping extends one routing word across long key sequences while preserving the hardware-aligned compute-tile structure.

\begin{figure}[ht]
    \centering
    \vspace{-5pt}
    \includegraphics[width=1\linewidth]{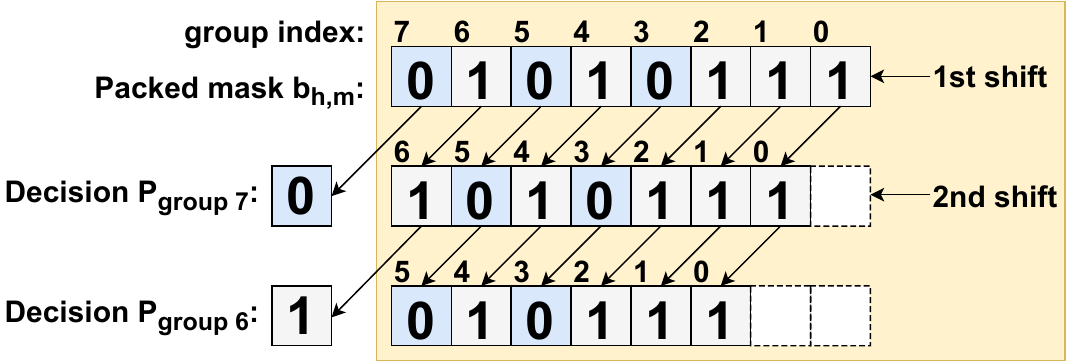}
    \caption{Constant-time shift-and-mask routing lookup. Shifting bit $g_j$ to the least-significant position yields the arithmetic-path decision $R_{h_k,m,g_j}$ for the corresponding key-tile group.}
    \label{fig:bit_extraction}
    \vspace{-5pt}
\end{figure}

\textbf{Bitmask packing.}
After grouping, the routing map $R$ is indexed by KV head $h_k$, query-tile row $m$, and key-tile group $g_j\in\{0,\dots,T_{\text{mask}}-1\}$. For each $(h_k,m)$, we pack the group decisions into a 64-bit integer, or packed \emph{bitmask word},
{\small
\[
b_{h_k,m}=\sum_{g_j=0}^{T_{\text{mask}}-1}R_{h_k,m,g_j}\,2^{g_j},\; R_{h_k,m,g_j}\in\{0,1\}.
\]
}

\textbf{Constant-time lookup.}
Inside the kernel, one shift-and-mask operation retrieves the routing decision for group $g_j$ (Figure~\ref{fig:bit_extraction}):
\[
R_{h_k,m,g_j}
= \left( (b_{h_k,m} \gg g_j) \;\&\; 1 \right).
\]
The general representation requires one 64-bit word per $(h_k,m)$, totaling $\mathcal{O}(H_kT_m)$ routing metadata. Packed bitmasks therefore provide compact kernel-native routing metadata consumed directly by the streaming inner loop.

\subsection{Tile-Group Mixed-Precision Attention}

\modelname preserves the FlashAttention-style online-softmax recurrence defined in Section~\ref{sec:preliminary}; routing changes only the arithmetic path used to construct each score tile $S_m^{\,n}$.

The FP16 and INT8 paths contribute distinct rounding, accumulation, and rescaling behavior. The \textbf{FP16} and \textbf{INT8} paths use different multiplication, accumulation, and scale-restoration procedures. After INT8 scale restoration, score tiles from both paths enter a common floating-point domain before updating the shared running maximum, normalizer, and output accumulator.
\modelname maintains this shared online-softmax state in FP16, while the routing map $R$ assigns \textbf{FP16} or \textbf{INT8} score computation to each legal tile group. Both arithmetic paths apply the same $1/\sqrt{d}$ scaling and exponentiation implementation before contributing to the shared normalization.

\textbf{Tile-group dispatch.}
For query-tile row $m$ and key-tile index $n$, the routing-group index is
$g_j=\lfloor n/g\rfloor$, with $\mathrm{BLOCK}_{N}^{\text{mask}}=g\,\mathrm{BLOCK}_N$.
The corresponding routing bit defines the routed score tile $S_m^{\,n}$:
{\small
\[
S_m^{\,n} =
\begin{cases}
\operatorname{Rescale}_{m,n}\!\left(Q_{8,m}K_{8,n}^{\top}\right)/\sqrt{d},
& R_{h_k,m,g_j}=1,\\[2pt]
Q_mK_n^{\top}/\sqrt{d},
& R_{h_k,m,g_j}=0.
\end{cases}
\]
}
\vspace{-10pt}

\textbf{INT8 path and scale alignment.}
For the primary evaluated score-routing path, \modelname quantizes $Q$ and $K$ blockwise to produce $Q_8$ and $K_8$ together with per-block scales, while $V$ and the $PV$ computation remain in FP16.
The quantize-once mode prepares $Q_8$, $K_8$, and their scales once per attention call and reuses them across all INT8-routed tile groups.

\begin{figure*}[thb]
    \centering
    \vspace{-5pt}
    \begin{subfigure}[b]{0.485\linewidth}
        \centering
        \includegraphics[width=\linewidth]{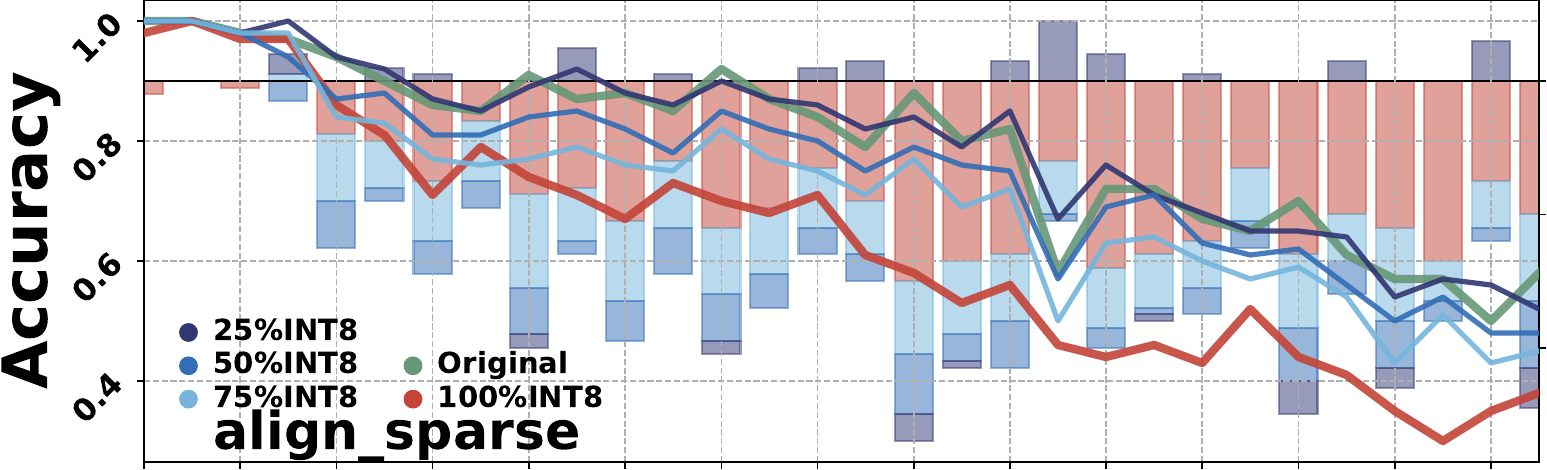}
    \end{subfigure}
    \hfill
    \begin{subfigure}[b]{0.487\linewidth}
        \centering
        \includegraphics[width=\linewidth]{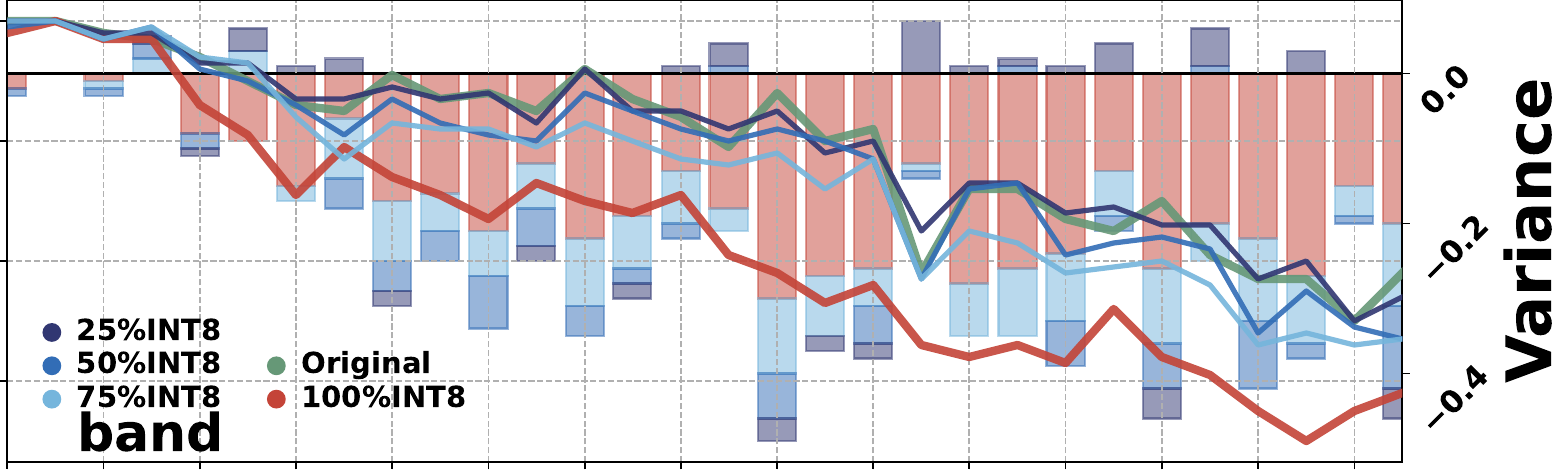}
    \end{subfigure}

    \begin{subfigure}[b]{0.485\linewidth}
        \centering
        \includegraphics[width=\linewidth]{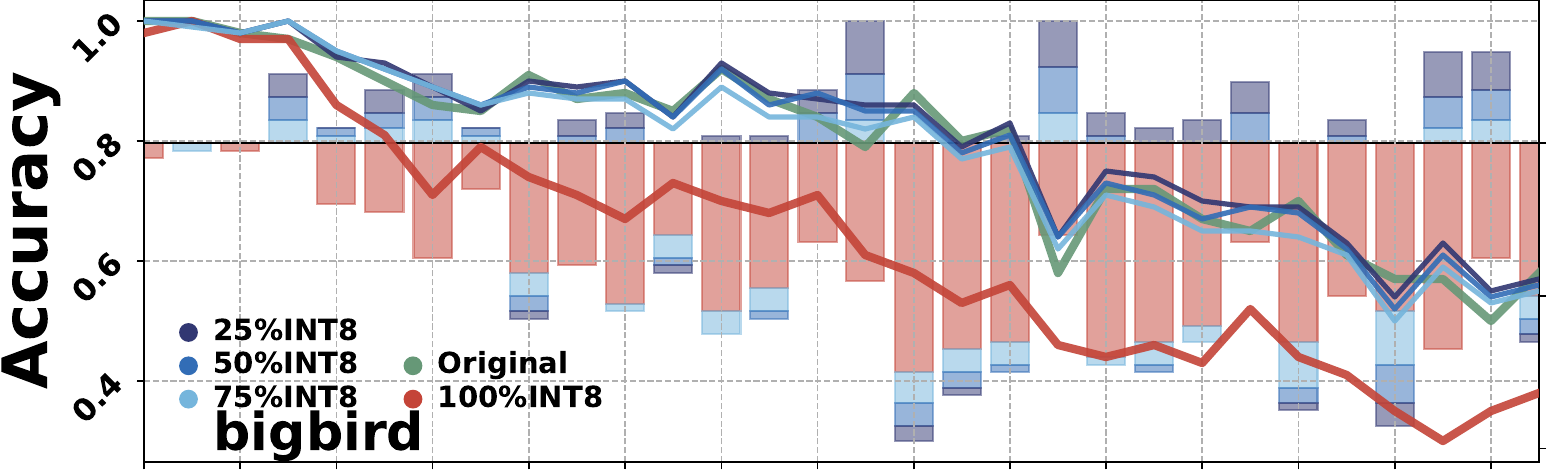}
    \end{subfigure}
    \hfill
    \begin{subfigure}[b]{0.487\linewidth}
        \centering
        \includegraphics[width=\linewidth]{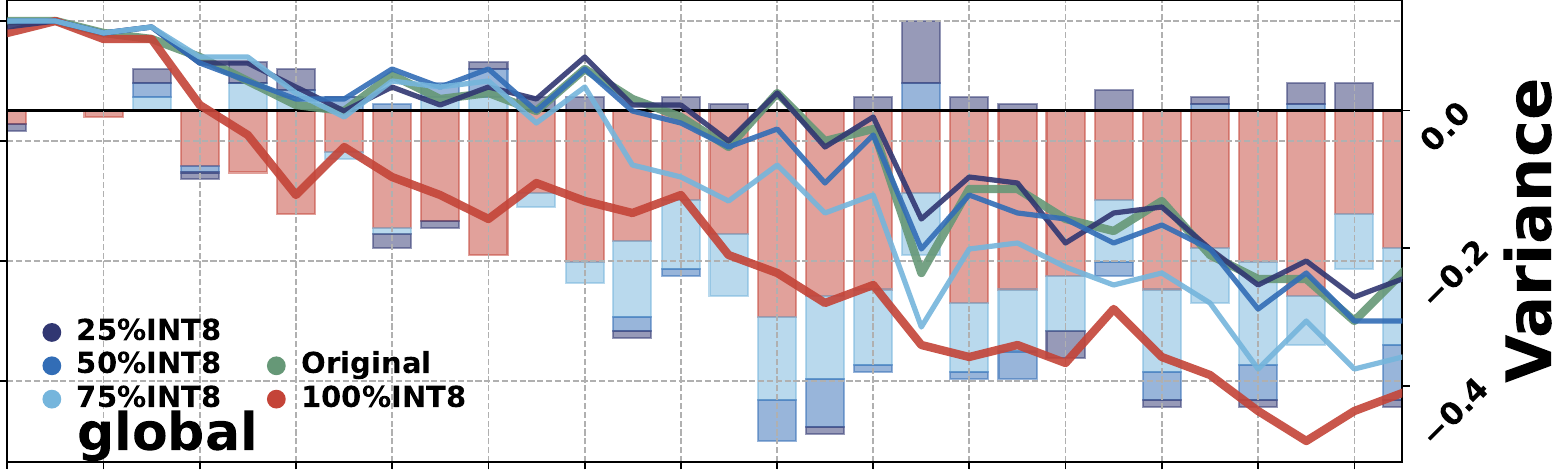}
    \end{subfigure}

    \begin{subfigure}[b]{0.485\linewidth}
        \centering
        \includegraphics[width=\linewidth]{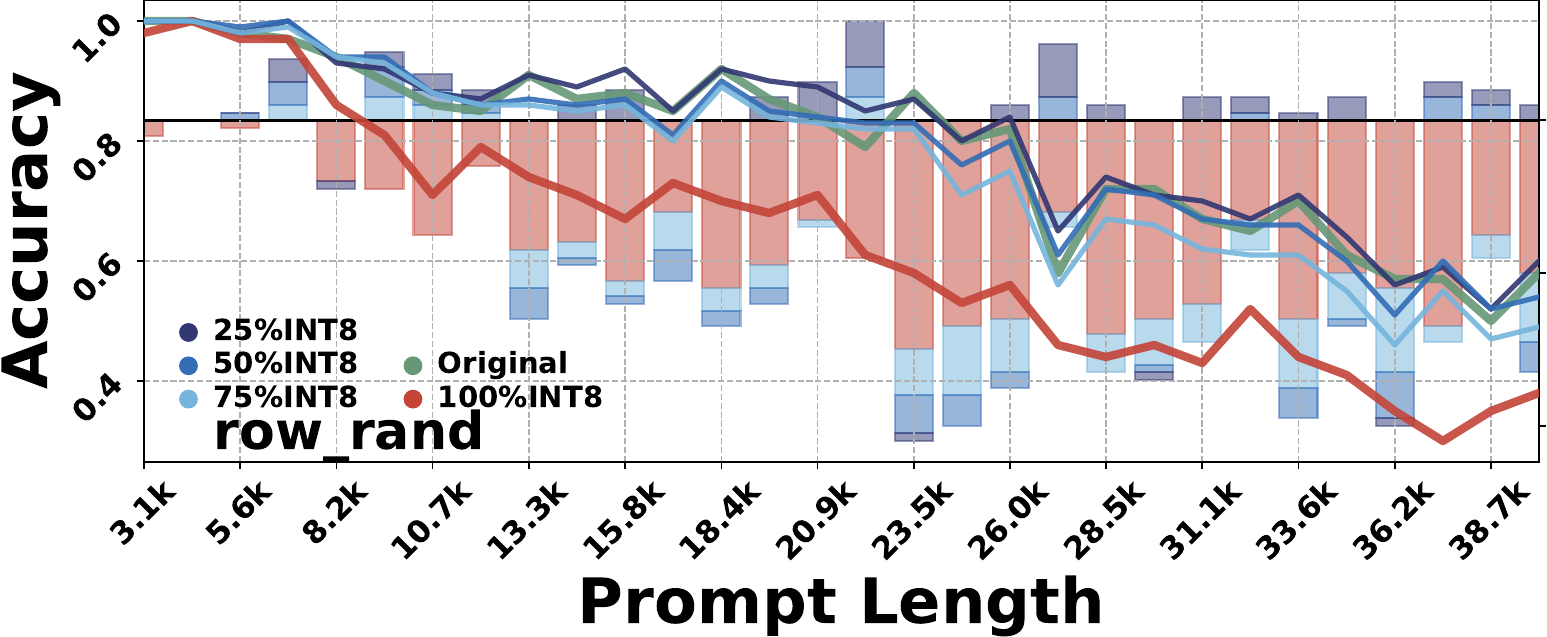}
    \end{subfigure}
    \hfill
    \begin{subfigure}[b]{0.497\linewidth}
        \centering
        \includegraphics[width=\linewidth]{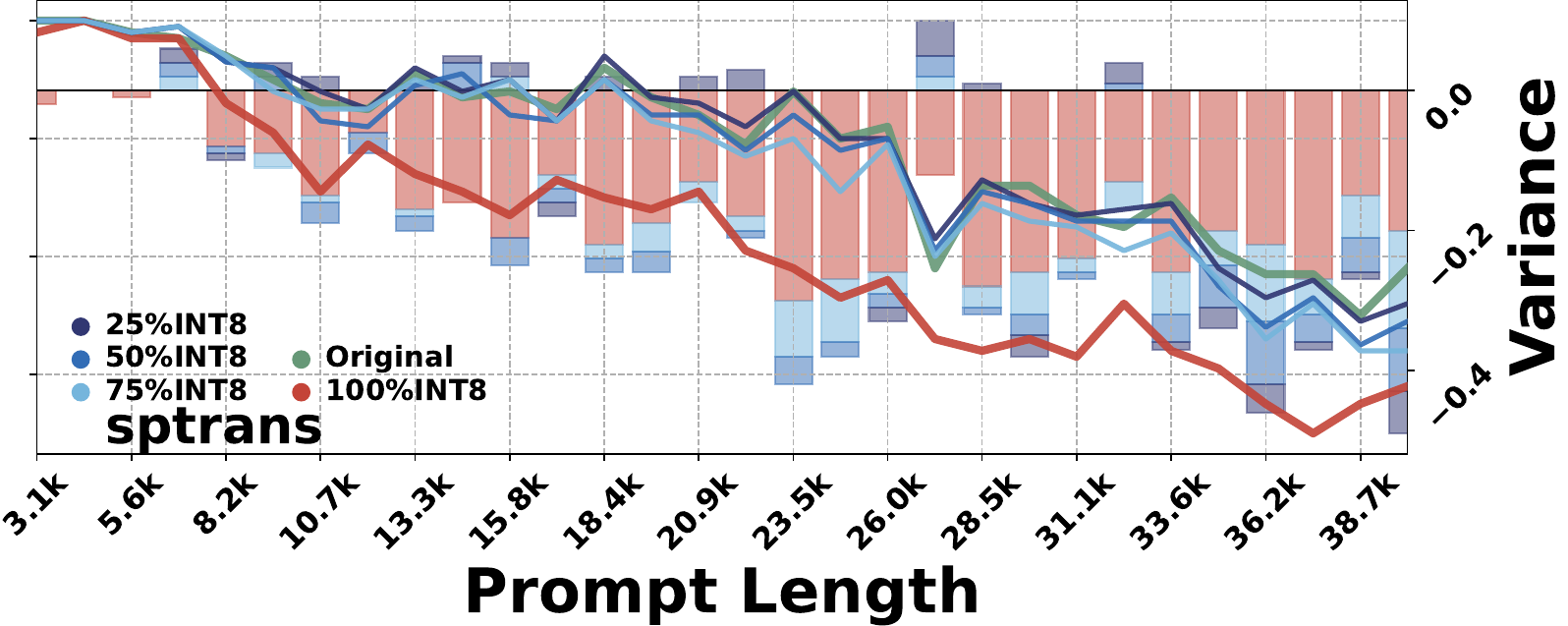}
    \end{subfigure}
    \vspace{-10pt}
    \caption{
    Line-level retrieval accuracy on the LongEval benchmark for \textbf{LLaMA 3.2 3B} under different tile-group routing layouts, evaluated across prompt lengths from 3.1k to 38.7k tokens.
    Line plots (left y-axis) report exact-match retrieval accuracy, while bar plots (right y-axis, $\Delta$ accuracy vs.\ FP16) show differences relative to the FP16 attention baseline.
    Each panel corresponds to a routing layout with 25\%, 50\%, or 75\% of tile groups routed to INT8.
    Colors indicate routing settings: \textcolor{baselinegreen}{green} denotes the original FP16 attention baseline, \textcolor{onered}{red} denotes \textbf{One (all legal score-tile groups routed to INT8)}, and \textcolor{darkblue}{blue} denotes mixed configurations with partially INT8-routed tile groups.
    }
    \vspace{-10pt}
    \label{fig:longeval_llama3b}
\end{figure*}

The INT8 path computes $Q_{8,m}K_{8,n}^{\top}$ with INT8 MMA and INT32 accumulation, then applies block scales and $1/\sqrt{d}$ to produce $S_m^{\,n}$. FP16-routed tiles compute $Q_mK_n^\top/\sqrt{d}$ through the FP16 path and update the same online-softmax state. Variable-length execution stores compact per-block scales indexed through the fused kernel's prefix-sum metadata.
Appendix~\ref{app:storage_layout} distinguishes this cache interface from the prefill path and details the tensor, scale, temporary-buffer, and data-movement layouts. 

Finally, tile-group policies become inner-loop dispatch through constant-time bitmask lookup. The selected \textbf{FP16} and \textbf{INT8} score paths update a shared FP16 online-softmax state. Together, compact metadata, constant-time lookup, and shared-state execution integrate heterogeneous score arithmetic into one dense fused attention computation.


\vspace{-5pt}
\section{Experiments}
\vspace{-5pt}

\subsection{Experimental Setup}

We evaluate \modelname from four perspectives: (i) long-context retrieval, (ii) long-context question answering, (iii) prefill efficiency, and (iv) numerical behavior under routed precision.
The main evaluation uses LLaMA 3.2 3B~\cite{meta2024llama3.2}. Appendix~\ref{app:model-performance} extends quality evaluation to Vicuna-7B, Qwen-2-7B~\cite{team2024qwen2}, and Qwen-2.5-7B~\cite{qwen2.5}, while Appendix~\ref{app:efficiency} reports efficiency across the same model families.

\noindent
\textbf{Dataset and Metrics.} 
We evaluate long-context question answering on LV-Eval~\cite{yuan2024lv}, which covers 11 English and Chinese datasets at context lengths from 16k to 64k tokens.
We use LongEval~\cite{li2023long} to evaluate line-level retrieval with exact-match accuracy. Each LongEval example embeds a uniquely labeled target line in a sequence of up to 54k tokens and queries the model for its associated content.

\noindent
\textbf{Baselines.}
We compare against (i) dense FP16 attention as the full-precision reference, (ii) \textbf{One}, which routes all legal score-tile groups to INT8 using the same \modelname kernel substrate, (iii) FlashAttention as the IO-aware FP16 execution baseline, (iv) MInference~\cite{jiang2024minference} and FlexPrefill~\cite{lai2025flexprefill} as sparse long-context baselines, and (v) SageAttention~\cite{zhang2025sageattentionaccurate8bitattention} as a representative INT8 attention kernel.

\noindent
\textbf{Hardware and Measurement.}
All experiments run on NVIDIA A100 40GB GPUs.
\modelname's offline autotuner selects an A100 kernel configuration that remains fixed throughout evaluation.
Throughput evaluation uses batch size 8, three warmup iterations, and five timed iterations with the same hardware and model wrapper for all methods. Timings include quantization, scale restoration, routing, memory staging, and kernel-scheduling costs.
We therefore report end-to-end prefill throughput rather than isolated MMA throughput.

\begin{table*}[thb]
\centering
\scriptsize
\setlength{\tabcolsep}{1.8pt}
\renewcommand{\arraystretch}{0.9}

\begin{adjustbox}{width=\textwidth}
\begin{tabular}{c @{\vsep}
                c @{\vsep}
                c @{\vsep}
                c @{\vsep}
                c c c @{\vsep}
                c c c @{\vsep}
                c @{\vsep}
                c @{\vsep}
                c}
\toprule
\textbf{Dataset} &
\textbf{Len} &
\textbf{FP16} &
\textbf{One} &
\textbf{SpTrans25} &
\textbf{SpTrans50} &
\textbf{SpTrans75} &
\textbf{BigBird25} &
\textbf{BigBird50} &
\textbf{BigBird75} &
\textbf{MInference} &
\textbf{FlexPrefill} &
\textbf{SageAttn} \\
\midrule

\multirow{3}{*}{\ds{1}}
& 16k \setbase{32.04}\setone{28.78}
& \cellcolor{baselinegreen}\base
& \cellcolor{onered}\one
& \tcell{31.75} & \tcell{32.18} & \tcell{30.25}
& \tcell{31.27} & \tcell{29.64} & \tcell{29.75}
& \tcell{26.93} & \tcell{27.11} & \tcell{29.79} \\
& 32k \setbase{15.08}\setone{11.62}
& \cellcolor{baselinegreen}\base
& \cellcolor{onered}\one
& \tcell{15.56} & \tcell{14.77} & \tcell{14.37}
& \tcell{15.22} & \tcell{14.02} & \tcell{13.55}
& \tcell{11.55} & \tcell{11.74} & \tcell{13.50} \\
& 64k \setbase{7.75}\setone{5.42}
& \cellcolor{baselinegreen}\base
& \cellcolor{onered}\one
& \tcell{8.01} & \tcell{7.44} & \tcell{6.83}
& \tcell{8.01} & \tcell{6.47} & \tcell{5.72}
& \tcell{5.31} & \tcell{5.39} & \tcell{5.77} \\
\midrule

\multirow{3}{*}{\ds{2}}
& 16k \setbase{18.49}\setone{15.62}
& \cellcolor{baselinegreen}\base
& \cellcolor{onered}\one
& \tcell{18.64} & \tcell{18.08} & \tcell{16.78}
& \tcell{18.61} & \tcell{16.98} & \tcell{16.62}
& \tcell{14.61} & \tcell{14.92} & \tcell{16.67} \\
& 32k \setbase{15.12}\setone{12.02}
& \cellcolor{baselinegreen}\base
& \cellcolor{onered}\one
& \tcell{15.27} & \tcell{14.69} & \tcell{14.18}
& \tcell{15.14} & \tcell{13.53} & \tcell{13.28}
& \tcell{12.01} & \tcell{11.64} & \tcell{14.59} \\
& 64k \setbase{11.84}\setone{7.79}
& \cellcolor{baselinegreen}\base
& \cellcolor{onered}\one
& \tcell{12.06} & \tcell{11.34} & \tcell{11.04}
& \tcell{12.04} & \tcell{10.28} & \tcell{10.03}
& \tcell{7.96} & \tcell{6.81} & \tcell{11.22} \\
\midrule

\multirow{3}{*}{\ds{3}}
& 16k \setbase{6.72}\setone{4.45}
& \cellcolor{baselinegreen}\base
& \cellcolor{onered}\one
& \tcell{21.04} & \tcell{20.53} & \tcell{21.65}
& \tcell{6.29} & \tcell{6.97} & \tcell{4.57}
& \tcell{5.41} & \tcell{5.01} & \tcell{5.88} \\
& 32k \setbase{3.78}\setone{1.95}
& \cellcolor{baselinegreen}\base
& \cellcolor{onered}\one
& \tcell{3.35} & \tcell{2.78} & \tcell{2.68}
& \tcell{2.35} & \tcell{1.81} & \tcell{1.20}
& \tcell{2.75} & \tcell{2.51} & \tcell{3.15} \\
& 64k \setbase{3.38}\setone{1.42}
& \cellcolor{baselinegreen}\base
& \cellcolor{onered}\one
& \tcell{3.24} & \tcell{2.67} & \tcell{2.47}
& \tcell{3.24} & \tcell{1.70} & \tcell{1.37}
& \tcell{2.13} & \tcell{1.91} & \tcell{2.73} \\
\midrule

\multirow{3}{*}{\ds{4}}
& 16k \setbase{6.00}\setone{2.27}
& \cellcolor{baselinegreen}\base
& \cellcolor{onered}\one
& \tcell{6.00} & \tcell{5.43} & \tcell{5.04}
& \tcell{6.00} & \tcell{4.46} & \tcell{4.49}
& \tcell{4.77} & \tcell{4.53} & \tcell{5.19} \\
& 32k \setbase{18.00}\setone{13.98}
& \cellcolor{baselinegreen}\base
& \cellcolor{onered}\one
& \tcell{18.00} & \tcell{17.43} & \tcell{17.54}
& \tcell{18.50} & \tcell{16.46} & \tcell{15.99}
& \tcell{12.99} & \tcell{11.99} & \tcell{15.08} \\
& 64k \setbase{14.50}\setone{10.07}
& \cellcolor{baselinegreen}\base
& \cellcolor{onered}\one
& \tcell{15.00} & \tcell{14.43} & \tcell{14.04}
& \tcell{15.00} & \tcell{13.46} & \tcell{12.49}
& \tcell{9.03} & \tcell{8.31} & \tcell{11.55} \\
\midrule

\multirow{3}{*}{\ds{5}}
& 16k \setbase{17.51}\setone{12.86}
& \cellcolor{baselinegreen}\base
& \cellcolor{onered}\one
& \tcell{17.66} & \tcell{17.09} & \tcell{16.88}
& \tcell{17.66} & \tcell{15.17} & \tcell{14.84}
& \tcell{13.97} & \tcell{13.16} & \tcell{15.27} \\
& 32k \setbase{10.80}\setone{8.38}
& \cellcolor{baselinegreen}\base
& \cellcolor{onered}\one
& \tcell{11.01} & \tcell{10.44} & \tcell{10.06}
& \tcell{10.61} & \tcell{9.07} & \tcell{9.00}
& \tcell{7.81} & \tcell{7.20} & \tcell{9.11} \\
& 64k \setbase{6.82}\setone{5.32}
& \cellcolor{baselinegreen}\base
& \cellcolor{onered}\one
& \tcell{7.63} & \tcell{6.25} & \tcell{5.97}
& \tcell{7.63} & \tcell{6.09} & \tcell{5.62}
& \tcell{4.26} & \tcell{3.86} & \tcell{5.50} \\
\midrule

\multirow{3}{*}{\ds{6}}
& 16k \setbase{12.33}\setone{8.51}
& \cellcolor{baselinegreen}\base
& \cellcolor{onered}\one
& \tcell{12.31} & \tcell{11.74} & \tcell{11.03}
& \tcell{12.38} & \tcell{10.40} & \tcell{10.30}
& \tcell{9.82} & \tcell{9.28} & \tcell{10.69} \\
& 32k \setbase{6.73}\setone{3.22}
& \cellcolor{baselinegreen}\base
& \cellcolor{onered}\one
& \tcell{7.08} & \tcell{6.51} & \tcell{5.95}
& \tcell{6.52} & \tcell{5.18} & \tcell{4.45}
& \tcell{4.88} & \tcell{4.48} & \tcell{5.69} \\
& 64k \setbase{1.84}\setone{0.74}
& \cellcolor{baselinegreen}\base
& \cellcolor{onered}\one
& \tcell{1.97} & \tcell{1.37} & \tcell{1.22}
& \tcell{1.91} & \tcell{1.35} & \tcell{1.13}
& \tcell{1.12} & \tcell{1.07} & \tcell{1.44} \\
\midrule

\multirow{3}{*}{\ds{7}}
& 16k \setbase{21.75}\setone{15.69}
& \cellcolor{baselinegreen}\base
& \cellcolor{onered}\one
& \tcell{22.07} & \tcell{21.16} & \tcell{21.05}
& \tcell{21.96} & \tcell{20.81} & \tcell{20.09}
& \tcell{15.59} & \tcell{14.05} & \tcell{19.94} \\
& 32k \setbase{19.89}\setone{12.11}
& \cellcolor{baselinegreen}\base
& \cellcolor{onered}\one
& \tcell{19.91} & \tcell{19.37} & \tcell{18.56}
& \tcell{20.17} & \tcell{18.41} & \tcell{17.79}
& \tcell{12.23} & \tcell{10.95} & \tcell{15.92} \\
& 64k \setbase{14.18}\setone{6.93}
& \cellcolor{baselinegreen}\base
& \cellcolor{onered}\one
& \tcell{14.23} & \tcell{13.66} & \tcell{13.34}
& \tcell{14.09} & \tcell{12.37} & \tcell{11.97}
& \tcell{7.85} & \tcell{6.67} & \tcell{11.24} \\
\midrule

\multirow{3}{*}{\ds{8}}
& 16k \setbase{18.24}\setone{12.72}
& \cellcolor{baselinegreen}\base
& \cellcolor{onered}\one
& \tcell{17.53} & \tcell{16.99} & \tcell{17.46}
& \tcell{17.57} & \tcell{16.45} & \tcell{15.96}
& \tcell{12.32} & \tcell{11.39} & \tcell{13.93} \\
& 32k \setbase{14.25}\setone{9.12}
& \cellcolor{baselinegreen}\base
& \cellcolor{onered}\one
& \tcell{13.63} & \tcell{13.18} & \tcell{13.00}
& \tcell{13.73} & \tcell{12.73} & \tcell{12.29}
& \tcell{9.26} & \tcell{7.40} & \tcell{10.54} \\
& 64k \setbase{11.08}\setone{7.54}
& \cellcolor{baselinegreen}\base
& \cellcolor{onered}\one
& \tcell{11.31} & \tcell{10.65} & \tcell{10.38}
& \tcell{11.05} & \tcell{9.60} & \tcell{9.04}
& \tcell{7.24} & \tcell{5.08} & \tcell{9.45} \\
\midrule

\multirow{3}{*}{\ds{9}}
& 16k \setbase{45.68}\setone{34.70}
& \cellcolor{baselinegreen}\base
& \cellcolor{onered}\one
& \tcell{45.77} & \tcell{44.60} & \tcell{44.88}
& \tcell{45.13} & \tcell{43.66} & \tcell{43.86}
& \tcell{38.16} & \tcell{35.24} & \tcell{42.12} \\
& 32k \setbase{26.84}\setone{18.17}
& \cellcolor{baselinegreen}\base
& \cellcolor{onered}\one
& \tcell{26.05} & \tcell{25.94} & \tcell{25.11}
& \tcell{26.47} & \tcell{24.94} & \tcell{23.78}
& \tcell{20.09} & \tcell{21.40} & \tcell{22.50} \\
& 64k \setbase{16.36}\setone{9.19}
& \cellcolor{baselinegreen}\base
& \cellcolor{onered}\one
& \tcell{16.03} & \tcell{15.25} & \tcell{14.71}
& \tcell{15.72} & \tcell{14.18} & \tcell{14.04}
& \tcell{10.24} & \tcell{10.69} & \tcell{13.21} \\
\midrule

\multirow{3}{*}{\ds{10}}
& 16k \setbase{28.88}\setone{22.37}
& \cellcolor{baselinegreen}\base
& \cellcolor{onered}\one
& \tcell{29.10} & \tcell{28.31} & \tcell{27.23}
& \tcell{28.95} & \tcell{27.57} & \tcell{26.33}
& \tcell{23.28} & \tcell{23.43} & \tcell{25.18} \\
& 32k \setbase{18.33}\setone{13.66}
& \cellcolor{baselinegreen}\base
& \cellcolor{onered}\one
& \tcell{18.26} & \tcell{17.65} & \tcell{17.41}
& \tcell{18.17} & \tcell{16.70} & \tcell{16.88}
& \tcell{12.70} & \tcell{10.28} & \tcell{13.94} \\
& 64k \setbase{15.75}\setone{12.81}
& \cellcolor{baselinegreen}\base
& \cellcolor{onered}\one
& \tcell{15.94} & \tcell{15.32} & \tcell{15.39}
& \tcell{15.93} & \tcell{14.22} & \tcell{13.88}
& \tcell{9.42} & \tcell{7.68} & \tcell{13.37} \\
\midrule

\multirow{3}{*}{\ds{11}}
& 16k \setbase{22.44}\setone{16.97}
& \cellcolor{baselinegreen}\base
& \cellcolor{onered}\one
& \tcell{22.94} & \tcell{22.37} & \tcell{21.74}
& \tcell{22.44} & \tcell{20.24} & \tcell{20.76}
& \tcell{15.99} & \tcell{12.64} & \tcell{18.19} \\
& 32k \setbase{14.85}\setone{12.02}
& \cellcolor{baselinegreen}\base
& \cellcolor{onered}\one
& \tcell{15.61} & \tcell{15.04} & \tcell{15.06}
& \tcell{16.11} & \tcell{14.57} & \tcell{13.34}
& \tcell{11.33} & \tcell{10.25} & \tcell{14.57} \\
& 64k \setbase{9.26}\setone{3.41}
& \cellcolor{baselinegreen}\base
& \cellcolor{onered}\one
& \tcell{8.92} & \tcell{8.35} & \tcell{7.40}
& \tcell{8.92} & \tcell{8.38} & \tcell{7.91}
& \tcell{3.10} & \tcell{4.02} & \tcell{5.79} \\
\bottomrule
\end{tabular}
\end{adjustbox}
\vspace{-10pt}
\caption{
LV-Eval long-context question answering accuracy for \textbf{LLaMA 3.2 3B}.
We compare FP16 attention, \textbf{One (100\% INT8)}, representative \modelname routing layouts, sparse long-context baselines (MInference and FlexPrefill), and an INT8 attention baseline (SageAttention).
Sparse baselines reduce computation by pruning token interactions, whereas \modelname preserves dense legal connectivity and routes legal tile groups to different score-computation paths.
Cells are color-coded by comparison to \textbf{One} and FP16: \textcolor{onered}{red} indicates performance below \textbf{One}, \textcolor{neutralblue}{blue} indicates performance between \textbf{One} and FP16, and \textcolor{baselinegreen}{green} indicates performance above FP16.
}
\label{tab:lveval_llama_sota}
\vspace{-15pt}
\end{table*}

\subsection{Model Performance}\label{sec:model_performance}
\vspace{-5pt}

\noindent
\textbf{Long-context Retrieval.}
LongEval evaluates exact retrieval from sequences of labeled lines.
Each input contains entries such as ``line wacky-cob: CONTENT'', and the query specifies a label.
The model returns the associated content, with exact-match accuracy measuring retrieval success.

Figure~\ref{fig:longeval_llama3b} reports LLaMA 3.2 3B retrieval accuracy from 3.1k to 38.7k tokens. FP16 and \textbf{One} provide the full-precision and matched uniform-INT8 references, respectively. Each panel shows one spatial routing layout at 25\%, 50\%, and 75\% INT8 tile-group coverage. Retrieval quality depends on FP16 placement in addition to nominal INT8 coverage. \texttt{row\_rand} and \texttt{sptrans} retain stronger accuracy as INT8 coverage increases, while \texttt{align\_sparse}, \texttt{band}, and \texttt{global} retain more quality under conservative coverage.
Together, routing layout determines where FP16 computation is retained, while coverage controls INT8 execution.

\noindent
\textbf{Long-context Question Answering.}
LV-Eval covers 11 English and Chinese long-context QA datasets under confusing-fact insertion (CFI), keyword and phrase replacement (KPR), and keyword-recall (AK) settings.
Datasets \ds{1} \textit{cmrc\_mixup} and \ds{2} \textit{dureader\_mixup} cover Chinese machine reading comprehension and question answering, and datasets \ds{3} \textit{factrecall\_en} and \ds{4} \textit{factrecall\_zh} cover bilingual factual recall.
Dataset \ds{5} \textit{hotpotwikiqa\_mixup} evaluates multi-hop retrieval, dataset \ds{6} \textit{lic\_mixup} evaluates precise information localization, and datasets \ds{7} \textit{loogle\_CR\_mixup}, \ds{8} \textit{loogle\_MIR\_mixup}, and \ds{9} \textit{loogle\_SD\_mixup} evaluate content recall, multi-information retrieval, and sequential dependency retrieval.
Datasets \ds{10} \textit{multifieldqa\_en\_mixup} and \ds{11} \textit{multifieldqa\_zh\_mixup} evaluate bilingual multi-field question answering.

\begin{table*}[thb]
\centering
\scriptsize
\setlength{\tabcolsep}{1.8pt}
\renewcommand{\arraystretch}{0.9}
\vspace{-15pt}
\begin{adjustbox}{width=\textwidth}
\begin{tabular}{
c @{\vsep}
c @{\vsep}
c @{\vsep}
c @{\vsep}
c @{\vsep}
c @{\vsep}
c @{\vsep}
c @{\vsep}
c @{\vsep}
c @{\vsep}
c @{\vsep}
c @{\vsep}
c @{\vsep}
c
}
\toprule
\textbf{Len} &
\textbf{Metric} &
\textbf{Torch} &
\textbf{FlashAttn} &
\textbf{One} &
\textbf{SpTrans25} &
\textbf{SpTrans50} &
\textbf{SpTrans75} &
\textbf{BigBird25} &
\textbf{BigBird50} &
\textbf{BigBird75} &
\textbf{MInference} &
\textbf{FlexPrefill} &
\textbf{SageAttn} \\
\midrule

1k & Thpt
& \setlow{17.45}\setup{32.27}\tcellFO{11.14}
& \cellcolor{onered!35}17.45
& \cellcolor{baselinegreen!35}32.27
& \tcellFO{27.66}
& \tcellFO{32.11}
& \tcellFO{33.50}
& \tcellFO{26.92}
& \tcellFO{31.48}
& \tcellFO{32.37}
& \tcellFO{1.96}
& \tcellFO{8.01}
& \tcellFO{19.91} \\

& TOPS
& \setlow{65.31}\setup{120.80}\tcellFO{41.70}
& \cellcolor{onered!35}65.31
& \cellcolor{baselinegreen!35}120.80
& \tcellFO{103.50}
& \tcellFO{120.20}
& \tcellFO{125.38}
& \tcellFO{100.78}
& \tcellFO{117.85}
& \tcellFO{121.17}
& \tcellFO{7.33}
& \tcellFO{29.98}
& \tcellFO{74.54} \\

\midrule

2k & Thpt
& \setlow{16.48}\setup{32.06}\tcellFO{7.78}
& \cellcolor{onered!35}16.48
& \cellcolor{baselinegreen!35}32.06
& \tcellFO{27.11}
& \tcellFO{31.46}
& \tcellFO{33.92}
& \tcellFO{26.57}
& \tcellFO{28.83}
& \tcellFO{32.71}
& \tcellFO{2.79}
& \tcellFO{12.32}
& \tcellFO{20.20} \\

& TOPS
& \setlow{64.64}\setup{125.70}\tcellFO{30.50}
& \cellcolor{onered!35}64.64
& \cellcolor{baselinegreen!35}125.70
& \tcellFO{106.32}
& \tcellFO{123.38}
& \tcellFO{132.99}
& \tcellFO{104.23}
& \tcellFO{113.05}
& \tcellFO{128.31}
& \tcellFO{10.93}
& \tcellFO{48.30}
& \tcellFO{79.19} \\

\midrule

4k & Thpt
& \setlow{14.33}\setup{29.80}\tcellFO{OOM}
& \cellcolor{onered!35}14.33
& \cellcolor{baselinegreen!35}29.80
& \tcellFO{27.14}
& \tcellFO{30.59}
& \tcellFO{31.80}
& \tcellFO{24.69}
& \tcellFO{26.84}
& \tcellFO{27.48}
& \tcellFO{4.70}
& \tcellFO{15.45}
& \tcellFO{19.91} \\

& TOPS
& \setlow{61.31}\setup{127.48}\tcellFO{OOM}
& \cellcolor{onered!35}61.31
& \cellcolor{baselinegreen!35}127.48
& \tcellFO{116.09}
& \tcellFO{130.85}
& \tcellFO{136.03}
& \tcellFO{105.63}
& \tcellFO{114.83}
& \tcellFO{117.56}
& \tcellFO{20.13}
& \tcellFO{66.11}
& \tcellFO{85.19} \\

\midrule

8k & Thpt
& \setlow{0}\setup{27.41}\tcellFO{OOM}
& \oom
& \cellcolor{baselinegreen!35}27.41
& \tcellFO{22.81}
& \tcellFO{25.69}
& \tcellFO{26.61}
& \tcellFO{23.02}
& \tcellFO{25.17}
& \tcellFO{25.84}
& \tcellFO{7.24}
& \tcellFO{15.80}
& \tcellFO{18.79} \\

& TOPS
& \setlow{0}\setup{136.76}\tcellFO{OOM}
& \oom
& \cellcolor{baselinegreen!35}136.76
& \tcellFO{113.83}
& \tcellFO{128.21}
& \tcellFO{132.84}
& \tcellFO{114.92}
& \tcellFO{125.58}
& \tcellFO{128.90}
& \tcellFO{36.13}
& \tcellFO{78.84}
& \tcellFO{93.79} \\

\bottomrule
\end{tabular}
\vspace{-10pt}
\end{adjustbox}
\vspace{-5pt}
\caption{
Implementation-level prefill throughput (Thpt, K tokens/s) and TOPS for \textbf{LLaMA 3.2 3B-Instruct}.
All methods use the same A100 40GB hardware, batch size 8, three warmup iterations, five timed iterations, and model wrapper; timings include method-specific quantization, scale restoration, routing, memory staging, and scheduling costs.
TOPS uses a common dense-attention operation count divided by measured end-to-end time.
Colors compare each row with FlashAttention and \textbf{One}: red indicates performance below FlashAttention, blue indicates performance between FlashAttention and \textbf{One}, and green indicates performance above \textbf{One}; OOM entries are gray.
}
\label{tab:efficiency_llama3.2_sota}
\vspace{-15pt}
\end{table*}

Table~\ref{tab:lveval_llama_sota} compares FP16, the uniform-INT8 reference \textbf{One}, representative \modelname configurations, sparse methods, and SageAttention. \textbf{One} often trails FP16, showing that uniform INT8 can be too coarse for long-context QA. \modelname narrows this gap by assigning FP16 to selected score-tile groups and INT8 to the rest while preserving all legal interactions. The strong \texttt{sptrans} result at 16k factual recall recurs across LLaMA and Qwen models at all coverage ratios, indicating a consistent layout--task interaction.
Appendix~\ref{app:model-performance} reports complete results across models and routing layouts.

\begin{table}[ht]
\vspace{-5pt}
\centering
\small
\setlength{\tabcolsep}{8pt}
\renewcommand{\arraystretch}{0.9}
\begin{tabular}{c @{\vsep} ccc}
\toprule
\textbf{Seq Len} &
\textbf{0\%} &
\textbf{5\%} &
\textbf{10\%} \\
\midrule
1k & $7.27{\times}10^{-5}$ & $7.47{\times}10^{-4}$ & $1.19{\times}10^{-3}$ \\
2k & $5.89{\times}10^{-5}$ & $4.29{\times}10^{-4}$ & $1.16{\times}10^{-3}$ \\
4k & $4.87{\times}10^{-5}$ & $6.71{\times}10^{-4}$ & $1.10{\times}10^{-3}$ \\
8k & $6.84{\times}10^{-5}$ & $8.34{\times}10^{-4}$ & $6.32{\times}10^{-3}$ \\
\midrule
\textbf{Seq Len} &
\textbf{15\%} &
\textbf{20\%} &
\textbf{25\%} \\
\midrule
1k & $1.56{\times}10^{-3}$ & $1.67{\times}10^{-3}$ & $2.03{\times}10^{-3}$ \\
2k & $1.49{\times}10^{-3}$ & $1.66{\times}10^{-3}$ & $1.84{\times}10^{-3}$ \\
4k & $1.41{\times}10^{-3}$ & $1.59{\times}10^{-3}$ & $1.78{\times}10^{-3}$ \\
8k & $6.59{\times}10^{-3}$ & $6.71{\times}10^{-3}$ & $6.84{\times}10^{-3}$ \\
\bottomrule
\end{tabular}
\vspace{-10pt}
\caption{
Mean absolute output deviation from the fixed Torch FP16 reference under different score-tile-group INT8 coverage ratios; lower values indicate closer agreement with the reference implementation.
}
\label{tab:int8_ratio_numerical_deviation}
\vspace{-10pt}
\end{table}

\vspace{-5pt}
\subsection{Efficiency}\label{sec:efficiency}
\vspace{-5pt}

Table~\ref{tab:efficiency_llama3.2_sota} reports prefill throughput and TOPS for LLaMA 3.2 3B-Instruct from 1k to 8k tokens. Under this protocol, \modelname improves throughput over FlashAttention by routing score-tile groups to INT8 Tensor Cores. At 4k tokens, \texttt{SpTrans75} reaches 31.80 K tokens/s, compared with 14.33 K tokens/s for FlashAttention and 29.80 K tokens/s for \textbf{One}. \modelname preserves dense legal connectivity while changing selected tile-group arithmetic paths. End-to-end throughput includes quantization, scale restoration, routing, memory staging, and kernel scheduling.
The ordering between \textbf{One} and high-coverage mixed configurations reflects the complete pipeline, including layout-dependent dispatch and memory behavior.
Appendix~\ref{app:efficiency} reports complete throughput and TOPS across models.

\vspace{-5pt}
\subsection{Attention-Kernel Numerical Behavior}\label{sec:numerical_analysis}
\vspace{-5pt}

We measure mean absolute deviation from a fixed Torch FP16 reference on randomized attention inputs.
Differences reflect quantization, scale restoration, accumulation, reduction order, and rounding across routed FP16 and INT8 score paths.

Table~\ref{tab:int8_ratio_numerical_deviation} reports deviation across INT8 coverage ratios and sequence lengths.
The 0\% configuration remains close to FP16, while deviation generally increases with INT8 coverage.
At 8k tokens, deviation increases markedly between 5\% and 10\% coverage and remains at a similar scale through 25\%, showing that coverage provides a practical numerical-control knob.
Appendix~\ref{app:numerical-analysis} extends the analysis with model-depth and sequence-length studies, fused-kernel comparisons, FP16-versus-FP32 accumulation controls, larger-model checks, and heavy-hitter exposure under static routing.

\vspace{-5pt}
\section{Conclusion}
\vspace{-5pt}
\modelname establishes score-tile-group precision routing as a kernel-native abstraction for long-context attention. Its kernel assigns FP16 and INT8 paths across legal tile groups, preserving the attention graph while varying precision within one computation. Scalable grouping lets each routing decision span adjacent key tiles while retaining hardware-aligned compute tiles and compact metadata. Packed bitmasks and constant-time lookup provide inner-loop control, while both paths update a shared online-softmax state. It supports training-free deployment, grouped-query attention, variable-length batching, and INT8 KV caches. Across LongEval, LV-Eval, prefill benchmarks, and numerical analyses on LLaMA, Qwen, and Vicuna, \modelname recovers long-context quality from uniform INT8 while improving prefill throughput over FP16. This establishes a controllable accuracy-efficiency frontier and demonstrates spatial precision control as a practical dimension for dense attention kernels.

\section*{Limitations}

\modelname targets forward inference during long-context prefill, the deployment setting evaluated throughout this work.
The current implementation instantiates tile-group routing with FP16 and INT8 Tensor Core paths on NVIDIA A100 GPUs.
Other numerical formats require format-specific scale handling and kernel scheduling.
Static routing templates provide deterministic policy construction, compact metadata, and constant-time kernel dispatch.
The kernel interface can consume alternative static or adaptive routing policies.

\section*{Ethical Considerations}

\modelname is a systems method for improving the efficiency of long-context LLM inference and does not introduce new training data, human-subject data, or model capabilities.
Its primary broader impact is computational.
Improved attention efficiency may reduce GPU time and energy consumption per supported workload, while lower inference cost may also increase aggregate deployment and total compute demand.

\bibliography{reference}


\appendix

\setcounter{dbltopnumber}{4}
\renewcommand{\dbltopfraction}{0.95}
\renewcommand{\dblfloatpagefraction}{0.65}
\setlength{\dbltextfloatsep}{6pt}
\makeatletter
\setlength{\@dblfptop}{0pt}
\makeatother

\section{Notations}\label{app:notations}

For clarity and ease of reference, we group notation in Table~\ref{tab:notation} by functional role rather than by order of appearance.
This organization reflects the structure of the attention computation and kernel design:
core attention definitions, tiling and indexing used by IO-aware kernels, online-softmax state variables,
blockwise quantization primitives, head mappings for grouped-query attention, and tile-group routing metadata.
Grouping symbols in this way allows readers to quickly locate related quantities when following the
kernel execution flow and routing logic described in Sections~\ref{sec:preliminary} and~\ref{sec:method}.

\begin{table*}[t]
\centering
\small
\setlength{\tabcolsep}{0pt}

\begin{tabular}{c l l}
\toprule
\textbf{Category} & \textbf{Symbol} & \textbf{Description} \\
\midrule

\multirow{6}{*}{\rotatebox{0}{\textbf{Attention}}}
& $Q, K, V$ &
Query, key, and value matrices:
$Q\in\mathbb{R}^{L_q\times d}$,
$K\in\mathbb{R}^{L_k\times d}$,
$V\in\mathbb{R}^{L_k\times d}$ \\

& $L_q, L_k$ &
Query and key lengths; $L_q=L_k=L$ for self-attention \\

& $d$ &
Query, key, and value head dimension \\

& $S$ &
Scaled score matrix:
$S=QK^\top/\sqrt{d}\in\mathbb{R}^{L_q\times L_k}$ \\

& $P$ &
Attention weights:
$P=\mathrm{softmax}(S)\in\mathbb{R}^{L_q\times L_k}$ \\

& $O$ &
Attention output:
$O=PV\in\mathbb{R}^{L_q\times d}$ \\

\midrule

\multirow{6}{*}{\rotatebox{0}{\textbf{Tiling}}}
& $b_q, b_{kv}$ &
Query and key/value tile sizes \\

& $\mathrm{BLOCK}_M, \mathrm{BLOCK}_N$ &
Hardware-aligned compute-tile sizes;
$\mathrm{BLOCK}_M\equiv b_q$,
$\mathrm{BLOCK}_N\equiv b_{kv}$ \\

& $T_m$ &
Query-tile count:
$T_m=\lceil L_q/\mathrm{BLOCK}_M\rceil$ \\

& $T_n$ &
Key/value-tile count:
$T_n=\lceil L_k/\mathrm{BLOCK}_N\rceil$ \\

& $i, j$ &
Query and key token indices \\

& $m, n$ &
Query and key/value tile indices \\

\midrule

\multirow{3}{*}{\rotatebox{0}{\textbf{Online Softmax}}}
& $\tilde m_m^{\,n}$ &
Row-wise maximum for query tile $m$ after tile $n$ \\

& $\tilde z_m^{\,n}$ &
Row-wise normalizer for query tile $m$ after tile $n$ \\

& $\tilde O_m^{\,n}$ &
Unnormalized output accumulator after tile $n$ \\

\midrule

\multirow{4}{*}{\rotatebox{0}{\textbf{Quantization}}}
& $\psi(\cdot)$ &
Blockwise quantization operator \\

& $\hat A, \hat B$ &
Low-precision forms of operands $A$ and $B$ \\

& $\delta_A, \delta_B$ &
Block scales for $\hat A$ and $\hat B$ \\

& $Q_8, K_8$ &
INT8 query and key representations \\

\midrule

\multirow{2}{*}{\rotatebox{0}{\textbf{Heads}}}
& $H_q, H_k$ &
Numbers of query and KV heads \\

& $h_q, h_k$ &
Query-head and mapped KV-head indices \\

\midrule

\multirow{7}{*}{\rotatebox{0}{\textbf{Routing Policy}}}
& $\mathrm{BLOCK}_N^{\text{mask}}$ &
Routing-group width along the key dimension \\

& $g$ &
Key-tile grouping factor:
$\mathrm{BLOCK}_N^{\text{mask}}=g\,\mathrm{BLOCK}_N$ \\

& $T_{\text{mask}}$ &
Groups per query-tile row:
$T_{\text{mask}}
=\lceil L_k/\mathrm{BLOCK}_N^{\text{mask}}\rceil\le 64$ \\

& $g_j$ &
Group index for key tile $n$:
$g_j=\lfloor n/g\rfloor$ \\

& $R_{h_k,m,g_j}$ &
Binary route: 1 for INT8 and 0 for FP16 \\

& $b_{h_k,m}$ &
Packed 64-bit routing word \\

& $\rho_{\mathrm{INT8}}$ &
Fraction of legal groups routed to INT8 \\

\bottomrule
\end{tabular}
\caption{
Notation for attention, tiling, online softmax, quantization, head mapping, and routing.
}
\label{tab:notation}
\end{table*}

\vspace{-5pt}
\section{Precision Layouts}\label{app:precision-layouts}

\vspace{-5pt}

\begin{figure}[t]
    \centering
    \includegraphics[width=1\linewidth]{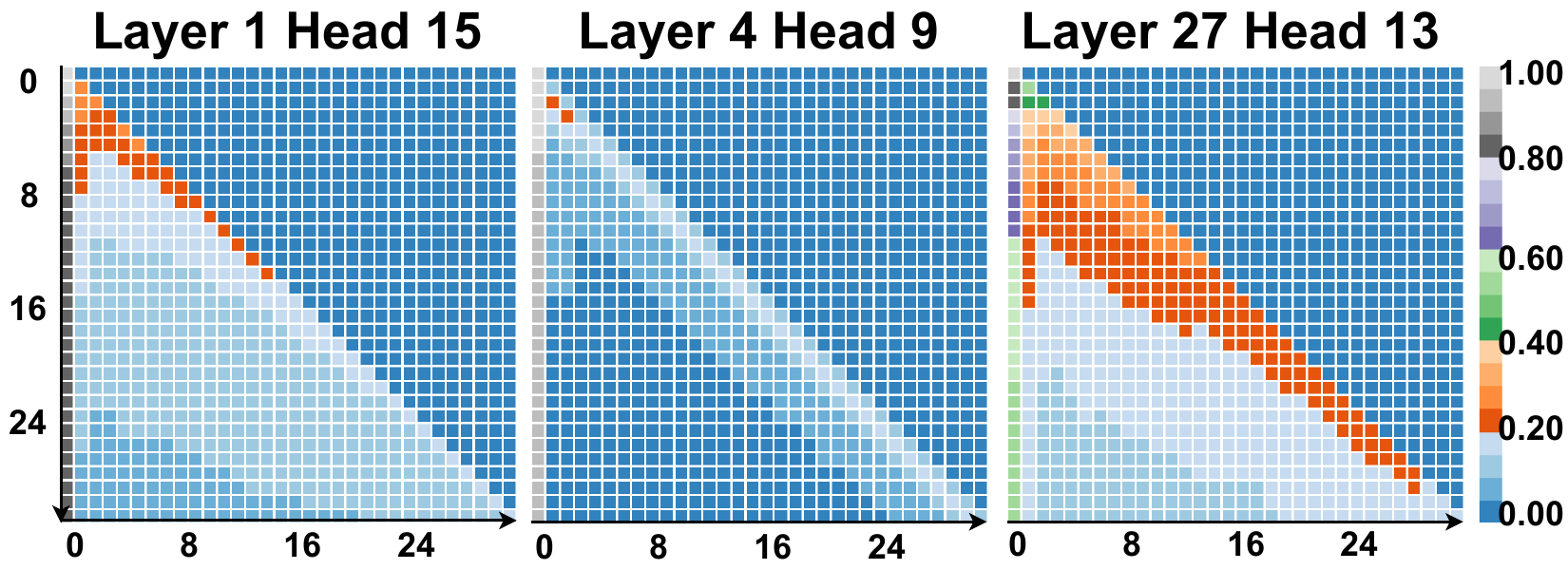}
    \caption{Visualization of attention layouts across layers and heads. The figure compares the attention values from a LLaMA 3.2 3B Instruct model on the Multi-News dataset.}
    \vspace{-20pt}
    \label{fig:attn_layouts}
\end{figure}

Figure~\ref{fig:attn_layouts} visualizes attention maps across layers and heads for LLaMA 3.2 3B on the Multi-News dataset.
The maps provide qualitative motivation for structured tile-group precision allocation by showing that attention values are distributed non-uniformly across the query--key plane.
The evaluated routing templates are statically constructed and do not use benchmark outputs for policy selection.
In \modelname, these layouts do not remove interactions; they only determine whether each legal tile group is routed to FP16 or INT8.
These observations motivate a precision-routing view of attention acceleration.

\modelname supports block-level routing layouts that specify whether each legal tile group is computed through the FP16 or INT8 score path.
These layouts reuse spatial structures studied in prior sparse-attention work as precision-routing templates.
The resulting policies preserve dense token connectivity while assigning selected spatial regions to FP16 and the remaining legal regions to INT8.

\noindent\textbf{Policy Construction and Sharing.}
For each sequence length and block geometry, a layout is instantiated as a two-dimensional routing template
\(
\overline{R}_{m,g_j}\in\{0,1\},
\)
where $m$ is the query-tile row and $g_j$ is the key-tile-group index.
The convention is
{\small
\[
\overline{R}_{m,g_j}
=
\begin{cases}
1, & \text{INT8 score path},\\
0, & \text{FP16 score path}.
\end{cases}
\]
}
Causal and sequence-boundary masks determine whether token interactions are legal independently of this precision decision.

The same template is reused across transformer layers and batch examples and is broadcast across KV heads:
\[
R_{h_k,m,g_j}
=
\overline{R}_{m,g_j},
\qquad
h_k\in\{0,\ldots,H_k-1\}.
\]
Query heads mapped to the same KV head and read the same packed routing word. The layer and batch indices are omitted from $R$ because the evaluated routing values are shared along these dimensions.

To keep one routing word sufficient at long sequence lengths, the routing-group width is selected as a multiple of the key compute-tile width:
{\small
\(
b_{64}=\left\lceil\frac{L_k}{64}\right\rceil,\;
\mathrm{BLOCK}_N^{\mathrm{mask}}
=\mathrm{BLOCK}_N\left\lceil
\frac{\max\!\left(b_{64},\mathrm{BLOCK}_N\right)}
{\mathrm{BLOCK}_N}
\right\rceil.
\)
}
Consequently,
\(
g
=
\frac{
\mathrm{BLOCK}_N^{\mathrm{mask}}
}{
\mathrm{BLOCK}_N
},
\;
T_{\mathrm{mask}}
=
\left\lceil
\frac{L_k}{\mathrm{BLOCK}_N^{\mathrm{mask}}}
\right\rceil
\le 64,
\)
and key compute tile $n$ uses routing-group index
\(
g_j=\left\lfloor\frac{n}{g}\right\rfloor.
\)

For each query-tile row, the decisions are packed into
\(
b_{h_k,m}
=
\sum_{g_j=0}^{T_{\mathrm{mask}}-1}
R_{h_k,m,g_j}\,2^{g_j}.
\)
The template is regenerated when the sequence length or selected block geometry changes and is cached otherwise.

\noindent\textbf{Coverage Accounting.}
Let
\(
\mathcal{G}
=
\left\{
(h_k,m,g_j):
\mathcal{I}_m\times\mathcal{J}_{g_j}
\right\}
\)
denote the set of legal routing groups.
The realized INT8 coverage is
\[
\widehat{\rho}_{\mathrm{INT8}}
=
\frac{
\sum_{(h_k,m,g_j)\in\mathcal{G}}
R_{h_k,m,g_j}
}{
|\mathcal{G}|
}.
\]
The reported 25\%, 50\%, and 75\% settings denote legal score-tile-group INT8 coverage.
They do not denote the fraction of total attention FLOPs executed in INT8.
For each layout, its parameters are selected at tile-group granularity to match the requested coverage as closely as possible.
Minor differences between requested and realized coverage may arise from discrete routing groups and sequence boundaries.

For query-tile row $m$, let
\(
\mathcal{G}_m
=
\left\{
g_j:
(h_k,m,g_j)\in\mathcal{G}
\right\}
\)
denote its legal key-tile groups, and let
\(
\mathcal{F}_m
=
\left\{
g_j\in\mathcal{G}_m:
R_{h_k,m,g_j}=0
\right\}
\)
denote its FP16-routed groups.
The layouts below differ in the spatial construction of $\mathcal{F}_m$.

\noindent\textbf{One (All INT8).}
All legal tile groups are routed to INT8:
\(
\mathcal{F}_m^{\mathrm{One}}=\varnothing,
\;
R_{h_k,m,g_j}=1.
\)
This pattern corresponds to uniform INT8 score computation and serves as the matched low-precision reference using the same \modelname execution substrate.

\noindent\textbf{Zero (All FP16).}
All legal tile groups are routed to FP16:
\(
\mathcal{F}_m^{\mathrm{Zero}}=\mathcal{G}_m,
\;
R_{h_k,m,g_j}=0.
\)
This pattern corresponds to all-FP16 of the \modelname kernel and serves as its full-precision reference.

\noindent\textbf{Band (Local Template).}~\cite{beltagy2020longformer, dai2019transformer}
Band retains a contiguous FP16 region around the query--key diagonal.
Let
\(
\gamma(m)
=
\left\lfloor
\frac{m\,\mathrm{BLOCK}_M}
{\mathrm{BLOCK}_N^{\mathrm{mask}}}
\right\rfloor
\)
denote the key-group index aligned with the beginning of query-tile row $m$.
For a half-width $w$, the FP16 set is
\(
\mathcal{F}_m^{\mathrm{Band}}(w)
=
\left\{
g_j\in\mathcal{G}_m:
|g_j-\gamma(m)|\le w
\right\}.
\)
Groups inside the band are routed to FP16, while the remaining legal groups are routed to INT8.
The width $w$ is selected to match the requested INT8 coverage at tile-group granularity.
This template transfers the locality bias of sliding-window attention into precision allocation while preserving the complete attention.

\noindent\textbf{Global.}~\cite{zaheer2021bigbirdtransformerslonger, beltagy2020longformer}
Let
\(
\mathcal{G}_{\mathrm{global}}
\subseteq
\{0,\ldots,T_{\mathrm{mask}}-1\}
\)
denote a fixed set of key-tile groups associated with designated global positions.
Global uses
\(
\mathcal{F}_m^{\mathrm{Global}}
=
\mathcal{G}_{\mathrm{global}}
\cap
\mathcal{G}_m.
\)
The selected global groups remain in FP16 for every query-tile row, while the remaining legal groups are routed to INT8.
The number of global groups is selected to match the requested INT8 coverage.
This template follows the global-token bias used in BigBird- and Longformer-style mechanisms.

\noindent\textbf{Row-Random.}~\cite{zaheer2021bigbirdtransformerslonger}
Each query-tile row independently selects a fixed-size subset of legal key-tile groups uniformly at random for FP16 routing:
\(
\mathcal{F}_m^{\mathrm{RowRand}}
\sim
\operatorname{UniformSubset}
\left(
\mathcal{G}_m,
k_m
\right),
\)
where
\(
k_m
=
\operatorname{round}
\left(
(1-\rho_{\mathrm{INT8}})
|\mathcal{G}_m|
\right).
\)
Sampling is performed without replacement using a fixed random seed, and the resulting routing template is reused throughout evaluation.
The remaining groups are routed to INT8.
This template follows the random component of BigBird and distributes FP16 across nearby and long-range positions.

\noindent\textbf{Aligned Sparse.}~\cite{dai2019transformerxlattentivelanguagemodels, press2021train}
Aligned Sparse assigns a right-aligned region of legal key-tile groups to FP16.
Let
\(
a(m)
=
\min
\left(
|\mathcal{G}_m|,
\left\lceil
\alpha+\beta m
\right\rceil
\right)
\)
denote the number of FP16 groups assigned to query-tile row $m$, where $\alpha$ controls the initial width and $\beta$ controls its growth.
If
\(
g_m^{\max}
=
\max \mathcal{G}_m,
\)
then
\(
\mathcal{F}_m^{\mathrm{Align}}
=
\left\{
g_j\in\mathcal{G}_m:
g_j\ge g_m^{\max}-a(m)+1
\right\}.
\)
The number of FP16-routed groups therefore increases monotonically with the query index, producing a right-aligned template with an expanding FP16 region.
The parameters $\alpha$ and $\beta$ are selected to match the requested INT8 coverage.

\noindent\textbf{BigBird.}~\cite{zaheer2021bigbirdtransformerslonger}
This template combines
(i) local-band routing,
(ii) global-token routing, and
(iii) row-wise random routing.
Its FP16 set is
\(
\mathcal{F}_m^{\mathrm{BigBird}}
=
\mathcal{F}_m^{\mathrm{Band}}
\cup
\mathcal{F}_m^{\mathrm{Global}}
\cup
\mathcal{F}_m^{\mathrm{RowRand}}.
\)
A legal tile group selected by any component remains on the FP16 path, while the remaining legal groups use INT8.
The local, global, and random component sizes are selected jointly to match the requested tile-group INT8 coverage.
\modelname therefore interprets the corresponding spatial topology as a composite precision-routing template rather than a structural attention mask.

\noindent\textbf{Sparse Transformer (SpTrans).}~\cite{child2019generatinglongsequencessparse}
SpTrans combines
(i) the local stride containing the current query position and
(ii) the final $c$ positions of each preceding stride.
For stride length $s$ and query position $i$, let
\(
u(i)
=
\left\lfloor
\frac{i}{s}
\right\rfloor
\)
denote its current stride.
The selected FP16 key positions are
\(
\mathcal{K}_i^{\mathrm{SpTrans}}
=
\left[
u(i)s,\,
i
\right]
\cup
\bigcup_{r=0}^{u(i)-1}
\left[
\max\!\left(rs,(r+1)s-c\right),\,
(r+1)s-1
\right].
\)
For query-tile row $m$, the corresponding FP16 routing set is
\(
\mathcal{F}_m^{\mathrm{SpTrans}}
=
\left\{
g_j\in\mathcal{G}_m:
\exists i\in\mathcal{I}_m
\text{ such that }
\mathcal{J}_{g_j}
\cap
\mathcal{K}_i^{\mathrm{SpTrans}}
\neq\varnothing
\right\}.
\)
These selected tile groups are routed to FP16, while the remaining legal tile groups are routed to INT8.
The stride length $s$ and tail width $c$ determine the spatial structure, and the resulting tile-group allocation is configured for the requested INT8 coverage.

\noindent\textbf{Layouts Reported in the Main Evaluation.}
The main LV-Eval table reports BigBird and SpTrans as representative structured layouts with complementary spatial organizations and empirical quality--efficiency behavior.
Complete results for Band, Global, Row-Random, Aligned Sparse, BigBird, and SpTrans are reported in this appendix across models, context lengths, and INT8 coverage levels.

The supported routing layouts do not introduce sparsity into the attention computation.
Instead, \modelname translates established spatial structures into hardware-aligned, block-level precision-routing decisions.
This decouples \emph{which interactions are legal} from \emph{which arithmetic path computes them}, enabling training-free heterogeneous-precision execution with dense token connectivity.

\section{Quantization Configuration}
\label{app:quantization-configuration}

This section specifies the numerical configuration of the primary score-routing implementation.
Blockwise INT8 quantization is applied to $Q$ and $K$, while $V$ and the $PV$ computation remain in FP16.
Both score paths enter one common floating-point score domain before updating the shared online-softmax state.

\noindent\textbf{Blockwise INT8 Quantization.}
For a valid floating-point block $X^{(r)}$, \modelname uses signed symmetric INT8 quantization with absmax scale
\(
\delta_X^{(r)}
=
\frac{
\max_{x\in X^{(r)}} |x|
}{
127
}.
\)
The quantized representation is obtained by nearest-integer conversion:
\(
\widehat{X}^{(r)}
=
\operatorname{round}
\left(
\frac{X^{(r)}}{\delta_X^{(r)}}
\right),
\;
\widehat{X}^{(r)}\in\{-127,\ldots,127\}.
\)
Because the scale is determined by the block maximum, finite normalized values lie within the stated signed range.

The quantization kernel converts block values to FP32 when computing the scale and quantized representation.
The resulting scale tensors are stored in the operand dtype by the production wrapper; the FP16 experiments therefore use FP16 scale tensors.

\noindent\textbf{Quantization Granularity.}
The evaluated score-routing path uses
\(
\mathrm{BLK}_Q=128,
\qquad
\mathrm{BLK}_K=64.
\)
Each query block has shape
\(
128\times d,
\)
and each key block has shape
\(
64\times d,
\)
with one scale per block and attention head.

For variable-length batches, the scale tensors have logical shapes
\(
\left[
\sum_b
\left\lceil
\frac{L_{q,b}}{128}
\right\rceil,
H_q
\right]
\quad\text{and}\quad
\left[
\sum_b
\left\lceil
\frac{L_{k,b}}{64}
\right\rceil,
H_k
\right].
\)
Prefix-sum metadata maps each sequence and token block to its scale entry.

\noindent\textbf{INT8 Score Computation.}
For query block $r$ and key block $s$, an INT8-routed score tile first computes
\(
C_{r,s}^{\mathrm{INT32}}
=
Q_8^{(r)}
\left(K_8^{(s)}\right)^\top
\)
using INT8 Tensor Core multiplication with INT32 accumulation.
The accumulator is converted to floating point and rescaled as
\(
S_{r,s}^{\mathrm{INT8}}
=
\frac{
\delta_Q^{(r)}
\delta_K^{(s)}
C_{r,s}^{\mathrm{INT32}}
}{
\sqrt{d}
}.
\)
An FP16-routed tile computes
\(
S_{r,s}^{\mathrm{FP16}}
=
\frac{
Q^{(r)}
\left(K^{(s)}\right)^\top
}{
\sqrt{d}
}.
\)
Both paths enter FP16 score domain before exponentiation and update shared online-softmax state.

The running maximum, normalizer, and output accumulator are stored in FP16.
Tile maxima, exponentiation, and row reductions use FP32 intermediates before their results are incorporated into the shared FP16 state.
The resulting probability tile and the $PV$ update use FP16.

\noindent\textbf{Quantize-Once Execution.}
The implementation-level prefill path generates $Q_8$, $K_8$, and their compact block scales once before the attention launch.
These operands are reused by all INT8-routed score-tile groups, while FP16-routed groups read the original FP16 $Q$ and $K$ tensors.
The all-FP16 configuration skips this quantization step.
Reported implementation-level timings include operand quantization, scale preparation, routing, rescaling, memory staging, and fused attention.

\noindent\textbf{Additional Execution Interfaces.}
The implementation also provides an on-the-fly operand-generation mode and an INT8 key/value cache interface.
The cache interface stores $K_8$ and $V_8$ together with compact per-block scales.
During decode, cached $K_8$ participates in the routed INT8 score path; cached $V_8$ is converted to FP16 on chip, with its block scale folded into the probability tile before the $PV$ dot product.
This interface is separate from the primary prefill configuration, which quantizes the routed $Q/K$ score operands and retains FP16 $V$ for every $PV$ update.

\section{Storage and Execution Layout}
\label{app:storage_layout}

Building on the numerical configuration in Appendix~\ref{app:quantization-configuration}, \modelname keeps the dense FP16 $Q,K,V$ tensors in the standard contiguous HBM layout and stores routing metadata separately as one packed 64-bit word for each $(h_k,m)$, where $h_k$ is a KV-head index and $m$ is a query-tile row.
The fused inner loop reads this word to select the FP16 or INT8 score path for each legal tile group.
Tensor addressing, token connectivity, and tile legality retain the dense FlashAttention-style execution structure.

\begin{figure*}[t]
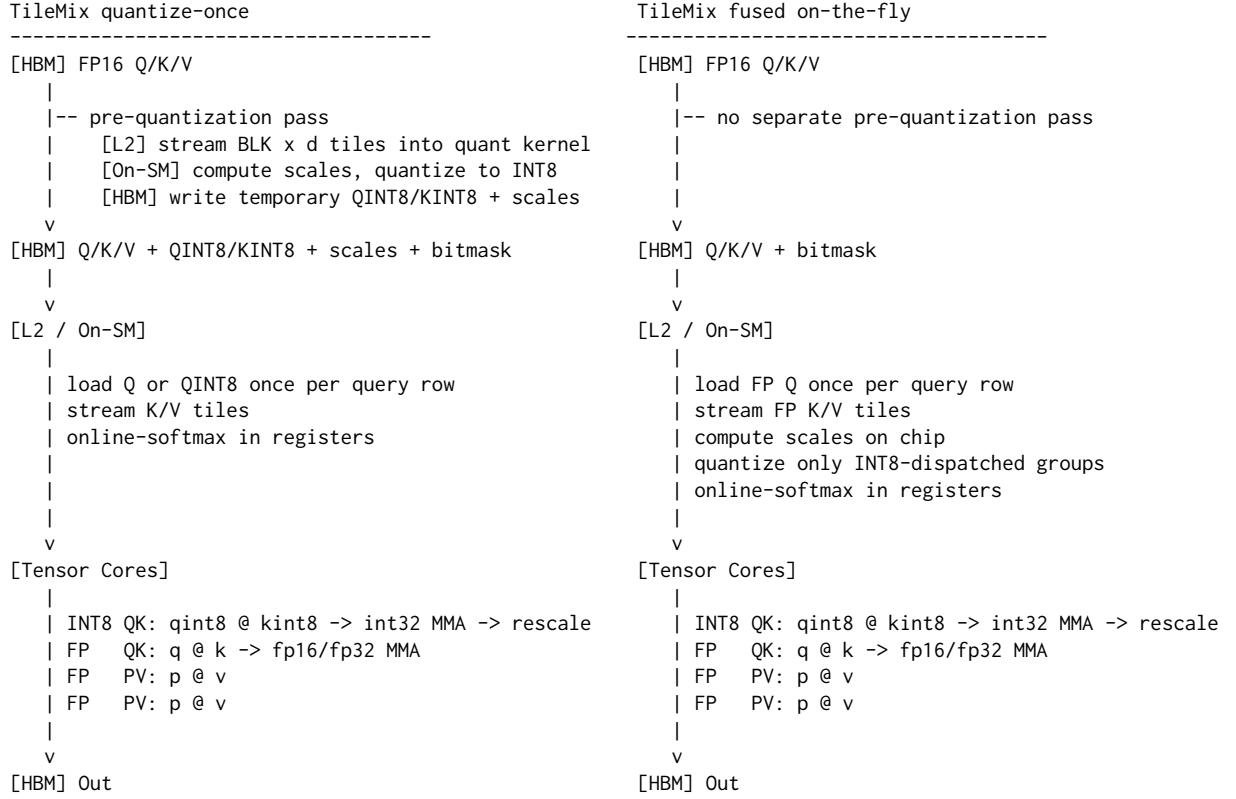

\centering

\begin{adjustbox}{width=\textwidth,center}
\begin{BVerbatim}[fontsize=\tiny]
TileMix quantize-once                                  TileMix fused on-the-fly
-------------------------------------                 -------------------------------------
[HBM] FP16 Q/K/V                                       [HBM] FP16 Q/K/V
   |                                                      |
   |-- pre-quantization pass                              |-- no separate pre-quantization pass
   |    [L2] stream BLK x d tiles into quant kernel       |
   |    [On-SM] compute scales, quantize to INT8          |
   |    [HBM] write temporary QINT8/KINT8 + scales        |
   v                                                      v
[HBM] Q/K/V + QINT8/KINT8 + scales + bitmask           [HBM] Q/K/V + bitmask
   |                                                      |
   v                                                      v
[L2 / On-SM]                                           [L2 / On-SM]
   |                                                      |
   | load Q or QINT8 once per query row                   | load FP Q once per query row
   | stream K/V tiles                                     | stream FP K/V tiles
   | online-softmax in registers                          | compute scales on chip
   |                                                      | quantize only INT8-dispatched groups
   |                                                      | online-softmax in registers
   |                                                      |
   v                                                      v
[Tensor Cores]                                         [Tensor Cores]
   |                                                      |
   | INT8 QK: qint8 @ kint8 -> int32 MMA -> rescale       | INT8 QK: qint8 @ kint8 -> int32 MMA -> rescale
   | FP   QK: q @ k -> fp16/fp32 MMA                      | FP   QK: q @ k -> fp16/fp32 MMA
   | FP   PV: p @ v                                       | FP   PV: p @ v
   | FP   PV: p @ v                                       | FP   PV: p @ v
   |                                                      |
   v                                                      v
[HBM] Out                                              [HBM] Out
\end{BVerbatim}
\end{adjustbox}

\caption{
Storage and execution layout of the two \modelname operand-preparation modes
for the primary prefill path.
Quantize-once materializes temporary INT8 $Q/K$ tensors and their scales once
per attention call and reuses them across INT8-routed score-tile groups.
Fused on-the-fly generates INT8 $Q/K$ fragments inside the kernel only for
INT8-routed groups.
Both modes retain $V$ and the $PV$ computation in FP16.
The separate INT8 key/value cache interface is described in the surrounding text.
}
\label{fig:tilemix_storage_layout}
\end{figure*}

\modelname supports two operand-preparation modes, as illustrated in Figure~\ref{fig:tilemix_storage_layout}.
The \textit{quantize-once} mode is used by the reported quality and implementation-level efficiency evaluations.
It generates temporary INT8 $Q/K$ tensors and their block scales once per attention call and reuses them across all INT8-routed groups.
The original FP16 $Q/K$ tensors remain available to FP16-routed groups, while $V$ and $PV$ computation remain FP16 throughout the primary evaluated path.

The \textit{fused on-the-fly} mode is an additional implementation mode that loads the FP16 operands and generates INT8 fragments only for groups selected by the routing map.
This mode removes the attention-call-local $Q_8/K_8$ scratch tensors from HBM and performs scale computation and quantization on-chip.
Both modes use the same packed routing policy, routed score computation, and shared online-softmax update.

The kernel interface additionally supports INT8 key/value caches with per-block scale metadata.
In this cache path, cached $K_8$ supplies the routed INT8 score computation, while cached $V_8$ is converted to FP16 on chip and used in the floating-point $PV$ update.
This cache interface is described separately from the primary prefill path illustrated in Figure~\ref{fig:tilemix_storage_layout}.
The primary evaluation uses routed FP16/INT8 $QK^\top$ score computation with FP16 $V/PV$, as specified in Appendix~\ref{app:quantization-configuration}.

For clarity, Table~\ref{tab:tilemix_memory_accounting} reports attention-call-local memory and data-movement accounting under one representative configuration:
$H_q=32$, $H_k=8$, $L_q=L_k=8192$, $d=128$, $\mathrm{BLK}_Q=128$, and $\mathrm{BLK}_K=64$.
Here, $H_q$ and $H_k$ are the query-head and KV-head counts, $L_q$ and $L_k$ are the query and key sequence lengths, $d$ is the per-head dimension, and $\mathrm{BLK}_Q$ and $\mathrm{BLK}_K$ are the query and key quantization-block sizes.

The accounting distinguishes:
(i) tensors resident in HBM during one attention call;
(ii) global/L2 operand reads for one routed QK or optional PV tile event; and
(iii) the simultaneously live on-SM working state for one FP16 tile event.
The INT8 QK column corresponds to the primary score-routing path, while the $V_8$-backed PV column separately characterizes value-cache traffic in the INT8 key/value cache interface.
Resident HBM for the \modelname quantize-once row reports the primary score-only prefill configuration.
These quantities describe operator-local storage and traffic, distinct from full-model peak GPU memory.

\begin{table*}[t]
\centering
\small
\setlength{\tabcolsep}{4pt}
\renewcommand{\arraystretch}{1.12}

\begin{adjustbox}{width=\textwidth}
\begin{tabular}{lcccccc}
\toprule
\textbf{Route} &
\makecell{\textbf{Resident}\\\textbf{HBM}} &
\makecell{\textbf{INT8 QK tile}\\\textbf{global/L2 read}} &
\makecell{\textbf{$V_8$-backed PV tile}\\\textbf{global/L2 read}} &
\makecell{\textbf{FP QK tile}\\\textbf{on-SM live set}} &
\makecell{\textbf{FP PV tile}\\\textbf{on-SM live set}} &
\makecell{\textbf{FP QK/PV}\\\textbf{Tensor Core accumulator}} \\
\midrule

FlashAttention &
96 MB &
\textemdash &
\textemdash &
128 KB &
128 KB &
64 KB \\

\modelname quantize-once &
136.01 MiB &
16.25 KB &
16.25 KB &
96 KB &
96 KB &
32 KB \\

\modelname fused on-the-fly &
96 MB &
32 KB &
32 KB &
96 KB &
96 KB &
32 KB \\

\bottomrule
\end{tabular}
\end{adjustbox}
\vspace{-5pt}
\caption{
Memory and data-movement accounting for FlashAttention and \modelname under
$H_q=32$, $H_k=8$, $L_q=L_k=8192$, $d=128$, $\mathrm{BLK}_Q=128$, and $\mathrm{BLK}_K=64$.
Resident HBM counts persistent tensors and route-specific temporary scratch tensors during the attention call.
Global/L2 reads are reported for one INT8-routed QK or PV tile event.
The on-SM live set reports the simultaneously live on-chip working state for the corresponding FP16-routed tile event.
}
\label{tab:tilemix_memory_accounting}
\end{table*}

Under this representative configuration, quantize-once uses 136.01 MiB of resident HBM because it materializes reusable INT8 scratch tensors and their scales for the duration of the attention call.
This preparation reduces the operand read for one INT8-routed QK tile event from 32 KB in fused on-the-fly execution to 16.25 KB.
The same traffic relation applies to $V_8$-backed PV reads in the optional INT8 key/value cache interface.

Fused on-the-fly retains the same 96 MB resident-HBM footprint as the FlashAttention configuration in this accounting and generates INT8 operands from FP16 tiles on chip.
For FP16-routed QK and PV events, the \modelname tile path uses a 96 KB on-SM live set and a 32 KB Tensor Core accumulator, compared with 128 KB and 64 KB, respectively, under the reported FlashAttention accounting. These measurements characterize the local storage and data-movement behavior of the attention implementations; full-model OOM behavior is considered separately from this operator-level accounting.

\section{Model Performance}\label{app:model-performance}

\textbf{Long-context Retrieval.}

\begin{figure*}[t]
    \centering

    \begin{subfigure}[b]{0.485\linewidth}
        \centering
        \includegraphics[width=\linewidth]{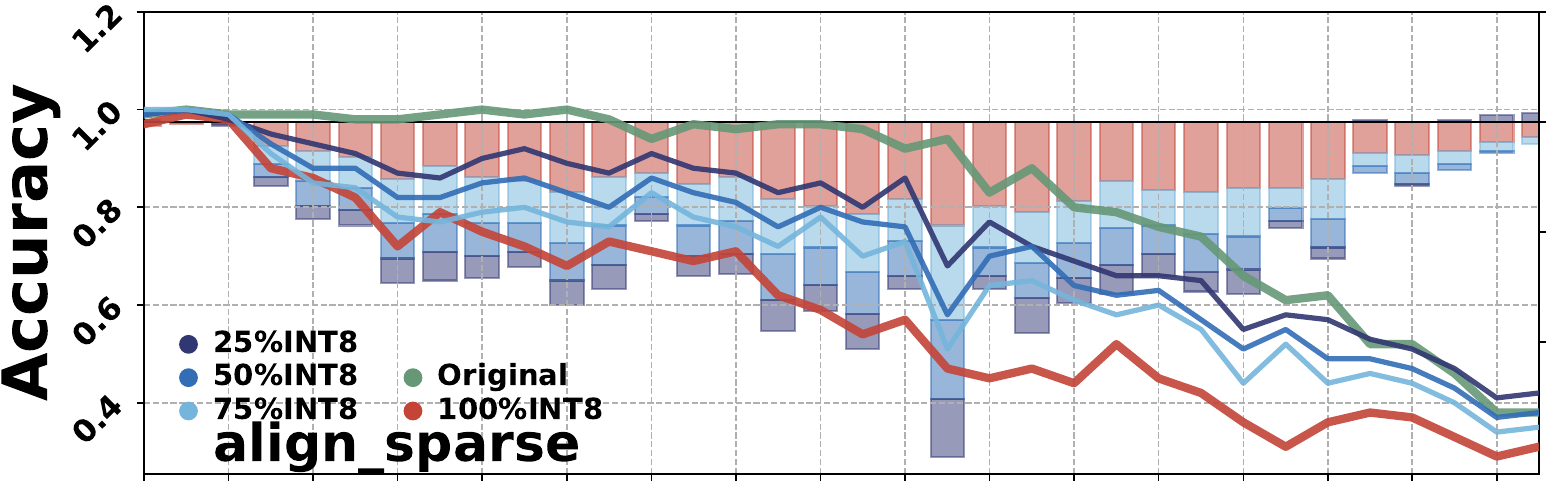}
    \end{subfigure}
    \hfill
    \begin{subfigure}[b]{0.487\linewidth}
        \centering
        \includegraphics[width=\linewidth]{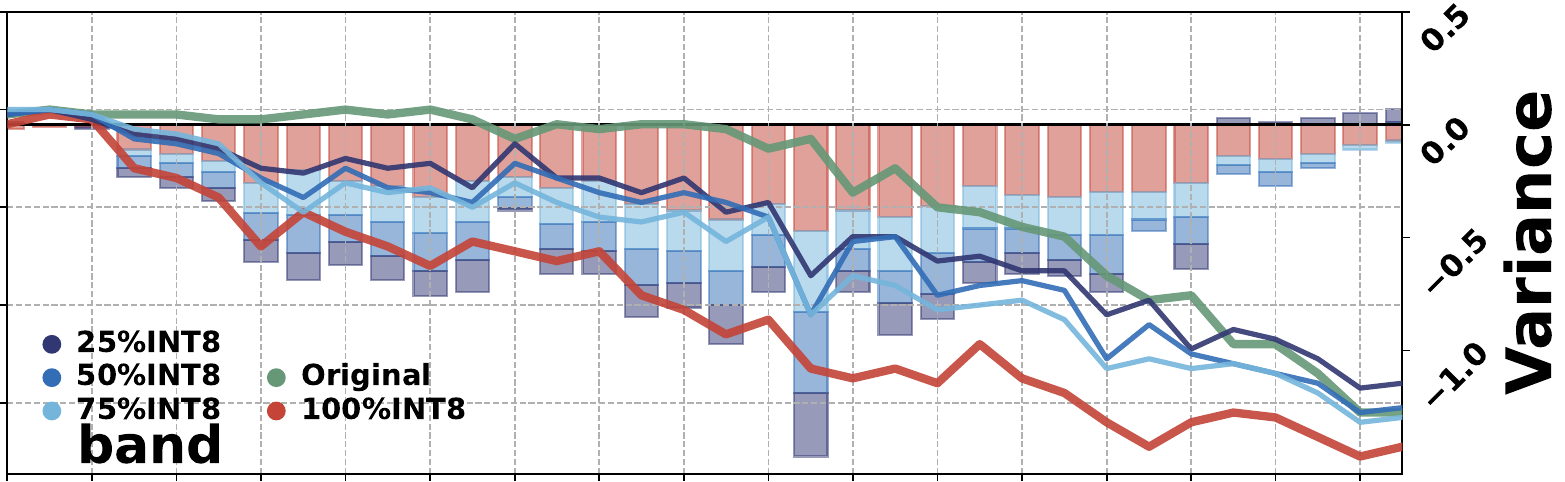}
    \end{subfigure}

    \begin{subfigure}[b]{0.485\linewidth}
        \centering
        \includegraphics[width=\linewidth]{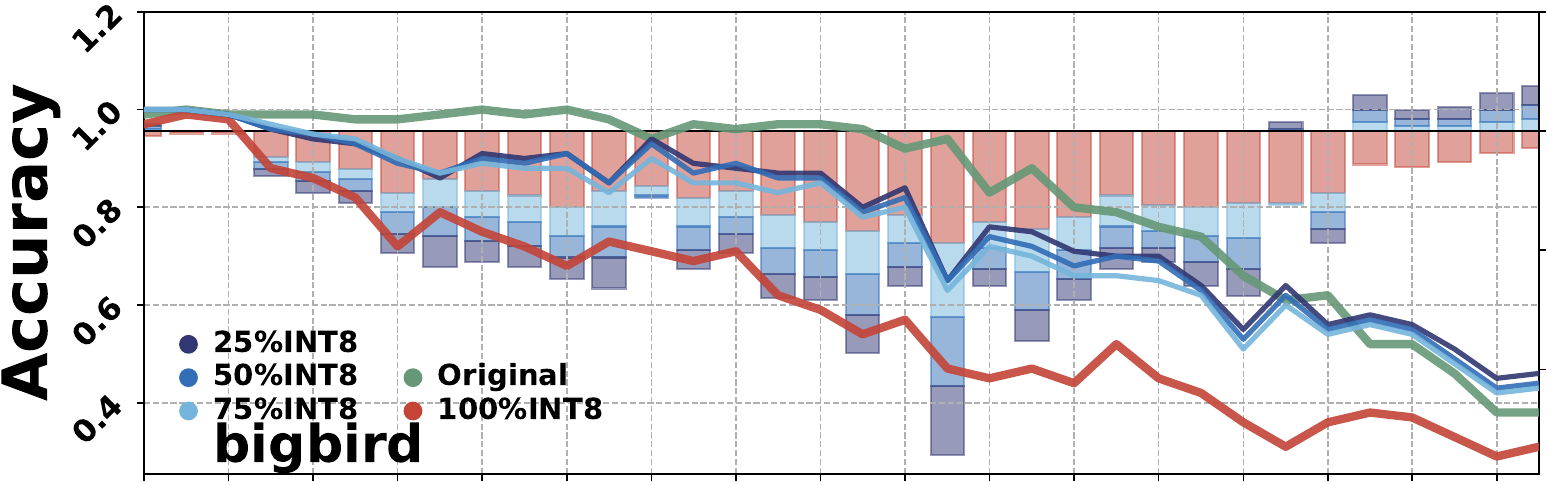}
    \end{subfigure}
    \hfill
    \begin{subfigure}[b]{0.487\linewidth}
        \centering
        \includegraphics[width=\linewidth]{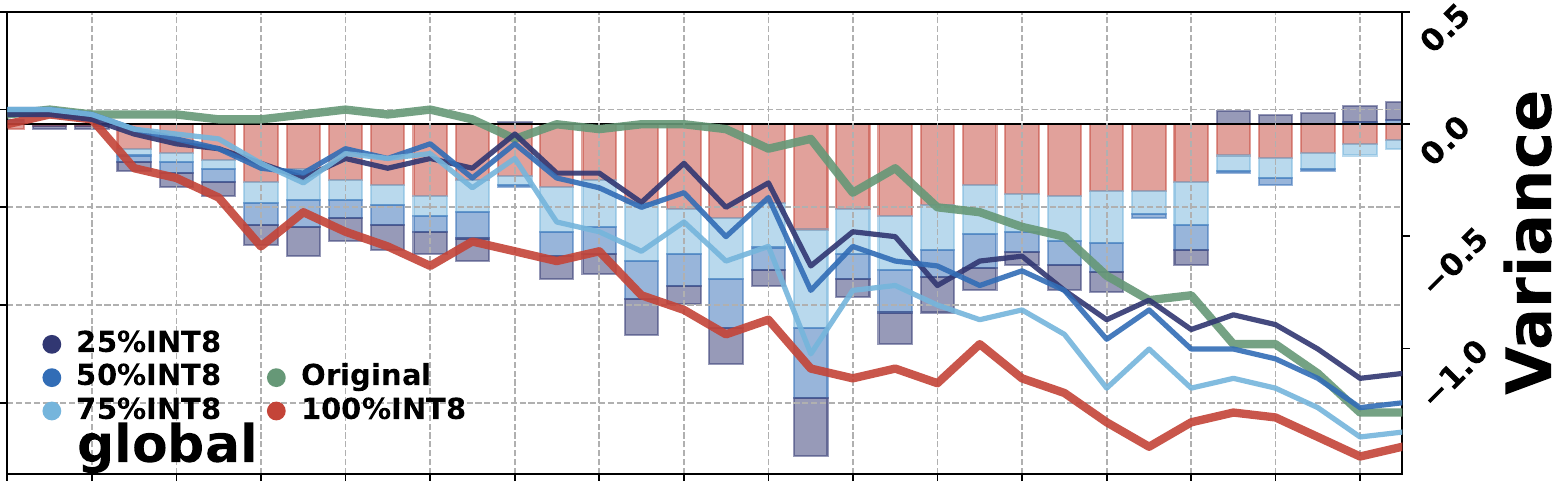}
    \end{subfigure}

    \begin{subfigure}[b]{0.485\linewidth}
        \centering
        \includegraphics[width=\linewidth]{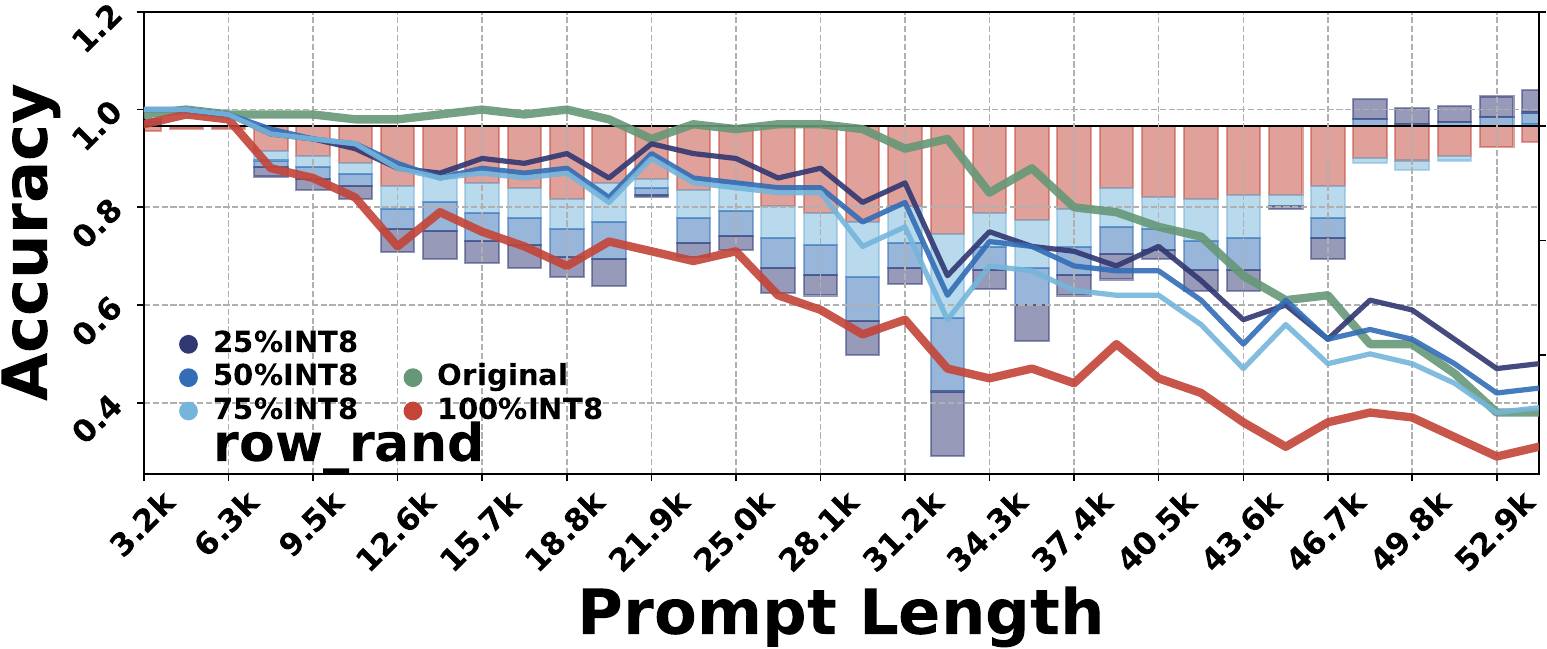}
    \end{subfigure}
    \hfill
    \begin{subfigure}[b]{0.497\linewidth}
        \centering
        \includegraphics[width=\linewidth]{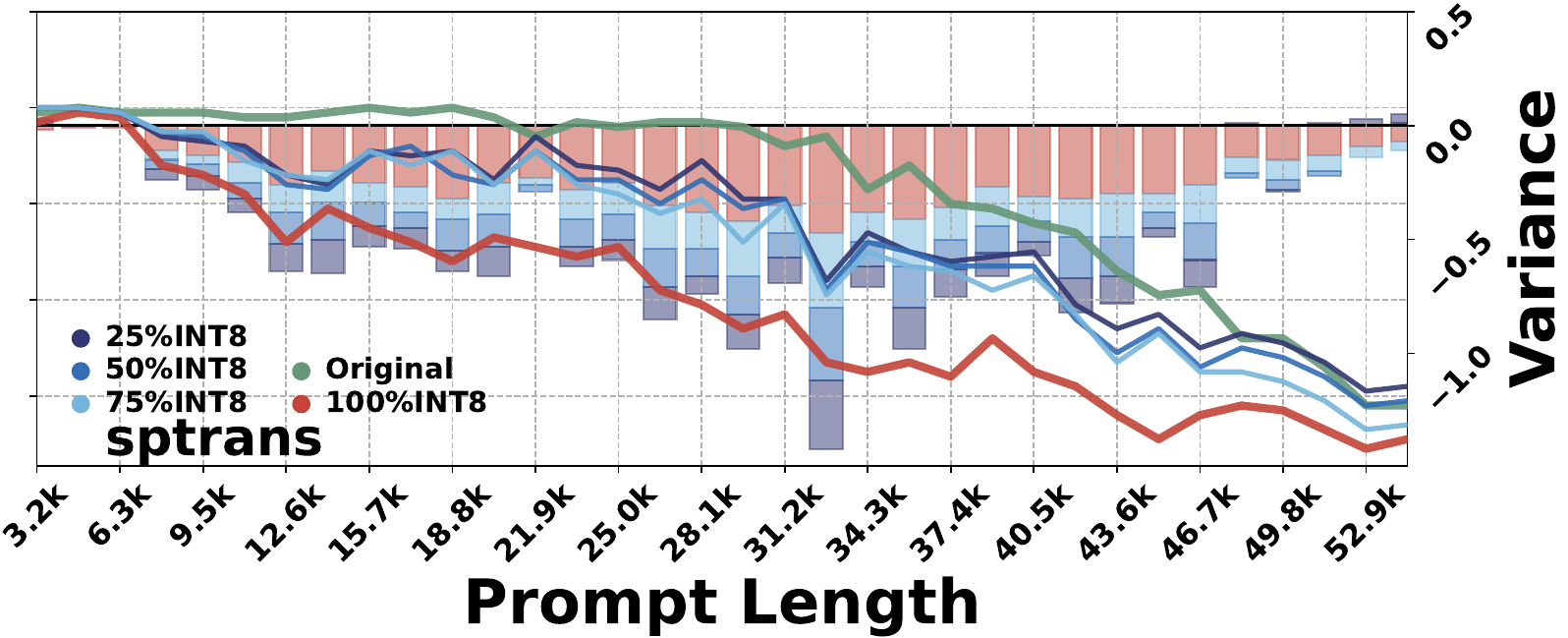}
    \end{subfigure}
     \vspace{-10pt}
    \caption{
    Line-level retrieval accuracy on the LongEval benchmark for \textbf{Qwen 2.5 7B} under different tile-group routing layouts, evaluated across prompt lengths from 3.1k to 38.7k tokens.
    Line plots (left y-axis) report exact-match retrieval accuracy, while bar plots (right y-axis, $\Delta$ accuracy vs.\ FP16) show differences relative to the FP16 attention baseline.
    Each panel corresponds to a routing layout with 25\%, 50\%, or 75\% of tile groups routed to INT8.
    Colors indicate routing settings: \textcolor{baselinegreen}{green} denotes the original FP16 attention baseline, \textcolor{onered}{red} denotes \textbf{One} (all legal score-tile groups routed to INT8), and \textcolor{darkblue}{blue} denotes mixed configurations with partially INT8-routed tile groups.
    }
    \label{fig:longeval_qwen2.5}
\end{figure*}

\begin{figure*}[t]
    \centering

    \begin{subfigure}[b]{0.485\linewidth}
        \centering
        \includegraphics[width=\linewidth]{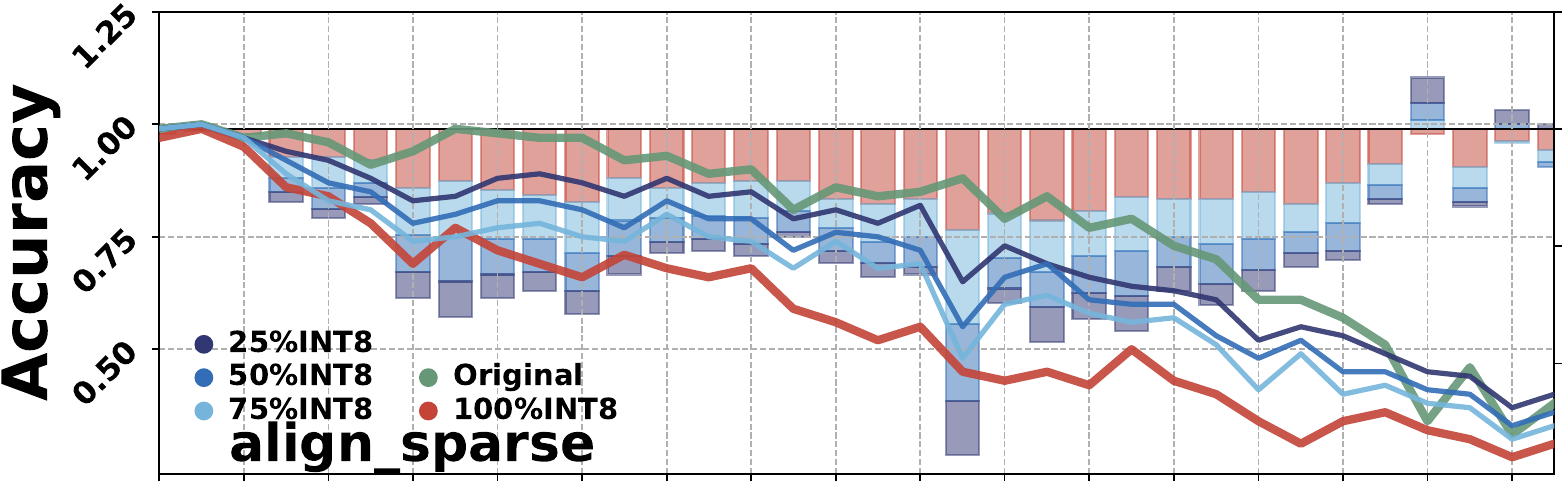}
    \end{subfigure}
    \hfill
    \begin{subfigure}[b]{0.487\linewidth}
        \centering
        \includegraphics[width=\linewidth]{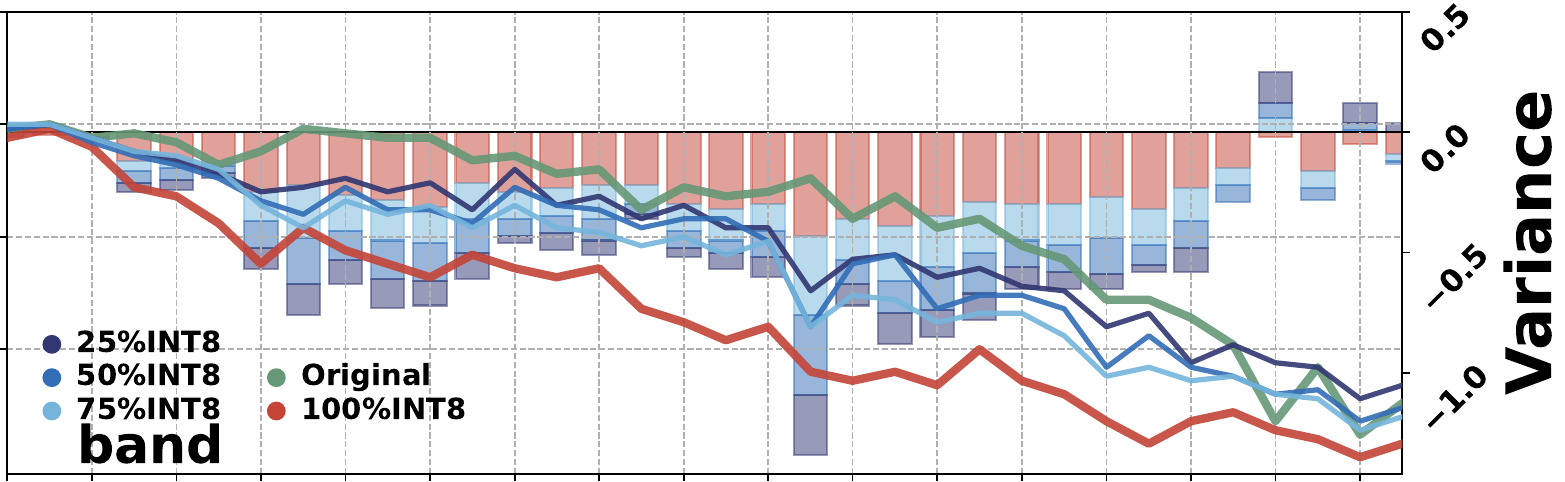}
    \end{subfigure}

    \begin{subfigure}[b]{0.485\linewidth}
        \centering
        \includegraphics[width=\linewidth]{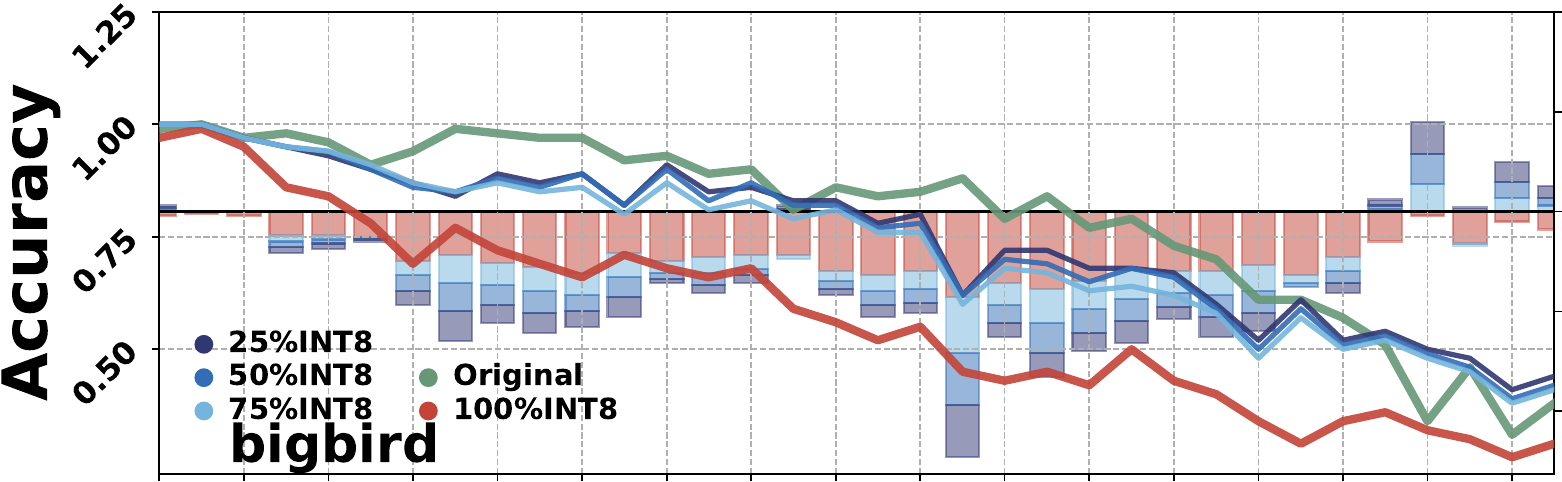}
    \end{subfigure}
    \hfill
    \begin{subfigure}[b]{0.487\linewidth}
        \centering
        \includegraphics[width=\linewidth]{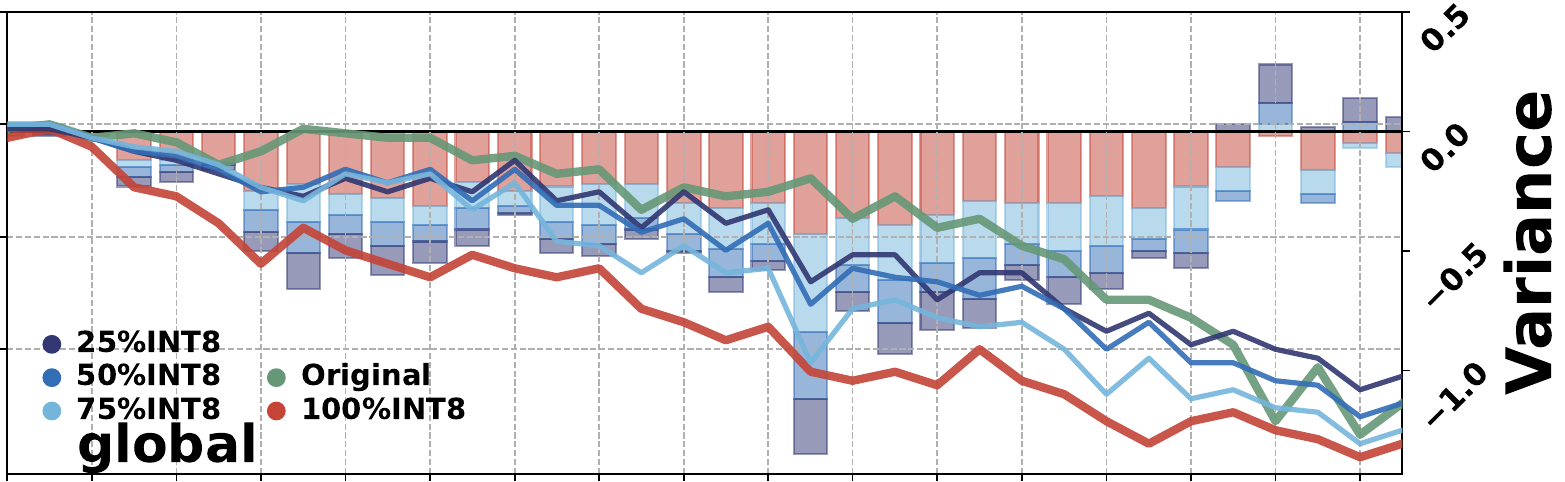}
    \end{subfigure}

    \begin{subfigure}[b]{0.485\linewidth}
        \centering
        \includegraphics[width=\linewidth]{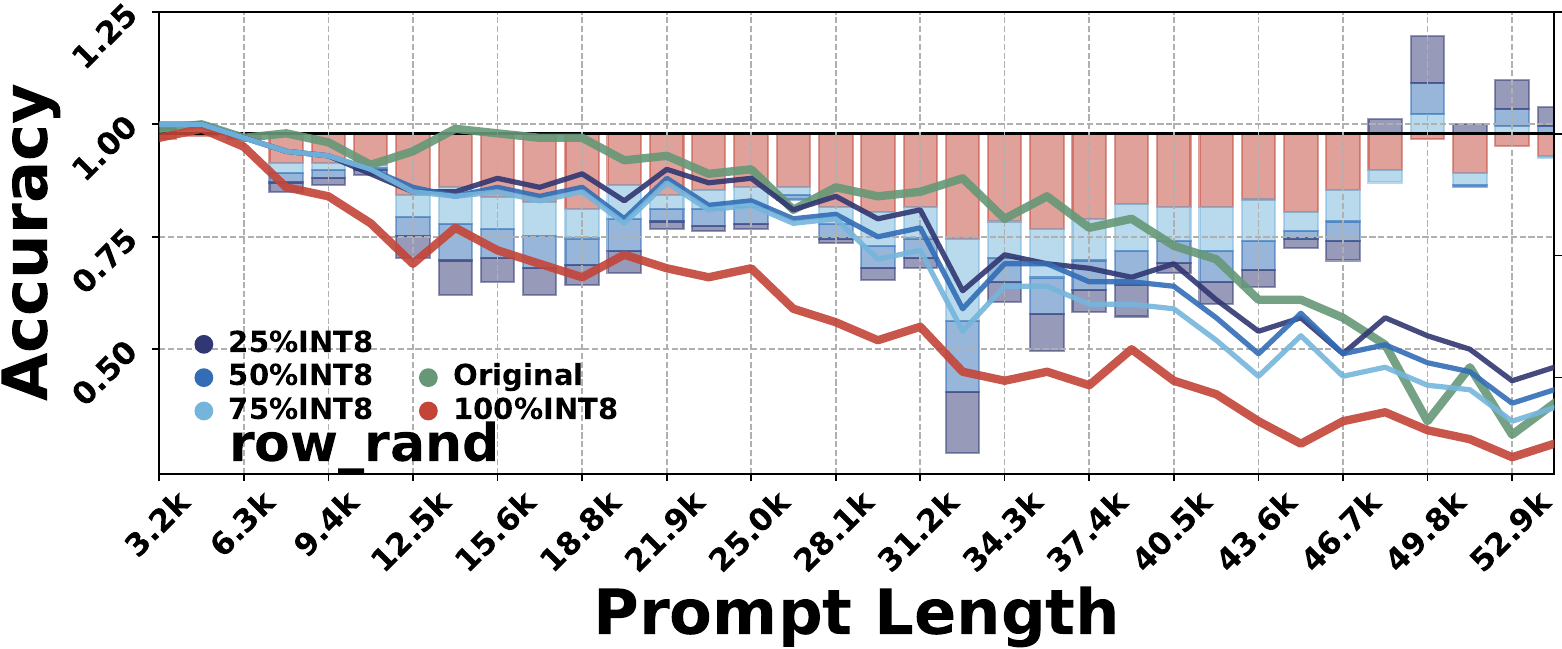}
    \end{subfigure}
    \hfill
    \begin{subfigure}[b]{0.497\linewidth}
        \centering
        \includegraphics[width=\linewidth]{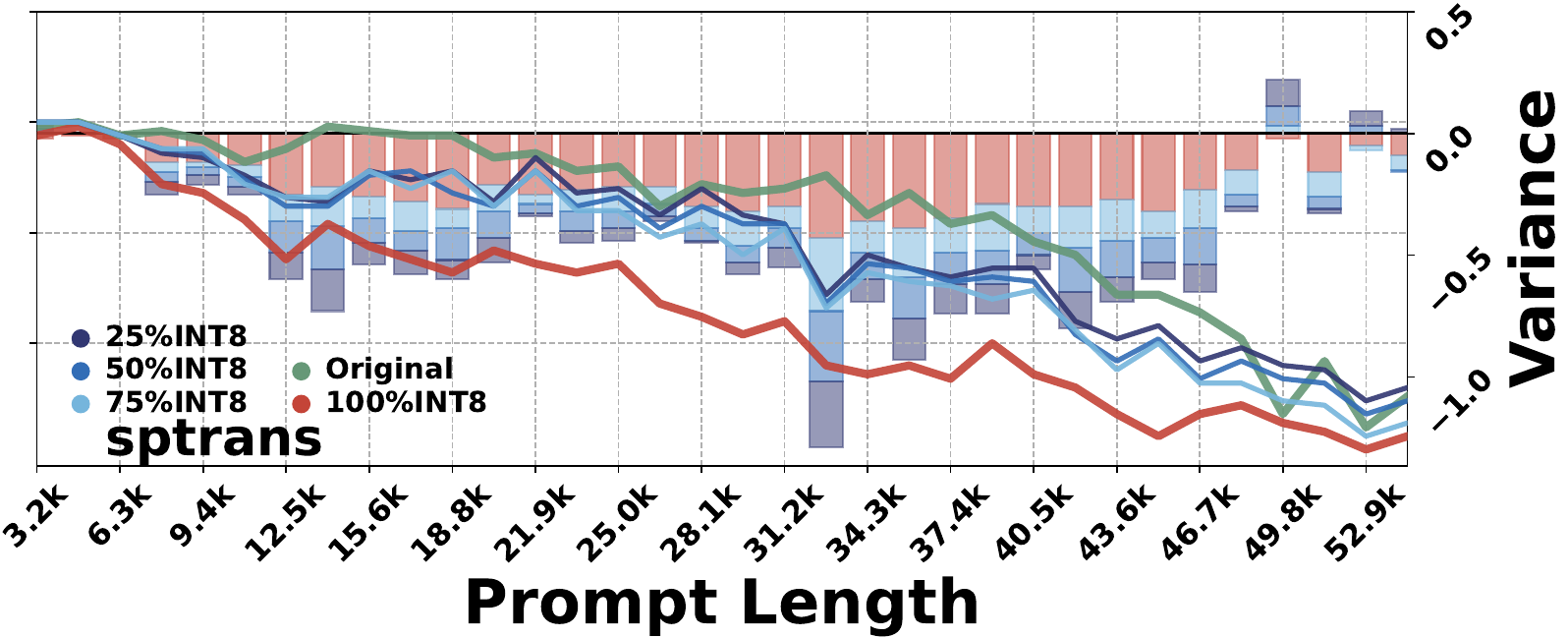}
    \end{subfigure}
    \vspace{-5pt}
    \caption{
    Line-level retrieval accuracy on the LongEval benchmark for \textbf{Qwen 2 7B} under different precision policy layouts, evaluated across prompt lengths from 3.1k to 38.7k tokens.
    }
    \label{fig:longeval_qwen2}
    \vspace{-10pt}
\end{figure*}

Figure~\ref{fig:longeval_qwen2.5} and Figure~\ref{fig:longeval_qwen2} report LongEval results for Qwen 2.5 7B and Qwen 2 7B across extended context lengths. Vicuna~7B, however, supports a maximum context length of 16k tokens. To avoid sparsely populated plots and to present results at all supported lengths clearly, we report Vicuna~7B performance in tabular form (Table~\ref{tab:longeval_vicuna}) instead of figures. The table follows the same evaluation protocol, precision layouts, and INT8 ratios as the figure-based results, allowing direct comparison across models within their respective context limits.

\begin{table*}[t]
\centering
\scriptsize
\setlength{\tabcolsep}{1.8pt}
\renewcommand{\arraystretch}{0.9}

\begin{adjustbox}{width=\textwidth}
\begin{tabular}{c @{\vsep}
                c @{\vsep}
                c @{\vsep}
                c @{\vsep}
                c c c @{\vsep}
                c c c @{\vsep}
                c c c @{\vsep}
                c c c @{\vsep}
                c c c @{\vsep}
                c c c}
\toprule
\multirow{2}{*}{\textbf{Line}} &
\multirow{2}{*}{\textbf{Len}} &
\multirow{2}{*}{\textbf{FP16}} &
\multirow{2}{*}{\textbf{INT8}} &
\multicolumn{3}{c}{\textbf{AlignSparse}} &
\multicolumn{3}{c}{\textbf{Band}} &
\multicolumn{3}{c}{\textbf{BigBird}} &
\multicolumn{3}{c}{\textbf{Global}} &
\multicolumn{3}{c}{\textbf{RowRand}} &
\multicolumn{3}{c}{\textbf{SpTrans}} \\
\cmidrule(lr){5-7}\cmidrule(lr){8-10}\cmidrule(lr){11-13}
\cmidrule(lr){14-16}\cmidrule(lr){17-19}\cmidrule(lr){20-22}
& & & &
\textbf{25} & \textbf{50} & \textbf{75} &
\textbf{25} & \textbf{50} & \textbf{75} &
\textbf{25} & \textbf{50} & \textbf{75} &
\textbf{25} & \textbf{50} & \textbf{75} &
\textbf{25} & \textbf{50} & \textbf{75} &
\textbf{25} & \textbf{50} & \textbf{75} \\
\midrule

200 & 4776.33 & 0.99 & 0.97
& 0.99 & 0.99 & 1.00
& 0.99 & 0.99 & 1.00
& 1.00 & 1.00 & 1.00
& 0.99 & 1.00 & 1.00
& 1.00 & 1.00 & 1.00
& 1.00 & 1.00 & 1.00 \\

300 & 7081.42 & 1.00 & 0.99
& 1.00 & 1.00 & 1.00
& 1.00 & 1.00 & 1.00
& 1.00 & 1.00 & 1.00
& 0.99 & 1.00 & 1.00
& 1.00 & 1.00 & 1.00
& 1.00 & 1.00 & 1.00 \\

400 & 9384.62 & 0.99 & 0.98
& 0.98 & 0.99 & 0.99
& 0.98 & 0.99 & 0.99
& 0.99 & 0.99 & 0.99
& 0.98 & 0.99 & 0.99
& 0.99 & 0.99 & 0.99
& 0.99 & 0.99 & 0.99 \\

500 & 11691.21 & 0.85 & 0.76
& 0.82 & 0.80 & 0.78
& 0.82 & 0.81 & 0.82
& 0.82 & 0.82 & 0.83
& 0.82 & 0.82 & 0.82
& 0.82 & 0.82 & 0.82
& 0.81 & 0.82 & 0.82 \\

600 & 13991.58 & 0.73 & 0.63
& 0.69 & 0.65 & 0.63
& 0.69 & 0.69 & 0.68
& 0.69 & 0.70 & 0.70
& 0.69 & 0.69 & 0.70
& 0.69 & 0.69 & 0.69
& 0.69 & 0.69 & 0.67 \\

700 & 16294.20 & 0.52 & 0.44
& 0.48 & 0.47 & 0.45
& 0.49 & 0.48 & 0.46
& 0.49 & 0.49 & 0.50
& 0.49 & 0.49 & 0.50
& 0.49 & 0.49 & 0.48
& 0.49 & 0.48 & 0.47 \\

\bottomrule
\end{tabular}
\end{adjustbox}
\caption{
Line-level retrieval accuracy on the LongEval benchmark for \textbf{Vicuna 7B} under different precision layouts inspired by sparse-attention patterns.
The task requires the model to scan long input sequences and exactly retrieve the content associated with a queried line identifier.
Accuracy is measured as exact-match retrieval rate, reported across increasing input lengths.
Results compare full FP16 attention, pure INT8 attention, and mixed-precision configurations with varying precision layouts and INT8 ratios, illustrating how different precision-allocation patterns affect robustness under low-precision execution.
}
\label{tab:longeval_vicuna}
\end{table*}

Together, Figure~\ref{fig:longeval_qwen2.5}, Figure~\ref{fig:longeval_qwen2}, and Table~\ref{tab:longeval_vicuna} show similar qualitative trends across Qwen 2.5 7B, Qwen 2 7B, and Vicuna 7B.
Across models, uniform INT8 attention (\textbf{One}) often degrades retrieval accuracy as prompt length increases, showing that a single low-precision arithmetic path can be too aggressive for long-context retrieval.
In contrast, mixed-routing configurations generally recover accuracy by keeping selected tile groups on the FP16 path while routing the remaining groups to INT8.

Across the three model families, mixed-routing configurations recover substantial retrieval quality relative to \textbf{One}, while sensitivity to layout and INT8 coverage varies with the model and prompt length.
\texttt{bigbird}, \texttt{row\_rand}, and \texttt{sptrans} provide the most consistent recovery across the evaluated models, particularly at higher INT8 coverage.

Overall, these results suggest that tile-group precision routing provides a useful precision-budget mechanism across model families, while the best routing layout and INT8 ratio remain model- and task-dependent.


\noindent
\textbf{Long-context Question Answering.}
Table~\ref{tab:lveval_qwen2_7b_sota_merged} isolates the comparison against sparse long-context and INT8 attention baselines on Qwen 2 7B. Across the reported LV-Eval subsets, SpTrans mixed-precision configurations close much of the gap to FP16 while outperforming or matching sparse baselines in most settings. This supports the central distinction of \modelname: it recovers long-context quality through precision routing while preserving dense token connectivity.

\newsavebox{\TableSevenBox}
\begin{lrbox}{\TableSevenBox}
\begin{minipage}{\textwidth}
\centering
\scriptsize
\setlength{\tabcolsep}{3.2pt}
\renewcommand{\arraystretch}{0.95}
\vspace{-10pt}

\begin{adjustbox}{width=\textwidth}
\begin{tabular}{c @{\vsep}
                c @{\vsep}
                c @{\vsep}
                c @{\vsep}
                c @{\vsep}
                c @{\vsep}
                c c c}
\toprule
\textbf{Dataset} &
\textbf{Len} &
\textbf{One} &
\textbf{SpTrans25} &
\textbf{SpTrans50} &
\textbf{SpTrans75} &
\textbf{MInference} &
\textbf{FlexPrefill} &
\textbf{SageAttn} \\
\midrule

\multirow{3}{*}{\ds{1}}
& 16k \setbase{36.00}\setone{31.60}
& \cellcolor{onered}\one
& \tcell{36.00} & \tcell{34.30} & \tcell{33.40}
& \tcell{29.00} & \tcell{26.13} & \tcell{32.60} \\
& 32k \setbase{20.00}\setone{14.70}
& \cellcolor{onered}\one
& \tcell{20.00} & \tcell{19.00} & \tcell{18.50}
& \tcell{15.00} & \tcell{11.89} & \tcell{16.90} \\
& 64k \setbase{9.70}\setone{6.60}
& \cellcolor{onered}\one
& \tcell{9.70} & \tcell{8.80} & \tcell{8.20}
& \tcell{6.40} & \tcell{8.98} & \tcell{9.10} \\
\midrule

\multirow{3}{*}{\ds{2}}
& 16k \setbase{16.70}\setone{13.90}
& \cellcolor{onered}\one
& \tcell{16.70} & \tcell{16.20} & \tcell{15.10}
& \tcell{14.10} & \tcell{15.74} & \tcell{14.20} \\
& 32k \setbase{15.10}\setone{11.40}
& \cellcolor{onered}\one
& \tcell{15.10} & \tcell{14.60} & \tcell{14.00}
& \tcell{11.50} & \tcell{15.24} & \tcell{13.50} \\
& 64k \setbase{13.20}\setone{8.70}
& \cellcolor{onered}\one
& \tcell{13.20} & \tcell{12.50} & \tcell{12.10}
& \tcell{9.50} & \tcell{12.65} & \tcell{10.80} \\
\midrule

\multirow{3}{*}{\ds{7}}
& 16k \setbase{17.10}\setone{12.20}
& \cellcolor{onered}\one
& \tcell{17.10} & \tcell{16.50} & \tcell{16.30}
& \tcell{14.90} & \tcell{17.09} & \tcell{15.80} \\
& 32k \setbase{15.30}\setone{9.30}
& \cellcolor{onered}\one
& \tcell{15.30} & \tcell{14.90} & \tcell{14.40}
& \tcell{12.60} & \tcell{16.08} & \tcell{13.80} \\
& 64k \setbase{11.70}\setone{5.80}
& \cellcolor{onered}\one
& \tcell{11.70} & \tcell{11.10} & \tcell{10.80}
& \tcell{9.00} & \tcell{10.29} & \tcell{11.40} \\
\midrule

\multirow{3}{*}{\ds{8}}
& 16k \setbase{16.90}\setone{12.10}
& \cellcolor{onered}\one
& \tcell{16.90} & \tcell{16.30} & \tcell{16.70}
& \tcell{15.20} & \tcell{13.65} & \tcell{15.10} \\
& 32k \setbase{12.60}\setone{8.40}
& \cellcolor{onered}\one
& \tcell{12.60} & \tcell{12.20} & \tcell{12.00}
& \tcell{10.80} & \tcell{12.26} & \tcell{10.80} \\
& 64k \setbase{10.80}\setone{7.30}
& \cellcolor{onered}\one
& \tcell{10.80} & \tcell{10.10} & \tcell{9.90}
& \tcell{8.10} & \tcell{9.46} & \tcell{9.50} \\
\midrule

\multirow{3}{*}{\ds{9}}
& 16k \setbase{33.80}\setone{25.70}
& \cellcolor{onered}\one
& \tcell{33.80} & \tcell{33.10} & \tcell{33.30}
& \tcell{28.50} & \tcell{29.07} & \tcell{31.40} \\
& 32k \setbase{22.90}\setone{15.70}
& \cellcolor{onered}\one
& \tcell{22.90} & \tcell{22.60} & \tcell{22.10}
& \tcell{18.10} & \tcell{20.71} & \tcell{21.00} \\
& 64k \setbase{15.90}\setone{9.00}
& \cellcolor{onered}\one
& \tcell{15.90} & \tcell{15.50} & \tcell{15.00}
& \tcell{11.40} & \tcell{14.53} & \tcell{13.70} \\
\midrule

\multirow{3}{*}{\ds{10}}
& 16k \setbase{25.10}\setone{19.30}
& \cellcolor{onered}\one
& \tcell{25.10} & \tcell{24.60} & \tcell{24.00}
& \tcell{21.20} & \tcell{23.89} & \tcell{22.60} \\
& 32k \setbase{19.30}\setone{14.20}
& \cellcolor{onered}\one
& \tcell{19.30} & \tcell{18.60} & \tcell{18.20}
& \tcell{15.30} & \tcell{17.67} & \tcell{17.90} \\
& 64k \setbase{17.50}\setone{14.00}
& \cellcolor{onered}\one
& \tcell{17.50} & \tcell{16.90} & \tcell{17.00}
& \tcell{12.80} & \tcell{15.95} & \tcell{15.90} \\
\midrule

\multirow{3}{*}{\ds{11}}
& 16k \setbase{34.30}\setone{25.30}
& \cellcolor{onered}\one
& \tcell{34.30} & \tcell{33.50} & \tcell{32.30}
& \tcell{28.70} & \tcell{22.74} & \tcell{31.40} \\
& 32k \setbase{25.30}\setone{20.10}
& \cellcolor{onered}\one
& \tcell{25.30} & \tcell{24.60} & \tcell{24.50}
& \tcell{19.30} & \tcell{15.98} & \tcell{23.50} \\
& 64k \setbase{18.40}\setone{6.90}
& \cellcolor{onered}\one
& \tcell{18.40} & \tcell{17.20} & \tcell{15.30}
& \tcell{13.00} & \tcell{15.22} & \tcell{16.10} \\
\bottomrule
\end{tabular}
\end{adjustbox}
\vspace{-10pt}
\captionof{table}{
LV-Eval long-context question answering results on Qwen2-7B.
We compare \modelname pure INT8 (\textbf{One}), \modelname SpTrans mixed-precision variants, and long-context attention baselines.
SpTrans25, SpTrans50, and SpTrans75 denote configurations with 25\%, 50\%, and 75\% INT8-routed tile groups.
Datasets are denoted by the same indices defined in the experimental setup.
}
\label{tab:lveval_qwen2_7b_sota_merged}
\end{minipage}
\end{lrbox}

Tables~\ref{tab:lveval_llama}, \ref{tab:lveval_qwen2}, and~\ref{tab:lveval_qwen2.5} report LV-Eval accuracy for \textbf{LLaMA 3.2 3B}, \textbf{Qwen 2 7B}, and \textbf{Qwen 2.5 7B} across context lengths from 16k to 64k under FP16 attention (Baseline), \textbf{One} (100\% INT8 tiles), and mixed-precision configurations with different precision layouts and INT8 ratios (25/50/75\%).

\begin{table*}[p]
\centering
\usebox{\TableSevenBox}
\vspace{-5pt}
\begin{minipage}{\textwidth}
\centering
\scriptsize
\setlength{\tabcolsep}{1.8pt}
\renewcommand{\arraystretch}{0.7}

\begin{adjustbox}{width=\textwidth}
\begin{tabular}{c @{\vsep}
                c @{\vsep}
                c @{\vsep}
                c @{\vsep}
                c c c @{\vsep}
                c c c @{\vsep}
                c c c @{\vsep}
                c c c @{\vsep}
                c c c @{\vsep}
                c c c}
\toprule
\multirow{2}{*}{\textbf{Dataset}} &
\multirow{2}{*}{\textbf{Len}} &
\multirow{2}{*}{\textbf{FP16}} &
\multirow{2}{*}{\textbf{One}} &
\multicolumn{3}{c}{\textbf{AlignSparse}} &
\multicolumn{3}{c}{\textbf{Band}} &
\multicolumn{3}{c}{\textbf{BigBird}} &
\multicolumn{3}{c}{\textbf{Global}} &
\multicolumn{3}{c}{\textbf{RowRand}} &
\multicolumn{3}{c}{\textbf{SpTrans}} \\
\cmidrule(lr){5-7}\cmidrule(lr){8-10}\cmidrule(lr){11-13}
\cmidrule(lr){14-16}\cmidrule(lr){17-19}\cmidrule(lr){20-22}
& &
& &
\textbf{25} & \textbf{50} & \textbf{75} &
\textbf{25} & \textbf{50} & \textbf{75} &
\textbf{25} & \textbf{50} & \textbf{75} &
\textbf{25} & \textbf{50} & \textbf{75} &
\textbf{25} & \textbf{50} & \textbf{75} &
\textbf{25} & \textbf{50} & \textbf{75} \\
\midrule

\multirow{3}{*}{\ds{1}}
& 16k \setbase{32.04}\setone{28.78} & \cellcolor{baselinegreen}\base
& \cellcolor{onered}\one
& \tcell{32.91} & \tcell{32.77} & \tcell{32.78}
& \tcell{32.52} & \tcell{30.76} & \tcell{30.86}
& \tcell{31.27} & \tcell{29.64} & \tcell{29.75}
& \tcell{31.75} & \tcell{30.18} & \tcell{29.87}
& \tcell{31.89} & \tcell{30.94} & \tcell{30.64}
& \tcell{31.75} & \tcell{32.18} & \tcell{30.25} \\
& 32k \setbase{15.08}\setone{11.62} & \cellcolor{baselinegreen}\base
& \cellcolor{onered}\one
& \tcell{15.75} & \tcell{13.38} & \tcell{12.42}
& \tcell{12.28} & \tcell{13.18} & \tcell{13.61}
& \tcell{15.22} & \tcell{14.02} & \tcell{13.55}
& \tcell{15.56} & \tcell{14.24} & \tcell{13.86}
& \tcell{15.26} & \tcell{13.23} & \tcell{12.61}
& \tcell{15.56} & \tcell{14.77} & \tcell{14.37} \\
& 64k \setbase{7.75}\setone{5.42} & \cellcolor{baselinegreen}\base
& \cellcolor{onered}\one
& \tcell{7.81} & \tcell{6.86} & \tcell{5.86}
& \tcell{7.81} & \tcell{5.84} & \tcell{5.81}
& \tcell{8.01} & \tcell{6.47} & \tcell{5.72}
& \tcell{8.01} & \tcell{7.03} & \tcell{6.25}
& \tcell{8.01} & \tcell{6.32} & \tcell{6.08}
& \tcell{8.01} & \tcell{7.44} & \tcell{6.83} \\
\midrule

\multirow{3}{*}{\ds{2}}
& 16k \setbase{18.49}\setone{15.62} & \cellcolor{baselinegreen}\base
& \cellcolor{onered}\one
& \tcell{18.95} & \tcell{17.08} & \tcell{16.83}
& \tcell{19.16} & \tcell{16.57} & \tcell{17.23}
& \tcell{18.61} & \tcell{16.98} & \tcell{16.62}
& \tcell{18.64} & \tcell{17.62} & \tcell{16.88}
& \tcell{18.39} & \tcell{16.58} & \tcell{16.56}
& \tcell{18.64} & \tcell{18.08} & \tcell{16.78} \\
& 32k \setbase{15.12}\setone{12.02} & \cellcolor{baselinegreen}\base
& \cellcolor{onered}\one
& \tcell{15.18} & \tcell{14.34} & \tcell{13.16}
& \tcell{15.48} & \tcell{13.59} & \tcell{13.09}
& \tcell{15.14} & \tcell{13.53} & \tcell{13.28}
& \tcell{15.27} & \tcell{14.78} & \tcell{13.59}
& \tcell{15.58} & \tcell{14.15} & \tcell{13.09}
& \tcell{15.27} & \tcell{14.69} & \tcell{14.18} \\
& 64k \setbase{11.84}\setone{7.79} & \cellcolor{baselinegreen}\base
& \cellcolor{onered}\one
& \tcell{12.15} & \tcell{10.00} & \tcell{9.12}
& \tcell{12.30} & \tcell{10.17} & \tcell{10.10}
& \tcell{12.04} & \tcell{10.28} & \tcell{10.03}
& \tcell{12.06} & \tcell{10.96} & \tcell{10.28}
& \tcell{12.19} & \tcell{10.39} & \tcell{10.13}
& \tcell{12.06} & \tcell{11.34} & \tcell{11.04} \\
\midrule

\multirow{3}{*}{\ds{3}}
& 16k \setbase{6.72}\setone{4.45} & \cellcolor{baselinegreen}\base
& \cellcolor{onered}\one
& \tcell{6.46} & \tcell{5.62} & \tcell{4.93}
& \tcell{6.52} & \tcell{24.37} & \tcell{4.37}
& \tcell{6.29} & \tcell{6.97} & \tcell{4.57}
& \tcell{6.29} & \tcell{5.31} & \tcell{4.50}
& \tcell{6.57} & \tcell{4.72} & \tcell{4.61}
& \tcell{21.04} & \tcell{20.53} & \tcell{21.65} \\
& 32k \setbase{3.78}\setone{1.95} & \cellcolor{baselinegreen}\base
& \cellcolor{onered}\one
& \tcell{3.77} & \tcell{3.07} & \tcell{2.92}
& \tcell{3.64} & \tcell{1.35} & \tcell{1.49}
& \tcell{2.35} & \tcell{1.81} & \tcell{1.20}
& \tcell{3.35} & \tcell{2.37} & \tcell{1.70}
& \tcell{5.96} & \tcell{2.04} & \tcell{2.65}
& \tcell{3.35} & \tcell{2.78} & \tcell{2.68} \\
& 64k \setbase{3.38}\setone{1.42} & \cellcolor{baselinegreen}\base
& \cellcolor{onered}\one
& \tcell{3.43} & \tcell{2.34} & \tcell{1.54}
& \tcell{3.40} & \tcell{1.42} & \tcell{1.41}
& \tcell{3.24} & \tcell{1.70} & \tcell{1.37}
& \tcell{3.24} & \tcell{2.26} & \tcell{1.45}
& \tcell{3.40} & \tcell{1.51} & \tcell{1.43}
& \tcell{3.24} & \tcell{2.67} & \tcell{2.47} \\
\midrule

\multirow{3}{*}{\ds{4}}
& 16k \setbase{6.00}\setone{2.27} & \cellcolor{baselinegreen}\base
& \cellcolor{onered}\one
& \tcell{7.50} & \tcell{6.00} & \tcell{5.50}
& \tcell{5.50} & \tcell{4.50} & \tcell{4.20}
& \tcell{6.00} & \tcell{4.46} & \tcell{4.49}
& \tcell{6.00} & \tcell{5.02} & \tcell{4.21}
& \tcell{6.00} & \tcell{4.76} & \tcell{5.02}
& \tcell{6.00} & \tcell{5.43} & \tcell{5.04} \\
& 32k \setbase{18.00}\setone{13.98} & \cellcolor{baselinegreen}\base
& \cellcolor{onered}\one
& \tcell{18.50} & \tcell{17.10} & \tcell{14.20}
& \tcell{18.50} & \tcell{16.50} & \tcell{17.00}
& \tcell{18.50} & \tcell{16.46} & \tcell{15.99}
& \tcell{18.00} & \tcell{17.02} & \tcell{16.71}
& \tcell{18.50} & \tcell{16.26} & \tcell{16.02}
& \tcell{18.00} & \tcell{17.43} & \tcell{17.54} \\
& 64k \setbase{14.50}\setone{10.07} & \cellcolor{baselinegreen}\base
& \cellcolor{onered}\one
& \tcell{15.00} & \tcell{14.20} & \tcell{12.50}
& \tcell{14.00} & \tcell{12.00} & \tcell{12.00}
& \tcell{15.00} & \tcell{13.46} & \tcell{12.49}
& \tcell{15.00} & \tcell{14.02} & \tcell{12.71}
& \tcell{15.50} & \tcell{13.26} & \tcell{13.02}
& \tcell{15.00} & \tcell{14.43} & \tcell{14.04} \\
\midrule

\multirow{3}{*}{\ds{5}}
& 16k \setbase{17.51}\setone{12.86} & \cellcolor{baselinegreen}\base
& \cellcolor{onered}\one
& \tcell{17.21} & \tcell{17.21} & \tcell{16.19}
& \tcell{17.69} & \tcell{15.51} & \tcell{15.25}
& \tcell{17.66} & \tcell{15.17} & \tcell{14.84}
& \tcell{17.66} & \tcell{15.87} & \tcell{15.87}
& \tcell{17.66} & \tcell{15.92} & \tcell{16.49}
& \tcell{17.66} & \tcell{17.09} & \tcell{16.88} \\
& 32k \setbase{10.80}\setone{8.38} & \cellcolor{baselinegreen}\base
& \cellcolor{onered}\one
& \tcell{11.01} & \tcell{10.15} & \tcell{9.01}
& \tcell{10.85} & \tcell{8.79} & \tcell{8.49}
& \tcell{10.61} & \tcell{9.07} & \tcell{9.00}
& \tcell{11.01} & \tcell{10.03} & \tcell{8.75}
& \tcell{10.80} & \tcell{9.28} & \tcell{9.21}
& \tcell{11.01} & \tcell{10.44} & \tcell{10.06} \\
& 64k \setbase{6.82}\setone{5.32} & \cellcolor{baselinegreen}\base
& \cellcolor{onered}\one
& \tcell{7.70} & \tcell{6.39} & \tcell{5.70}
& \tcell{6.89} & \tcell{5.97} & \tcell{5.39}
& \tcell{7.63} & \tcell{6.09} & \tcell{5.62}
& \tcell{7.63} & \tcell{6.65} & \tcell{5.84}
& \tcell{7.70} & \tcell{5.96} & \tcell{4.91}
& \tcell{7.63} & \tcell{6.25} & \tcell{5.97} \\
\midrule

\multirow{3}{*}{\ds{6}}
& 16k \setbase{12.33}\setone{8.51} & \cellcolor{baselinegreen}\base
& \cellcolor{onered}\one
& \tcell{12.43} & \tcell{11.77} & \tcell{9.79}
& \tcell{11.66} & \tcell{9.70} & \tcell{6.91}
& \tcell{12.38} & \tcell{10.40} & \tcell{10.30}
& \tcell{12.31} & \tcell{11.33} & \tcell{10.52}
& \tcell{12.02} & \tcell{10.27} & \tcell{10.53}
& \tcell{12.31} & \tcell{11.74} & \tcell{11.03} \\
& 32k \setbase{6.73}\setone{3.22} & \cellcolor{baselinegreen}\base
& \cellcolor{onered}\one
& \tcell{6.12} & \tcell{5.76} & \tcell{3.88}
& \tcell{7.04} & \tcell{4.84} & \tcell{4.90}
& \tcell{6.52} & \tcell{5.18} & \tcell{4.45}
& \tcell{7.08} & \tcell{6.10} & \tcell{4.73}
& \tcell{6.66} & \tcell{5.19} & \tcell{4.56}
& \tcell{7.08} & \tcell{6.51} & \tcell{5.95} \\
& 64k \setbase{1.84}\setone{0.74} & \cellcolor{baselinegreen}\base
& \cellcolor{onered}\one
& \tcell{1.99} & \tcell{1.87} & \tcell{1.02}
& \tcell{0.84} & \tcell{0.86} & \tcell{0.86}
& \tcell{1.91} & \tcell{1.35} & \tcell{1.13}
& \tcell{1.87} & \tcell{1.59} & \tcell{1.05}
& \tcell{1.99} & \tcell{1.89} & \tcell{1.24}
& \tcell{1.97} & \tcell{1.37} & \tcell{1.22} \\
\midrule

\multirow{3}{*}{\ds{7}}
& 16k \setbase{21.75}\setone{15.69} & \cellcolor{baselinegreen}\base
& \cellcolor{onered}\one
& \tcell{21.88} & \tcell{22.36} & \tcell{20.47}
& \tcell{27.84} & \tcell{19.86} & \tcell{19.03}
& \tcell{21.96} & \tcell{20.81} & \tcell{20.09}
& \tcell{22.07} & \tcell{21.08} & \tcell{20.23}
& \tcell{22.41} & \tcell{20.10} & \tcell{20.08}
& \tcell{22.07} & \tcell{21.16} & \tcell{21.05} \\
& 32k \setbase{19.89}\setone{12.11} & \cellcolor{baselinegreen}\base
& \cellcolor{onered}\one
& \tcell{19.16} & \tcell{17.35} & \tcell{14.23}
& \tcell{19.39} & \tcell{17.36} & \tcell{16.63}
& \tcell{20.17} & \tcell{18.41} & \tcell{17.79}
& \tcell{19.91} & \tcell{19.07} & \tcell{17.99}
& \tcell{19.88} & \tcell{17.96} & \tcell{17.47}
& \tcell{19.91} & \tcell{19.37} & \tcell{18.56} \\
& 64k \setbase{14.18}\setone{6.93} & \cellcolor{baselinegreen}\base
& \cellcolor{onered}\one
& \tcell{14.13} & \tcell{13.98} & \tcell{10.91}
& \tcell{14.13} & \tcell{12.45} & \tcell{11.25}
& \tcell{14.09} & \tcell{12.37} & \tcell{11.97}
& \tcell{14.23} & \tcell{13.31} & \tcell{12.23}
& \tcell{14.14} & \tcell{12.27} & \tcell{11.79}
& \tcell{14.23} & \tcell{13.66} & \tcell{13.34} \\
\midrule

\multirow{3}{*}{\ds{8}}
& 16k \setbase{18.24}\setone{12.72} & \cellcolor{baselinegreen}\base
& \cellcolor{onered}\one
& \tcell{18.16} & \tcell{16.12} & \tcell{13.13}
& \tcell{18.50} & \tcell{16.65} & \tcell{15.53}
& \tcell{17.57} & \tcell{16.45} & \tcell{15.96}
& \tcell{17.53} & \tcell{16.97} & \tcell{15.75}
& \tcell{18.12} & \tcell{16.27} & \tcell{16.28}
& \tcell{17.53} & \tcell{16.99} & \tcell{17.46} \\
& 32k \setbase{14.25}\setone{9.12} & \cellcolor{baselinegreen}\base
& \cellcolor{onered}\one
& \tcell{13.98} & \tcell{12.13} & \tcell{11.43}
& \tcell{14.44} & \tcell{12.64} & \tcell{11.40}
& \tcell{13.73} & \tcell{12.73} & \tcell{12.29}
& \tcell{13.63} & \tcell{13.03} & \tcell{12.21}
& \tcell{14.19} & \tcell{12.35} & \tcell{12.17}
& \tcell{13.63} & \tcell{13.18} & \tcell{13.00} \\
& 64k \setbase{11.08}\setone{7.54} & \cellcolor{baselinegreen}\base
& \cellcolor{onered}\one
& \tcell{11.06} & \tcell{9.11} & \tcell{8.28}
& \tcell{11.22} & \tcell{9.23} & \tcell{8.07}
& \tcell{11.05} & \tcell{9.60} & \tcell{9.04}
& \tcell{11.31} & \tcell{10.09} & \tcell{9.56}
& \tcell{11.08} & \tcell{9.18} & \tcell{8.73}
& \tcell{11.31} & \tcell{10.65} & \tcell{10.38} \\
\midrule

\multirow{3}{*}{\ds{9}}
& 16k \setbase{45.68}\setone{34.70} & \cellcolor{baselinegreen}\base
& \cellcolor{onered}\one
& \tcell{46.02} & \tcell{43.19} & \tcell{39.69}
& \tcell{45.91} & \tcell{43.60} & \tcell{42.61}
& \tcell{45.13} & \tcell{43.66} & \tcell{43.86}
& \tcell{45.77} & \tcell{44.79} & \tcell{43.52}
& \tcell{45.52} & \tcell{43.93} & \tcell{44.04}
& \tcell{45.77} & \tcell{44.60} & \tcell{44.88} \\
& 32k \setbase{26.84}\setone{18.17} & \cellcolor{baselinegreen}\base
& \cellcolor{onered}\one
& \tcell{25.76} & \tcell{23.20} & \tcell{21.69}
& \tcell{26.08} & \tcell{24.36} & \tcell{23.41}
& \tcell{26.47} & \tcell{24.94} & \tcell{23.78}
& \tcell{26.05} & \tcell{25.49} & \tcell{24.71}
& \tcell{26.90} & \tcell{25.05} & \tcell{24.58}
& \tcell{26.05} & \tcell{25.94} & \tcell{25.11} \\
& 64k \setbase{16.36}\setone{9.19} & \cellcolor{baselinegreen}\base
& \cellcolor{onered}\one
& \tcell{15.60} & \tcell{14.58} & \tcell{11.96}
& \tcell{16.30} & \tcell{14.21} & \tcell{12.45}
& \tcell{15.72} & \tcell{14.18} & \tcell{14.04}
& \tcell{16.03} & \tcell{14.85} & \tcell{14.24}
& \tcell{16.14} & \tcell{14.26} & \tcell{13.78}
& \tcell{16.03} & \tcell{15.25} & \tcell{14.71} \\
\midrule

\multirow{3}{*}{\ds{10}}
& 16k \setbase{28.88}\setone{22.37} & \cellcolor{baselinegreen}\base
& \cellcolor{onered}\one
& \tcell{28.26} & \tcell{26.21} & \tcell{25.80}
& \tcell{28.80} & \tcell{27.06} & \tcell{25.79}
& \tcell{28.95} & \tcell{27.57} & \tcell{26.33}
& \tcell{29.10} & \tcell{28.13} & \tcell{27.15}
& \tcell{28.17} & \tcell{26.31} & \tcell{26.42}
& \tcell{29.10} & \tcell{28.31} & \tcell{27.23} \\
& 32k \setbase{18.33}\setone{13.66} & \cellcolor{baselinegreen}\base
& \cellcolor{onered}\one
& \tcell{18.03} & \tcell{15.99} & \tcell{15.98}
& \tcell{17.84} & \tcell{16.20} & \tcell{15.13}
& \tcell{18.17} & \tcell{16.70} & \tcell{16.88}
& \tcell{18.26} & \tcell{17.35} & \tcell{16.44}
& \tcell{18.25} & \tcell{16.55} & \tcell{16.51}
& \tcell{18.26} & \tcell{17.65} & \tcell{17.41} \\
& 64k \setbase{15.75}\setone{12.81} & \cellcolor{baselinegreen}\base
& \cellcolor{onered}\one
& \tcell{16.08} & \tcell{15.91} & \tcell{14.98}
& \tcell{16.18} & \tcell{14.13} & \tcell{13.18}
& \tcell{15.93} & \tcell{14.22} & \tcell{13.88}
& \tcell{15.94} & \tcell{14.94} & \tcell{14.13}
& \tcell{15.85} & \tcell{14.77} & \tcell{14.41}
& \tcell{15.94} & \tcell{15.32} & \tcell{15.39} \\
\midrule

\multirow{3}{*}{\ds{11}}
& 16k \setbase{22.44}\setone{16.97} & \cellcolor{baselinegreen}\base
& \cellcolor{onered}\one
& \tcell{23.56} & \tcell{22.64} & \tcell{20.75}
& \tcell{22.36} & \tcell{21.03} & \tcell{19.89}
& \tcell{22.44} & \tcell{20.24} & \tcell{20.76}
& \tcell{20.29} & \tcell{21.46} & \tcell{26.06}
& \tcell{21.74} & \tcell{20.34} & \tcell{21.62}
& \tcell{22.94} & \tcell{22.37} & \tcell{21.74} \\
& 32k \setbase{14.85}\setone{12.02} & \cellcolor{baselinegreen}\base
& \cellcolor{onered}\one
& \tcell{15.53} & \tcell{15.28} & \tcell{14.49}
& \tcell{15.28} & \tcell{13.28} & \tcell{14.18}
& \tcell{16.11} & \tcell{14.57} & \tcell{13.34}
& \tcell{15.61} & \tcell{14.63} & \tcell{13.56}
& \tcell{15.81} & \tcell{13.57} & \tcell{13.54}
& \tcell{15.61} & \tcell{15.04} & \tcell{15.06} \\
& 64k \setbase{9.26}\setone{3.41} & \cellcolor{baselinegreen}\base
& \cellcolor{onered}\one
& \tcell{8.36} & \tcell{7.36} & \tcell{5.57}
& \tcell{8.36} & \tcell{6.36} & \tcell{5.36}
& \tcell{8.92} & \tcell{8.38} & \tcell{7.91}
& \tcell{8.92} & \tcell{7.94} & \tcell{7.13}
& \tcell{8.92} & \tcell{7.12} & \tcell{6.38}
& \tcell{8.92} & \tcell{8.35} & \tcell{7.40} \\
\bottomrule
\end{tabular}
\end{adjustbox}
\vspace{-10pt}
\captionof{table}{
LV-Eval long-context question answering for \textbf{LLaMA 3.2 3B}.
Cells are color-coded by comparison to \textbf{One} and \textbf{Baseline}: \textcolor{onered}{red} indicates performance below \textbf{One}, \textcolor{neutralblue}{blue} indicates performance between \textbf{One} and the \textbf{Baseline}, and \textcolor{baselinegreen}{green} indicates performance above the \textbf{Baseline}.
}
\label{tab:lveval_llama}
\vspace{-10pt}
\end{minipage}
\end{table*}

\newsavebox{\TableNineBox}
\begin{lrbox}{\TableNineBox}
\begin{minipage}{\textwidth}
\centering
\scriptsize
\setlength{\tabcolsep}{1.8pt}
\renewcommand{\arraystretch}{0.5}
\vspace{-5pt}
\begin{adjustbox}{width=\textwidth}
\begin{tabular}{c @{\vsep}
                c @{\vsep}
                c @{\vsep}
                c @{\vsep}
                c c c @{\vsep}
                c c c @{\vsep}
                c c c @{\vsep}
                c c c @{\vsep}
                c c c @{\vsep}
                c c c}
\toprule
\multirow{2}{*}{\textbf{Dataset}} &
\multirow{2}{*}{\textbf{Len}} &
\multirow{2}{*}{\textbf{FP16}} &
\multirow{2}{*}{\textbf{One}} &
\multicolumn{3}{c}{\textbf{AlignSparse}} &
\multicolumn{3}{c}{\textbf{Band}} &
\multicolumn{3}{c}{\textbf{BigBird}} &
\multicolumn{3}{c}{\textbf{Global}} &
\multicolumn{3}{c}{\textbf{RowRand}} &
\multicolumn{3}{c}{\textbf{SpTrans}} \\
\cmidrule(lr){5-7}\cmidrule(lr){8-10}\cmidrule(lr){11-13}
\cmidrule(lr){14-16}\cmidrule(lr){17-19}\cmidrule(lr){20-22}
& &
& &
\textbf{25} & \textbf{50} & \textbf{75} &
\textbf{25} & \textbf{50} & \textbf{75} &
\textbf{25} & \textbf{50} & \textbf{75} &
\textbf{25} & \textbf{50} & \textbf{75} &
\textbf{25} & \textbf{50} & \textbf{75} &
\textbf{25} & \textbf{50} & \textbf{75} \\
\midrule

\multirow{3}{*}{\ds{1}}
& 16k \setbase{34.97}\setone{31.6} & \cellcolor{baselinegreen}\base
& \cellcolor{onered}\one
& \tcell{36.0} & \tcell{35.6} & \tcell{35.5}
& \tcell{35.1} & \tcell{33.6} & \tcell{33.8}
& \tcell{34.4} & \tcell{32.7} & \tcell{32.9}
& \tcell{35.0} & \tcell{33.6} & \tcell{33.1}
& \tcell{35.8} & \tcell{34.7} & \tcell{34.1}
& \tcell{36.0} & \tcell{34.3} & \tcell{33.4} \\
& 32k \setbase{19.14}\setone{14.7} & \cellcolor{baselinegreen}\base
& \cellcolor{onered}\one
& \tcell{20.0} & \tcell{17.0} & \tcell{15.9}
& \tcell{15.6} & \tcell{16.7} & \tcell{17.0}
& \tcell{19.4} & \tcell{17.8} & \tcell{17.3}
& \tcell{19.8} & \tcell{18.1} & \tcell{17.6}
& \tcell{19.5} & \tcell{17.0} & \tcell{16.4}
& \tcell{20.0} & \tcell{19.0} & \tcell{18.5} \\
& 64k \setbase{9.34}\setone{6.6} & \cellcolor{baselinegreen}\base
& \cellcolor{onered}\one
& \tcell{9.6} & \tcell{8.4} & \tcell{7.3}
& \tcell{9.5} & \tcell{7.2} & \tcell{7.1}
& \tcell{9.7} & \tcell{7.8} & \tcell{7.1}
& \tcell{9.6} & \tcell{8.5} & \tcell{7.6}
& \tcell{9.8} & \tcell{7.7} & \tcell{7.4}
& \tcell{9.7} & \tcell{8.8} & \tcell{8.2} \\
\midrule

\multirow{3}{*}{\ds{2}}
& 16k \setbase{16.39}\setone{13.9} & \cellcolor{baselinegreen}\base
& \cellcolor{onered}\one
& \tcell{16.8} & \tcell{15.4} & \tcell{15.1}
& \tcell{17.1} & \tcell{15.0} & \tcell{15.5}
& \tcell{16.4} & \tcell{15.0} & \tcell{14.8}
& \tcell{16.6} & \tcell{15.7} & \tcell{15.0}
& \tcell{16.4} & \tcell{14.9} & \tcell{14.9}
& \tcell{16.7} & \tcell{16.2} & \tcell{15.1} \\
& 32k \setbase{14.42}\setone{11.4} & \cellcolor{baselinegreen}\base
& \cellcolor{onered}\one
& \tcell{14.3} & \tcell{13.4} & \tcell{12.6}
& \tcell{14.6} & \tcell{12.9} & \tcell{12.4}
& \tcell{14.5} & \tcell{13.0} & \tcell{12.6}
& \tcell{14.6} & \tcell{14.0} & \tcell{12.9}
& \tcell{14.8} & \tcell{13.4} & \tcell{12.7}
& \tcell{14.6} & \tcell{14.0} & \tcell{13.5} \\
& 64k \setbase{12.97}\setone{8.7} & \cellcolor{baselinegreen}\base
& \cellcolor{onered}\one
& \tcell{13.3} & \tcell{11.0} & \tcell{10.3}
& \tcell{13.4} & \tcell{11.2} & \tcell{11.0}
& \tcell{13.2} & \tcell{11.4} & \tcell{10.9}
& \tcell{13.3} & \tcell{12.0} & \tcell{11.2}
& \tcell{13.5} & \tcell{11.6} & \tcell{11.2}
& \tcell{13.2} & \tcell{12.5} & \tcell{12.1} \\
\midrule

\multirow{3}{*}{\ds{3}}
& 16k \setbase{16.39}\setone{10.7} & \cellcolor{baselinegreen}\base
& \cellcolor{onered}\one
& \tcell{15.8} & \tcell{13.9} & \tcell{12.2}
& \tcell{15.9} & \tcell{5.9} & \tcell{10.9}
& \tcell{15.4} & \tcell{16.6} & \tcell{11.0}
& \tcell{15.4} & \tcell{13.2} & \tcell{11.4}
& \tcell{16.1} & \tcell{11.5} & \tcell{11.2}
& \tcell{50.6} & \tcell{49.7} & \tcell{52.0} \\
& 32k \setbase{10.12}\setone{5.2} & \cellcolor{baselinegreen}\base
& \cellcolor{onered}\one
& \tcell{9.9} & \tcell{8.0} & \tcell{7.6}
& \tcell{9.6} & \tcell{3.5} & \tcell{3.9}
& \tcell{6.2} & \tcell{4.8} & \tcell{3.1}
& \tcell{9.2} & \tcell{6.5} & \tcell{4.7}
& \tcell{15.7} & \tcell{5.4} & \tcell{6.8}
& \tcell{9.1} & \tcell{7.4} & \tcell{7.1} \\
& 64k \setbase{5.33}\setone{2.2} & \cellcolor{baselinegreen}\base
& \cellcolor{onered}\one
& \tcell{5.4} & \tcell{3.9} & \tcell{2.6}
& \tcell{5.4} & \tcell{2.4} & \tcell{2.4}
& \tcell{5.2} & \tcell{2.7} & \tcell{2.3}
& \tcell{5.2} & \tcell{3.5} & \tcell{2.4}
& \tcell{5.5} & \tcell{2.5} & \tcell{2.4}
& \tcell{5.3} & \tcell{4.3} & \tcell{4.0} \\
\midrule


\multirow{3}{*}{\ds{5}}
& 16k \setbase{14.21}\setone{10.6} & \cellcolor{baselinegreen}\base
& \cellcolor{onered}\one
& \tcell{14.2} & \tcell{14.1} & \tcell{13.2}
& \tcell{14.4} & \tcell{12.8} & \tcell{12.5}
& \tcell{14.4} & \tcell{12.4} & \tcell{12.1}
& \tcell{14.4} & \tcell{12.9} & \tcell{12.8}
& \tcell{14.3} & \tcell{12.7} & \tcell{13.5}
& \tcell{14.5} & \tcell{14.0} & \tcell{13.8} \\
& 32k \setbase{8.16}\setone{6.3} & \cellcolor{baselinegreen}\base
& \cellcolor{onered}\one
& \tcell{8.3} & \tcell{7.6} & \tcell{6.7}
& \tcell{8.1} & \tcell{6.6} & \tcell{6.4}
& \tcell{8.0} & \tcell{6.9} & \tcell{6.8}
& \tcell{8.3} & \tcell{7.6} & \tcell{6.7}
& \tcell{8.2} & \tcell{7.1} & \tcell{7.0}
& \tcell{8.3} & \tcell{7.8} & \tcell{7.5} \\
& 64k \setbase{6.81}\setone{5.3} & \cellcolor{baselinegreen}\base
& \cellcolor{onered}\one
& \tcell{7.6} & \tcell{6.3} & \tcell{5.6}
& \tcell{7.4} & \tcell{6.4} & \tcell{5.7}
& \tcell{7.7} & \tcell{6.6} & \tcell{6.1}
& \tcell{7.7} & \tcell{6.7} & \tcell{6.0}
& \tcell{7.8} & \tcell{6.1} & \tcell{5.1}
& \tcell{7.7} & \tcell{6.3} & \tcell{6.0} \\
\midrule

\multirow{3}{*}{\ds{6}}
& 16k \setbase{23.28}\setone{16.0} & \cellcolor{baselinegreen}\base
& \cellcolor{onered}\one
& \tcell{23.3} & \tcell{22.1} & \tcell{18.6}
& \tcell{21.9} & \tcell{18.2} & \tcell{13.3}
& \tcell{23.2} & \tcell{19.7} & \tcell{19.4}
& \tcell{23.0} & \tcell{21.2} & \tcell{19.8}
& \tcell{22.6} & \tcell{19.3} & \tcell{19.9}
& \tcell{23.2} & \tcell{22.1} & \tcell{21.3} \\
& 32k \setbase{13.73}\setone{6.6} & \cellcolor{baselinegreen}\base
& \cellcolor{onered}\one
& \tcell{13.1} & \tcell{12.3} & \tcell{8.3}
& \tcell{14.4} & \tcell{9.9} & \tcell{10.0}
& \tcell{13.4} & \tcell{10.7} & \tcell{9.1}
& \tcell{14.6} & \tcell{12.6} & \tcell{9.7}
& \tcell{13.8} & \tcell{10.7} & \tcell{9.4}
& \tcell{14.6} & \tcell{13.5} & \tcell{12.3} \\
& 64k \setbase{6.91}\setone{2.9} & \cellcolor{baselinegreen}\base
& \cellcolor{onered}\one
& \tcell{7.4} & \tcell{7.0} & \tcell{3.8}
& \tcell{3.2} & \tcell{3.3} & \tcell{3.3}
& \tcell{7.2} & \tcell{5.1} & \tcell{4.3}
& \tcell{7.0} & \tcell{6.0} & \tcell{3.9}
& \tcell{7.4} & \tcell{7.0} & \tcell{4.7}
& \tcell{7.3} & \tcell{5.2} & \tcell{4.6} \\
\midrule

\multirow{3}{*}{\ds{7}}
& 16k \setbase{16.96}\setone{12.2} & \cellcolor{baselinegreen}\base
& \cellcolor{onered}\one
& \tcell{17.1} & \tcell{17.3} & \tcell{15.9}
& \tcell{21.1} & \tcell{15.1} & \tcell{14.6}
& \tcell{17.0} & \tcell{15.9} & \tcell{15.4}
& \tcell{17.2} & \tcell{16.4} & \tcell{15.8}
& \tcell{17.3} & \tcell{15.6} & \tcell{15.5}
& \tcell{17.1} & \tcell{16.5} & \tcell{16.3} \\
& 32k \setbase{15.09}\setone{9.3} & \cellcolor{baselinegreen}\base
& \cellcolor{onered}\one
& \tcell{14.7} & \tcell{13.6} & \tcell{11.1}
& \tcell{14.9} & \tcell{13.3} & \tcell{12.7}
& \tcell{15.4} & \tcell{14.1} & \tcell{13.5}
& \tcell{15.2} & \tcell{14.6} & \tcell{13.7}
& \tcell{15.1} & \tcell{13.8} & \tcell{13.4}
& \tcell{15.3} & \tcell{14.9} & \tcell{14.4} \\
& 64k \setbase{11.62}\setone{5.8} & \cellcolor{baselinegreen}\base
& \cellcolor{onered}\one
& \tcell{11.6} & \tcell{11.2} & \tcell{8.7}
& \tcell{11.5} & \tcell{10.1} & \tcell{9.2}
& \tcell{11.6} & \tcell{10.2} & \tcell{9.8}
& \tcell{11.6} & \tcell{10.9} & \tcell{10.2}
& \tcell{11.6} & \tcell{10.1} & \tcell{9.6}
& \tcell{11.7} & \tcell{11.1} & \tcell{10.8} \\
\midrule

\multirow{3}{*}{\ds{8}}
& 16k \setbase{17.36}\setone{12.1} & \cellcolor{baselinegreen}\base
& \cellcolor{onered}\one
& \tcell{17.2} & \tcell{15.3} & \tcell{12.5}
& \tcell{17.7} & \tcell{16.0} & \tcell{14.9}
& \tcell{16.8} & \tcell{15.7} & \tcell{15.2}
& \tcell{16.6} & \tcell{16.1} & \tcell{15.1}
& \tcell{17.2} & \tcell{15.6} & \tcell{15.7}
& \tcell{16.9} & \tcell{16.3} & \tcell{16.7} \\
& 32k \setbase{13.03}\setone{8.4} & \cellcolor{baselinegreen}\base
& \cellcolor{onered}\one
& \tcell{12.7} & \tcell{11.0} & \tcell{10.2}
& \tcell{13.3} & \tcell{11.6} & \tcell{10.5}
& \tcell{12.6} & \tcell{11.6} & \tcell{11.2}
& \tcell{12.4} & \tcell{11.8} & \tcell{11.1}
& \tcell{13.1} & \tcell{11.5} & \tcell{11.3}
& \tcell{12.6} & \tcell{12.2} & \tcell{12.0} \\
& 64k \setbase{10.73}\setone{7.3} & \cellcolor{baselinegreen}\base
& \cellcolor{onered}\one
& \tcell{10.8} & \tcell{9.1} & \tcell{8.2}
& \tcell{11.0} & \tcell{9.3} & \tcell{8.2}
& \tcell{10.7} & \tcell{9.2} & \tcell{8.7}
& \tcell{10.9} & \tcell{9.8} & \tcell{9.3}
& \tcell{10.7} & \tcell{9.3} & \tcell{8.9}
& \tcell{10.8} & \tcell{10.1} & \tcell{9.9} \\
\midrule

\multirow{3}{*}{\ds{9}}
& 16k \setbase{33.66}\setone{25.7} & \cellcolor{baselinegreen}\base
& \cellcolor{onered}\one
& \tcell{33.7} & \tcell{31.7} & \tcell{29.1}
& \tcell{33.8} & \tcell{32.2} & \tcell{31.5}
& \tcell{33.4} & \tcell{32.2} & \tcell{32.4}
& \tcell{33.7} & \tcell{33.0} & \tcell{32.1}
& \tcell{33.6} & \tcell{32.4} & \tcell{32.6}
& \tcell{33.8} & \tcell{33.1} & \tcell{33.3} \\
& 32k \setbase{23.30}\setone{15.7} & \cellcolor{baselinegreen}\base
& \cellcolor{onered}\one
& \tcell{22.6} & \tcell{20.5} & \tcell{19.0}
& \tcell{22.8} & \tcell{21.3} & \tcell{20.6}
& \tcell{23.2} & \tcell{21.9} & \tcell{20.9}
& \tcell{22.9} & \tcell{22.4} & \tcell{21.6}
& \tcell{23.6} & \tcell{22.0} & \tcell{21.5}
& \tcell{22.9} & \tcell{22.6} & \tcell{22.1} \\
& 64k \setbase{15.95}\setone{9.0} & \cellcolor{baselinegreen}\base
& \cellcolor{onered}\one
& \tcell{16.0} & \tcell{14.8} & \tcell{12.1}
& \tcell{15.9} & \tcell{13.9} & \tcell{12.5}
& \tcell{15.7} & \tcell{14.2} & \tcell{14.1}
& \tcell{16.0} & \tcell{14.9} & \tcell{14.3}
& \tcell{16.1} & \tcell{14.4} & \tcell{13.9}
& \tcell{15.9} & \tcell{15.5} & \tcell{15.0} \\
\midrule

\multirow{3}{*}{\ds{10}}
& 16k \setbase{24.88}\setone{19.3} & \cellcolor{baselinegreen}\base
& \cellcolor{onered}\one
& \tcell{24.4} & \tcell{22.9} & \tcell{22.5}
& \tcell{25.0} & \tcell{23.3} & \tcell{22.2}
& \tcell{25.2} & \tcell{23.9} & \tcell{23.0}
& \tcell{25.3} & \tcell{24.5} & \tcell{23.7}
& \tcell{24.7} & \tcell{23.1} & \tcell{23.3}
& \tcell{25.1} & \tcell{24.6} & \tcell{24.0} \\
& 32k \setbase{19.22}\setone{14.2} & \cellcolor{baselinegreen}\base
& \cellcolor{onered}\one
& \tcell{18.8} & \tcell{16.7} & \tcell{16.5}
& \tcell{18.8} & \tcell{17.1} & \tcell{16.1}
& \tcell{19.1} & \tcell{17.6} & \tcell{17.3}
& \tcell{19.3} & \tcell{18.4} & \tcell{17.5}
& \tcell{19.2} & \tcell{17.5} & \tcell{17.3}
& \tcell{19.3} & \tcell{18.6} & \tcell{18.2} \\
& 64k \setbase{17.20}\setone{14.0} & \cellcolor{baselinegreen}\base
& \cellcolor{onered}\one
& \tcell{17.7} & \tcell{17.2} & \tcell{16.3}
& \tcell{17.6} & \tcell{15.4} & \tcell{14.6}
& \tcell{17.4} & \tcell{15.7} & \tcell{15.4}
& \tcell{17.5} & \tcell{16.4} & \tcell{15.6}
& \tcell{17.3} & \tcell{16.2} & \tcell{15.9}
& \tcell{17.5} & \tcell{16.9} & \tcell{17.0} \\
\midrule

\multirow{3}{*}{\ds{11}}
& 16k \setbase{33.70}\setone{25.3} & \cellcolor{baselinegreen}\base
& \cellcolor{onered}\one
& \tcell{35.3} & \tcell{33.9} & \tcell{31.2}
& \tcell{33.5} & \tcell{31.5} & \tcell{29.8}
& \tcell{33.7} & \tcell{30.3} & \tcell{31.0}
& \tcell{33.9} & \tcell{32.4} & \tcell{31.2}
& \tcell{33.6} & \tcell{31.0} & \tcell{32.1}
& \tcell{34.3} & \tcell{33.5} & \tcell{32.3} \\
& 32k \setbase{24.82}\setone{20.1} & \cellcolor{baselinegreen}\base
& \cellcolor{onered}\one
& \tcell{25.6} & \tcell{25.0} & \tcell{23.7}
& \tcell{25.2} & \tcell{22.0} & \tcell{23.6}
& \tcell{26.2} & \tcell{23.7} & \tcell{21.8}
& \tcell{25.4} & \tcell{23.8} & \tcell{22.1}
& \tcell{25.9} & \tcell{22.3} & \tcell{22.2}
& \tcell{25.3} & \tcell{24.6} & \tcell{24.5} \\
& 64k \setbase{18.51}\setone{6.9} & \cellcolor{baselinegreen}\base
& \cellcolor{onered}\one
& \tcell{17.1} & \tcell{15.3} & \tcell{11.6}
& \tcell{17.1} & \tcell{13.1} & \tcell{11.1}
& \tcell{18.3} & \tcell{17.2} & \tcell{16.3}
& \tcell{18.3} & \tcell{16.4} & \tcell{14.9}
& \tcell{18.2} & \tcell{14.6} & \tcell{13.1}
& \tcell{18.4} & \tcell{17.2} & \tcell{15.3} \\

\bottomrule
\end{tabular}
\end{adjustbox}
\vspace{-10pt}
\captionof{table}{
LV-Eval for \textbf{Qwen 2 7B}.
Cells are color-coded by comparison to \textbf{One} and \textbf{Baseline}: \textcolor{onered}{red} indicates performance below \textbf{One}, \textcolor{neutralblue}{blue} indicates performance between \textbf{One} and the \textbf{Baseline}, and \textcolor{baselinegreen}{green} indicates performance above the \textbf{Baseline}.
}
\label{tab:lveval_qwen2}
\end{minipage}
\end{lrbox}

\newsavebox{\TableTenBox}
\begin{lrbox}{\TableTenBox}
\begin{minipage}{\textwidth}
\centering
\scriptsize
\setlength{\tabcolsep}{1.8pt}
\renewcommand{\arraystretch}{0.6}

\begin{adjustbox}{width=\textwidth}
\begin{tabular}{c @{\vsep}
                c @{\vsep}
                c @{\vsep}
                c @{\vsep}
                c c c @{\vsep}
                c c c @{\vsep}
                c c c @{\vsep}
                c c c @{\vsep}
                c c c @{\vsep}
                c c c}
\toprule
\multirow{2}{*}{\textbf{Dataset}} &
\multirow{2}{*}{\textbf{Len}} &
\multirow{2}{*}{\textbf{FP16}} &
\multirow{2}{*}{\textbf{One}} &
\multicolumn{3}{c}{\textbf{AlignSparse}} &
\multicolumn{3}{c}{\textbf{Band}} &
\multicolumn{3}{c}{\textbf{BigBird}} &
\multicolumn{3}{c}{\textbf{Global}} &
\multicolumn{3}{c}{\textbf{RowRand}} &
\multicolumn{3}{c}{\textbf{SpTrans}} \\
\cmidrule(lr){5-7}\cmidrule(lr){8-10}\cmidrule(lr){11-13}
\cmidrule(lr){14-16}\cmidrule(lr){17-19}\cmidrule(lr){20-22}
& &
& &
\textbf{25} & \textbf{50} & \textbf{75} &
\textbf{25} & \textbf{50} & \textbf{75} &
\textbf{25} & \textbf{50} & \textbf{75} &
\textbf{25} & \textbf{50} & \textbf{75} &
\textbf{25} & \textbf{50} & \textbf{75} &
\textbf{25} & \textbf{50} & \textbf{75} \\
\midrule
\multirow{3}{*}{\ds{1}}
& 16k \setbase{33.88}\setone{28.9} & \cellcolor{baselinegreen}\base
& \cellcolor{onered}\one
& \tcell{34.7} & \tcell{34.4} & \tcell{34.5}
& \tcell{33.9} & \tcell{32.0} & \tcell{32.1}
& \tcell{33.2} & \tcell{31.1} & \tcell{31.3}
& \tcell{33.7} & \tcell{32.0} & \tcell{31.7}
& \tcell{34.8} & \tcell{33.7} & \tcell{33.2}
& \tcell{34.9} & \tcell{33.1} & \tcell{32.2} \\
& 32k \setbase{17.61}\setone{13.6} & \cellcolor{baselinegreen}\base
& \cellcolor{onered}\one
& \tcell{18.3} & \tcell{15.6} & \tcell{14.6}
& \tcell{14.4} & \tcell{15.3} & \tcell{15.7}
& \tcell{17.8} & \tcell{16.4} & \tcell{15.9}
& \tcell{18.2} & \tcell{16.6} & \tcell{16.1}
& \tcell{17.8} & \tcell{15.5} & \tcell{14.9}
& \tcell{18.3} & \tcell{17.4} & \tcell{16.9} \\
& 64k \setbase{9.22}\setone{6.5} & \cellcolor{baselinegreen}\base
& \cellcolor{onered}\one
& \tcell{9.3} & \tcell{8.1} & \tcell{7.0}
& \tcell{9.3} & \tcell{7.0} & \tcell{6.9}
& \tcell{9.5} & \tcell{7.6} & \tcell{6.8}
& \tcell{9.4} & \tcell{8.3} & \tcell{7.4}
& \tcell{9.5} & \tcell{7.4} & \tcell{7.1}
& \tcell{9.4} & \tcell{8.5} & \tcell{7.9} \\
\midrule

\multirow{3}{*}{\ds{2}}
& 16k \setbase{17.37}\setone{14.8} & \cellcolor{baselinegreen}\base
& \cellcolor{onered}\one
& \tcell{17.8} & \tcell{16.2} & \tcell{15.9}
& \tcell{18.1} & \tcell{15.8} & \tcell{16.4}
& \tcell{17.4} & \tcell{15.9} & \tcell{15.6}
& \tcell{17.6} & \tcell{16.7} & \tcell{15.9}
& \tcell{17.4} & \tcell{15.8} & \tcell{15.7}
& \tcell{17.6} & \tcell{17.2} & \tcell{16.0} \\
& 32k \setbase{15.89}\setone{12.6} & \cellcolor{baselinegreen}\base
& \cellcolor{onered}\one
& \tcell{15.8} & \tcell{14.8} & \tcell{13.9}
& \tcell{16.0} & \tcell{14.1} & \tcell{13.6}
& \tcell{15.9} & \tcell{14.2} & \tcell{13.8}
& \tcell{16.0} & \tcell{15.4} & \tcell{14.2}
& \tcell{16.2} & \tcell{14.7} & \tcell{13.9}
& \tcell{16.0} & \tcell{15.3} & \tcell{14.8} \\
& 64k \setbase{12.12}\setone{8.1} & \cellcolor{baselinegreen}\base
& \cellcolor{onered}\one
& \tcell{12.4} & \tcell{10.2} & \tcell{9.5}
& \tcell{12.5} & \tcell{10.4} & \tcell{10.2}
& \tcell{12.3} & \tcell{10.6} & \tcell{10.1}
& \tcell{12.4} & \tcell{11.2} & \tcell{10.5}
& \tcell{12.6} & \tcell{10.8} & \tcell{10.4}
& \tcell{12.3} & \tcell{11.6} & \tcell{11.2} \\
\midrule

\multirow{3}{*}{\ds{3}}
& 16k \setbase{10.22}\setone{6.7} & \cellcolor{baselinegreen}\base
& \cellcolor{onered}\one
& \tcell{9.8} & \tcell{8.6} & \tcell{7.6}
& \tcell{9.9} & \tcell{3.7} & \tcell{6.8}
& \tcell{9.6} & \tcell{10.4} & \tcell{6.9}
& \tcell{9.6} & \tcell{8.2} & \tcell{7.0}
& \tcell{10.1} & \tcell{7.2} & \tcell{6.9}
& \tcell{31.4} & \tcell{30.7} & \tcell{32.2} \\
& 32k \setbase{6.03}\setone{3.1} & \cellcolor{baselinegreen}\base
& \cellcolor{onered}\one
& \tcell{6.0} & \tcell{4.8} & \tcell{4.6}
& \tcell{5.7} & \tcell{2.1} & \tcell{2.3}
& \tcell{3.7} & \tcell{2.9} & \tcell{1.9}
& \tcell{5.5} & \tcell{3.9} & \tcell{2.8}
& \tcell{9.4} & \tcell{3.3} & \tcell{4.1}
& \tcell{5.4} & \tcell{4.4} & \tcell{4.2} \\
& 64k \setbase{3.17}\setone{1.3} & \cellcolor{baselinegreen}\base
& \cellcolor{onered}\one
& \tcell{3.2} & \tcell{2.3} & \tcell{1.5}
& \tcell{3.2} & \tcell{1.4} & \tcell{1.4}
& \tcell{3.1} & \tcell{1.6} & \tcell{1.3}
& \tcell{3.1} & \tcell{2.1} & \tcell{1.4}
& \tcell{3.3} & \tcell{1.5} & \tcell{1.4}
& \tcell{3.2} & \tcell{2.6} & \tcell{2.4} \\
\midrule

\multirow{3}{*}{\ds{4}}
& 16k \setbase{4.64}\setone{1.7} & \cellcolor{baselinegreen}\base
& \cellcolor{onered}\one
& \tcell{5.8} & \tcell{4.6} & \tcell{4.3}
& \tcell{4.3} & \tcell{3.6} & \tcell{3.3}
& \tcell{4.6} & \tcell{3.5} & \tcell{3.5}
& \tcell{4.6} & \tcell{3.9} & \tcell{3.3}
& \tcell{4.7} & \tcell{3.8} & \tcell{4.0}
& \tcell{4.6} & \tcell{4.1} & \tcell{3.8} \\
& 32k \setbase{3.58}\setone{2.8} & \cellcolor{baselinegreen}\base
& \cellcolor{onered}\one
& \tcell{3.7} & \tcell{3.4} & \tcell{2.8}
& \tcell{3.7} & \tcell{3.3} & \tcell{3.4}
& \tcell{3.7} & \tcell{3.3} & \tcell{3.2}
& \tcell{3.6} & \tcell{3.4} & \tcell{3.3}
& \tcell{3.7} & \tcell{3.3} & \tcell{3.2}
& \tcell{3.6} & \tcell{3.5} & \tcell{3.6} \\
& 64k \setbase{1.74}\setone{1.2} & \cellcolor{baselinegreen}\base
& \cellcolor{onered}\one
& \tcell{1.8} & \tcell{1.7} & \tcell{1.5}
& \tcell{1.7} & \tcell{1.5} & \tcell{1.4}
& \tcell{1.8} & \tcell{1.6} & \tcell{1.5}
& \tcell{1.8} & \tcell{1.6} & \tcell{1.5}
& \tcell{1.9} & \tcell{1.6} & \tcell{1.5}
& \tcell{1.8} & \tcell{1.7} & \tcell{1.6} \\
\midrule

\multirow{3}{*}{\ds{5}}
& 16k \setbase{19.13}\setone{14.2} & \cellcolor{baselinegreen}\base
& \cellcolor{onered}\one
& \tcell{18.9} & \tcell{18.8} & \tcell{17.9}
& \tcell{19.2} & \tcell{17.1} & \tcell{16.7}
& \tcell{19.2} & \tcell{16.7} & \tcell{16.4}
& \tcell{19.2} & \tcell{17.3} & \tcell{17.1}
& \tcell{19.1} & \tcell{17.0} & \tcell{18.0}
& \tcell{19.3} & \tcell{18.7} & \tcell{18.2} \\
& 32k \setbase{8.72}\setone{6.8} & \cellcolor{baselinegreen}\base
& \cellcolor{onered}\one
& \tcell{8.9} & \tcell{8.1} & \tcell{7.3}
& \tcell{8.7} & \tcell{7.1} & \tcell{6.9}
& \tcell{8.6} & \tcell{7.4} & \tcell{7.3}
& \tcell{8.9} & \tcell{8.0} & \tcell{7.1}
& \tcell{8.8} & \tcell{7.6} & \tcell{7.5}
& \tcell{8.9} & \tcell{8.3} & \tcell{8.0} \\
& 64k \setbase{8.09}\setone{6.3} & \cellcolor{baselinegreen}\base
& \cellcolor{onered}\one
& \tcell{9.0} & \tcell{7.5} & \tcell{6.7}
& \tcell{8.8} & \tcell{7.7} & \tcell{6.9}
& \tcell{9.1} & \tcell{7.8} & \tcell{7.2}
& \tcell{9.1} & \tcell{7.9} & \tcell{7.1}
& \tcell{9.2} & \tcell{7.2} & \tcell{6.0}
& \tcell{9.0} & \tcell{7.4} & \tcell{7.1} \\
\midrule

\multirow{3}{*}{\ds{6}}
& 16k \setbase{23.19}\setone{16.0} & \cellcolor{baselinegreen}\base
& \cellcolor{onered}\one
& \tcell{23.2} & \tcell{22.0} & \tcell{18.5}
& \tcell{21.8} & \tcell{18.1} & \tcell{13.1}
& \tcell{23.1} & \tcell{19.6} & \tcell{19.3}
& \tcell{22.9} & \tcell{21.1} & \tcell{19.7}
& \tcell{22.5} & \tcell{19.2} & \tcell{19.8}
& \tcell{23.1} & \tcell{22.0} & \tcell{21.2} \\
& 32k \setbase{16.05}\setone{7.7} & \cellcolor{baselinegreen}\base
& \cellcolor{onered}\one
& \tcell{15.3} & \tcell{14.4} & \tcell{9.7}
& \tcell{16.9} & \tcell{11.6} & \tcell{11.8}
& \tcell{15.7} & \tcell{12.5} & \tcell{10.7}
& \tcell{17.1} & \tcell{14.8} & \tcell{11.4}
& \tcell{16.1} & \tcell{12.5} & \tcell{11.0}
& \tcell{17.1} & \tcell{15.8} & \tcell{14.4} \\
& 64k \setbase{5.58}\setone{2.3} & \cellcolor{baselinegreen}\base
& \cellcolor{onered}\one
& \tcell{6.0} & \tcell{5.7} & \tcell{3.1}
& \tcell{2.6} & \tcell{2.7} & \tcell{2.7}
& \tcell{5.8} & \tcell{4.1} & \tcell{3.4}
& \tcell{5.7} & \tcell{4.9} & \tcell{3.2}
& \tcell{6.0} & \tcell{5.7} & \tcell{3.8}
& \tcell{5.9} & \tcell{4.2} & \tcell{3.7} \\
\midrule

\multirow{3}{*}{\ds{7}}
& 16k \setbase{21.60}\setone{15.5} & \cellcolor{baselinegreen}\base
& \cellcolor{onered}\one
& \tcell{21.8} & \tcell{22.1} & \tcell{20.3}
& \tcell{26.8} & \tcell{19.1} & \tcell{18.5}
& \tcell{21.7} & \tcell{20.4} & \tcell{19.8}
& \tcell{21.9} & \tcell{20.9} & \tcell{20.1}
& \tcell{22.0} & \tcell{20.2} & \tcell{20.0}
& \tcell{21.8} & \tcell{21.0} & \tcell{20.9} \\
& 32k \setbase{18.85}\setone{11.6} & \cellcolor{baselinegreen}\base
& \cellcolor{onered}\one
& \tcell{18.3} & \tcell{16.9} & \tcell{13.8}
& \tcell{18.7} & \tcell{16.7} & \tcell{16.0}
& \tcell{19.4} & \tcell{17.8} & \tcell{17.0}
& \tcell{19.2} & \tcell{18.4} & \tcell{17.3}
& \tcell{19.1} & \tcell{17.4} & \tcell{16.9}
& \tcell{19.3} & \tcell{18.8} & \tcell{18.1} \\
& 64k \setbase{12.67}\setone{6.3} & \cellcolor{baselinegreen}\base
& \cellcolor{onered}\one
& \tcell{12.6} & \tcell{12.1} & \tcell{9.4}
& \tcell{12.5} & \tcell{11.0} & \tcell{10.0}
& \tcell{12.6} & \tcell{11.1} & \tcell{10.7}
& \tcell{12.7} & \tcell{11.9} & \tcell{11.1}
& \tcell{12.6} & \tcell{11.0} & \tcell{10.5}
& \tcell{12.7} & \tcell{12.1} & \tcell{11.8} \\
\midrule

\multirow{3}{*}{\ds{8}}
& 16k \setbase{21.27}\setone{14.8} & \cellcolor{baselinegreen}\base
& \cellcolor{onered}\one
& \tcell{21.1} & \tcell{18.7} & \tcell{15.2}
& \tcell{21.6} & \tcell{19.5} & \tcell{18.2}
& \tcell{20.5} & \tcell{19.2} & \tcell{18.6}
& \tcell{20.3} & \tcell{19.7} & \tcell{18.4}
& \tcell{21.1} & \tcell{19.1} & \tcell{19.2}
& \tcell{20.6} & \tcell{19.8} & \tcell{20.3} \\
& 32k \setbase{16.56}\setone{10.7} & \cellcolor{baselinegreen}\base
& \cellcolor{onered}\one
& \tcell{16.2} & \tcell{14.1} & \tcell{13.1}
& \tcell{16.9} & \tcell{14.8} & \tcell{13.4}
& \tcell{16.1} & \tcell{14.9} & \tcell{14.4}
& \tcell{15.9} & \tcell{15.2} & \tcell{14.3}
& \tcell{16.7} & \tcell{14.6} & \tcell{14.4}
& \tcell{16.1} & \tcell{15.6} & \tcell{15.4} \\
& 64k \setbase{12.90}\setone{8.8} & \cellcolor{baselinegreen}\base
& \cellcolor{onered}\one
& \tcell{12.9} & \tcell{10.9} & \tcell{9.8}
& \tcell{13.2} & \tcell{11.1} & \tcell{9.8}
& \tcell{12.8} & \tcell{11.0} & \tcell{10.4}
& \tcell{13.1} & \tcell{11.8} & \tcell{11.2}
& \tcell{12.9} & \tcell{11.2} & \tcell{10.7}
& \tcell{13.0} & \tcell{12.2} & \tcell{11.9} \\
\midrule

\multirow{3}{*}{\ds{9}}
& 16k \setbase{41.68}\setone{31.8} & \cellcolor{baselinegreen}\base
& \cellcolor{onered}\one
& \tcell{41.9} & \tcell{39.3} & \tcell{36.1}
& \tcell{41.8} & \tcell{39.9} & \tcell{39.0}
& \tcell{41.3} & \tcell{39.9} & \tcell{40.1}
& \tcell{41.7} & \tcell{40.9} & \tcell{39.7}
& \tcell{41.6} & \tcell{40.1} & \tcell{40.3}
& \tcell{41.8} & \tcell{41.0} & \tcell{41.2} \\
& 32k \setbase{26.25}\setone{17.7} & \cellcolor{baselinegreen}\base
& \cellcolor{onered}\one
& \tcell{25.4} & \tcell{23.0} & \tcell{21.3}
& \tcell{25.7} & \tcell{24.0} & \tcell{23.2}
& \tcell{26.1} & \tcell{24.6} & \tcell{23.5}
& \tcell{25.8} & \tcell{25.2} & \tcell{24.3}
& \tcell{26.6} & \tcell{24.8} & \tcell{24.2}
& \tcell{25.9} & \tcell{25.6} & \tcell{25.0} \\
& 64k \setbase{14.08}\setone{7.9} & \cellcolor{baselinegreen}\base
& \cellcolor{onered}\one
& \tcell{14.1} & \tcell{13.0} & \tcell{10.6}
& \tcell{14.0} & \tcell{12.2} & \tcell{11.0}
& \tcell{13.8} & \tcell{12.5} & \tcell{12.4}
& \tcell{14.1} & \tcell{13.1} & \tcell{12.6}
& \tcell{14.2} & \tcell{12.7} & \tcell{12.2}
& \tcell{14.0} & \tcell{13.6} & \tcell{13.1} \\
\midrule

\multirow{3}{*}{\ds{10}}
& 16k \setbase{26.81}\setone{20.8} & \cellcolor{baselinegreen}\base
& \cellcolor{onered}\one
& \tcell{26.3} & \tcell{24.6} & \tcell{24.2}
& \tcell{26.9} & \tcell{25.1} & \tcell{23.9}
& \tcell{27.1} & \tcell{25.7} & \tcell{24.7}
& \tcell{27.2} & \tcell{26.3} & \tcell{25.5}
& \tcell{26.6} & \tcell{24.9} & \tcell{25.1}
& \tcell{27.0} & \tcell{26.4} & \tcell{25.8} \\
& 32k \setbase{18.63}\setone{13.8} & \cellcolor{baselinegreen}\base
& \cellcolor{onered}\one
& \tcell{18.2} & \tcell{16.2} & \tcell{16.0}
& \tcell{18.2} & \tcell{16.6} & \tcell{15.6}
& \tcell{18.5} & \tcell{17.1} & \tcell{16.8}
& \tcell{18.7} & \tcell{17.8} & \tcell{16.9}
& \tcell{18.6} & \tcell{16.9} & \tcell{16.7}
& \tcell{18.7} & \tcell{18.0} & \tcell{17.6} \\
& 64k \setbase{16.42}\setone{13.4} & \cellcolor{baselinegreen}\base
& \cellcolor{onered}\one
& \tcell{16.9} & \tcell{16.4} & \tcell{15.5}
& \tcell{16.8} & \tcell{14.7} & \tcell{13.9}
& \tcell{16.6} & \tcell{15.0} & \tcell{14.7}
& \tcell{16.7} & \tcell{15.7} & \tcell{14.9}
& \tcell{16.5} & \tcell{15.4} & \tcell{15.1}
& \tcell{16.7} & \tcell{16.1} & \tcell{16.2} \\
\midrule

\multirow{3}{*}{\ds{11}}
& 16k \setbase{33.47}\setone{25.1} & \cellcolor{baselinegreen}\base
& \cellcolor{onered}\one
& \tcell{35.1} & \tcell{33.7} & \tcell{31.0}
& \tcell{33.3} & \tcell{31.3} & \tcell{29.6}
& \tcell{33.5} & \tcell{30.1} & \tcell{30.8}
& \tcell{33.7} & \tcell{32.2} & \tcell{31.0}
& \tcell{33.4} & \tcell{30.8} & \tcell{31.9}
& \tcell{34.1} & \tcell{33.3} & \tcell{32.1} \\
& 32k \setbase{22.36}\setone{18.1} & \cellcolor{baselinegreen}\base
& \cellcolor{onered}\one
& \tcell{23.1} & \tcell{22.6} & \tcell{21.4}
& \tcell{22.7} & \tcell{19.8} & \tcell{21.2}
& \tcell{23.6} & \tcell{21.3} & \tcell{19.6}
& \tcell{22.9} & \tcell{21.4} & \tcell{19.9}
& \tcell{23.3} & \tcell{20.1} & \tcell{20.0}
& \tcell{22.8} & \tcell{22.1} & \tcell{22.0} \\
& 64k \setbase{14.06}\setone{5.2} & \cellcolor{baselinegreen}\base
& \cellcolor{onered}\one
& \tcell{12.9} & \tcell{11.5} & \tcell{8.8}
& \tcell{13.0} & \tcell{9.9} & \tcell{8.4}
& \tcell{13.9} & \tcell{13.0} & \tcell{12.3}
& \tcell{13.9} & \tcell{12.4} & \tcell{11.2}
& \tcell{13.8} & \tcell{11.0} & \tcell{9.9}
& \tcell{14.0} & \tcell{13.1} & \tcell{11.6} \\
\bottomrule
\end{tabular}
\end{adjustbox}
\captionof{table}{
LV-Eval long-context question answering for \textbf{Qwen 2.5 7B} under different precision layouts and INT8 ratios (25/50/75\%).
Results are reported for FP16 attention (Baseline), \textbf{One} (all legal score-tile groups routed to INT8), and mixed-precision configurations.
Cells are color-coded by comparison to \textbf{One} and \textbf{Baseline}: \textcolor{onered}{red} indicates performance below \textbf{One}, \textcolor{neutralblue}{blue} indicates performance between \textbf{One} and the \textbf{Baseline}, and \textcolor{baselinegreen}{green} indicates performance above the \textbf{Baseline}.
}
\label{tab:lveval_qwen2.5}
\end{minipage}
\end{lrbox}

Across datasets and lengths, \textbf{One} generally underperforms the FP16 baseline, indicating that uniform INT8 attention can be too coarse for long-context QA.
Mixed-routing configurations usually narrow this gap, especially at moderate INT8 ratios.
The effect of increasing INT8 coverage is layout dependent: \texttt{align\_sparse}, \texttt{band}, and \texttt{global} often benefit from more conservative INT8 ratios, whereas \texttt{bigbird}, \texttt{row\_rand}, and \texttt{sptrans} often tolerate higher INT8 coverage.
A consistent layout--task interaction appears on the 16k \texttt{factrecall\_en} setting.
Across SpTrans25, SpTrans50, and SpTrans75, the scores reach 21.04/20.53/21.65 on LLaMA 3.2 3B, 50.6/49.7/52.0 on Qwen 2 7B, and 31.4/30.7/32.2 on Qwen 2.5 7B, compared with the corresponding FP16 scores of 6.72, 16.39, and 10.22.
The recurrence across model families and all three routing ratios shows a cross-model consistent layout--task interaction at this setting, while the corresponding SpTrans results at 32k and 64k return to the usual quality range.
Overall, these tables support the use of tile-group routing as a controllable precision-budget mechanism for long-context QA, while the best routing layout depends on the model, task, and context length.

\section{Efficiency}\label{app:efficiency}

Tables~\ref{tab:efficiency_llama3.2}, \ref{tab:efficiency_qwen2.5}, \ref{tab:efficiency_qwen2}, and~\ref{tab:efficiency_vicuna} report implementation-level prefill efficiency across LLaMA 3.2 3B, Qwen 2.5 7B, Qwen 2 7B, and Vicuna 7B using throughput (Thpt, K tokens/s) and TOPS from 1k to 8k where executable, with model-dependent maximum sequence lengths (batch size 8; 3 warmup iterations; 5 measurement iterations).
We compare the standard Torch implementation, FlashAttention as an IO-aware FP16 baseline, \textbf{One} (uniform all legal score-tile groups routed to INT8) as an efficiency-oriented reference, and \modelname mixed-routing variants under different routing layouts and INT8 ratios (25/50/75\%).

\newsavebox{\TableElevenBox}
\begin{lrbox}{\TableElevenBox}
\begin{minipage}{\textwidth}
\centering
\scriptsize
\setlength{\tabcolsep}{1.8pt}
\renewcommand{\arraystretch}{0.9}
\begin{adjustbox}{width=\textwidth}
\begin{tabular}{
c @{\vsep}  
c @{\vsep}  
c @{\vsep}  
c @{\vsep}  
c @{\vsep}  
c c c @{\vsep}          
c c c @{\vsep}          
c c c @{\vsep}          
c c c @{\vsep}          
c c c @{\vsep}          
c c c                     
}

\toprule
\multirow{2}{*}{\rotatebox{90}{\textbf{Len}}} &
\multirow{2}{*}{\rotatebox{90}{\scalebox{0.60}[1]{\textbf{Metric}}}} &
\multirow{2}{*}{\rotatebox{90}{\scalebox{0.68}[1]{\textbf{Torch}}}} &
\multirow{2}{*}{\rotatebox{90}{\scalebox{0.68}[1]{\textbf{Flash}}}} &
\multirow{2}{*}{\rotatebox{90}{\textbf{One}}} &
\multicolumn{3}{c}{\textbf{AlignSparse}} &
\multicolumn{3}{c}{\textbf{Band}} &
\multicolumn{3}{c}{\textbf{BigBird}} &
\multicolumn{3}{c}{\textbf{Global}} &
\multicolumn{3}{c}{\textbf{RowRand}} &
\multicolumn{3}{c}{\textbf{SpTrans}} \\

\cmidrule(lr){6-8}
\cmidrule(lr){9-11}
\cmidrule(lr){12-14}
\cmidrule(lr){15-17}
\cmidrule(lr){18-20}
\cmidrule(lr){21-23}

& & & & 
& \textbf{25} & \textbf{50} & \textbf{75}
& \textbf{25} & \textbf{50} & \textbf{75}
& \textbf{25} & \textbf{50} & \textbf{75}
& \textbf{25} & \textbf{50} & \textbf{75}
& \textbf{25} & \textbf{50} & \textbf{75}
& \textbf{25} & \textbf{50} & \textbf{75} \\

\midrule

1k & Thpt
& \setlow{17.45}\setup{32.27}\tcellFO{11.14}
& \cellcolor{onered!35}17.45
& \cellcolor{baselinegreen!35}32.27
& \tcellFO{29.22} & \tcellFO{32.02} & \tcellFO{33.48}
& \tcellFO{27.06} & \tcellFO{29.30} & \tcellFO{34.56}
& \tcellFO{26.92} & \tcellFO{31.48} & \tcellFO{32.37}
& \tcellFO{26.71} & \tcellFO{29.59} & \tcellFO{32.03}
& \tcellFO{28.66} & \tcellFO{32.09} & \tcellFO{34.95}
& \tcellFO{27.66} & \tcellFO{32.11} & \tcellFO{33.50} \\

& TOPS
& \setlow{65.31}\setup{120.80}\tcellFO{41.70}
& \cellcolor{onered!35}65.31
& \cellcolor{baselinegreen!35}120.80
& \tcellFO{109.38} & \tcellFO{119.86} & \tcellFO{125.29}
& \tcellFO{101.31} & \tcellFO{109.71} & \tcellFO{129.42}
& \tcellFO{100.78} & \tcellFO{117.85} & \tcellFO{121.17}
& \tcellFO{99.98} & \tcellFO{110.75} & \tcellFO{119.92}
& \tcellFO{107.25} & \tcellFO{120.12} & \tcellFO{130.80}
& \tcellFO{103.50} & \tcellFO{120.20} & \tcellFO{125.38} \\

\midrule

2k & Thpt
& \setlow{16.48}\setup{32.06}\tcellFO{7.78}
& \cellcolor{onered!35}16.48
& \cellcolor{baselinegreen!35}32.06
& \tcellFO{29.50} & \tcellFO{32.41} & \tcellFO{33.84}
& \tcellFO{26.77} & \tcellFO{28.89} & \tcellFO{34.30}
& \tcellFO{26.57} & \tcellFO{28.83} & \tcellFO{32.71}
& \tcellFO{26.45} & \tcellFO{29.29} & \tcellFO{31.74}
& \tcellFO{26.34} & \tcellFO{31.35} & \tcellFO{34.88}
& \tcellFO{27.11} & \tcellFO{31.46} & \tcellFO{33.92} \\

& TOPS
& \setlow{64.64}\setup{125.70}\tcellFO{30.50}
& \cellcolor{onered!35}64.64
& \cellcolor{baselinegreen!35}125.70
& \tcellFO{115.70} & \tcellFO{127.08} & \tcellFO{132.67}
& \tcellFO{104.99} & \tcellFO{113.30} & \tcellFO{134.48}
& \tcellFO{104.23} & \tcellFO{113.05} & \tcellFO{128.31}
& \tcellFO{103.69} & \tcellFO{114.83} & \tcellFO{124.46}
& \tcellFO{103.28} & \tcellFO{122.91} & \tcellFO{136.77}
& \tcellFO{106.32} & \tcellFO{123.38} & \tcellFO{132.99} \\

\midrule

4k & Thpt
& \setlow{14.33}\setup{29.80}\tcellFO{OOM}
& \cellcolor{onered!35}14.33
& \cellcolor{baselinegreen!35}29.80
& \tcellFO{27.71} & \tcellFO{30.35} & \tcellFO{31.62}
& \tcellFO{25.65} & \tcellFO{28.59} & \tcellFO{32.81}
& \tcellFO{24.69} & \tcellFO{26.84} & \tcellFO{27.48}
& \tcellFO{24.50} & \tcellFO{27.19} & \tcellFO{29.35}
& \tcellFO{26.97} & \tcellFO{30.06} & \tcellFO{32.89}
& \tcellFO{27.14} & \tcellFO{30.59} & \tcellFO{31.80} \\

& TOPS
& \setlow{61.31}\setup{127.48}\tcellFO{OOM}
& \cellcolor{onered!35}61.31
& \cellcolor{baselinegreen!35}127.48
& \tcellFO{118.53} & \tcellFO{129.85} & \tcellFO{135.26}
& \tcellFO{109.76} & \tcellFO{122.32} & \tcellFO{140.39}
& \tcellFO{105.63} & \tcellFO{114.83} & \tcellFO{117.56}
& \tcellFO{104.82} & \tcellFO{116.35} & \tcellFO{125.56}
& \tcellFO{115.37} & \tcellFO{128.62} & \tcellFO{140.73}
& \tcellFO{116.09} & \tcellFO{130.85} & \tcellFO{136.03} \\

\midrule

8k & Thpt
& \setlow{0}\setup{27.41}\tcellFO{OOM}
& \oom
& \cellcolor{baselinegreen!35}27.41
& \tcellFO{23.17} & \tcellFO{25.39} & \tcellFO{26.44}
& \tcellFO{23.23} & \tcellFO{25.03} & \tcellFO{28.19}
& \tcellFO{23.02} & \tcellFO{25.17} & \tcellFO{25.84}
& \tcellFO{22.84} & \tcellFO{25.39} & \tcellFO{27.47}
& \tcellFO{22.78} & \tcellFO{25.33} & \tcellFO{27.74}
& \tcellFO{22.81} & \tcellFO{25.69} & \tcellFO{26.61} \\

& TOPS
& \setlow{0}\setup{136.76}\tcellFO{OOM}
& \oom
& \cellcolor{baselinegreen!35}136.76
& \tcellFO{115.64} & \tcellFO{126.72} & \tcellFO{131.96}
& \tcellFO{115.89} & \tcellFO{124.89} & \tcellFO{140.73}
& \tcellFO{114.92} & \tcellFO{125.58} & \tcellFO{128.90}
& \tcellFO{113.98} & \tcellFO{126.74} & \tcellFO{137.09}
& \tcellFO{113.72} & \tcellFO{126.35} & \tcellFO{138.42}
& \tcellFO{113.83} & \tcellFO{128.21} & \tcellFO{132.84} \\

\bottomrule
\end{tabular}
\end{adjustbox}
\vspace{-10pt}
\captionof{table}{
Throughput (Thpt, K tokens/s) and TOPS on \textbf{LLaMA 3.2 3B-Instruct} across sequence lengths.
All methods are evaluated on the same A100 40GB hardware with batch size 8, 3 warmup iterations, and 5 measurement iterations under the same model wrapper.
\textbf{One} denotes all legal score-tile groups routed to INT8.
OOM entries are shown in gray and excluded from relative color comparisons.
}
\label{tab:efficiency_llama3.2}
\end{minipage}
\end{lrbox}

\begin{table*}[p]
\centering
\usebox{\TableNineBox}
\vspace{8pt}
\usebox{\TableElevenBox}
\vspace{8pt}
\begin{minipage}{\textwidth}
\centering
\scriptsize
\setlength{\tabcolsep}{1.8pt}
\renewcommand{\arraystretch}{0.9}

\begin{adjustbox}{width=\textwidth}
\begin{tabular}{
c @{\vsep}  
c @{\vsep}  
c @{\vsep}  
c @{\vsep}  
c @{\vsep}  
c c c @{\vsep}          
c c c @{\vsep}          
c c c @{\vsep}          
c c c @{\vsep}          
c c c @{\vsep}          
c c c                     
}

\toprule
\multirow{2}{*}{\rotatebox{90}{\textbf{Len}}} &
\multirow{2}{*}{\rotatebox{90}{\scalebox{0.60}[1]{\textbf{Metric}}}} &
\multirow{2}{*}{\rotatebox{90}{\scalebox{0.68}[1]{\textbf{Torch}}}} &
\multirow{2}{*}{\rotatebox{90}{\scalebox{0.68}[1]{\textbf{Flash}}}} &
\multirow{2}{*}{\rotatebox{90}{\textbf{One}}} &
\multicolumn{3}{c}{\textbf{AlignSparse}} &
\multicolumn{3}{c}{\textbf{Band}} &
\multicolumn{3}{c}{\textbf{BigBird}} &
\multicolumn{3}{c}{\textbf{Global}} &
\multicolumn{3}{c}{\textbf{RowRand}} &
\multicolumn{3}{c}{\textbf{SpTrans}} \\

\cmidrule(lr){6-8}
\cmidrule(lr){9-11}
\cmidrule(lr){12-14}
\cmidrule(lr){15-17}
\cmidrule(lr){18-20}
\cmidrule(lr){21-23}

& & & &
& \textbf{25} & \textbf{50} & \textbf{75}
& \textbf{25} & \textbf{50} & \textbf{75}
& \textbf{25} & \textbf{50} & \textbf{75}
& \textbf{25} & \textbf{50} & \textbf{75}
& \textbf{25} & \textbf{50} & \textbf{75}
& \textbf{25} & \textbf{50} & \textbf{75} \\

\midrule

1k & Thpt
& \setlow{7.98}\setup{19.27}\tcellFO{6.98}
& \cellcolor{onered!35}7.98
& \cellcolor{baselinegreen!35}19.27
& \tcellFO{15.10} & \tcellFO{16.12} & \tcellFO{18.06}
& \tcellFO{14.28} & \tcellFO{16.98} & \tcellFO{17.59}
& \tcellFO{15.03} & \tcellFO{16.47} & \tcellFO{18.18}
& \tcellFO{14.74} & \tcellFO{16.53} & \tcellFO{17.93}
& \tcellFO{14.90} & \tcellFO{15.87} & \tcellFO{17.96}
& \tcellFO{15.09} & \tcellFO{16.05} & \tcellFO{17.76} \\

& TOPS
& \setlow{63.01}\setup{152.10}\tcellFO{55.11}
& \cellcolor{onered!35}63.01
& \cellcolor{baselinegreen!35}152.10
& \tcellFO{119.20} & \tcellFO{127.24} & \tcellFO{142.63}
& \tcellFO{112.75} & \tcellFO{134.02} & \tcellFO{138.79}
& \tcellFO{118.71} & \tcellFO{130.01} & \tcellFO{143.58}
& \tcellFO{116.37} & \tcellFO{130.48} & \tcellFO{141.49}
& \tcellFO{117.63} & \tcellFO{125.32} & \tcellFO{141.87}
& \tcellFO{119.14} & \tcellFO{126.73} & \tcellFO{140.17} \\

\midrule

2k & Thpt
& \setlow{7.72}\setup{19.13}\tcellFO{5.21}
& \cellcolor{onered!35}7.72
& \cellcolor{baselinegreen!35}19.13
& \tcellFO{14.95} & \tcellFO{15.93} & \tcellFO{17.84}
& \tcellFO{14.18} & \tcellFO{16.80} & \tcellFO{17.41}
& \tcellFO{14.88} & \tcellFO{16.23} & \tcellFO{17.94}
& \tcellFO{14.63} & \tcellFO{16.35} & \tcellFO{17.76}
& \tcellFO{14.71} & \tcellFO{15.71} & \tcellFO{17.76}
& \tcellFO{14.89} & \tcellFO{15.84} & \tcellFO{17.52} \\

& TOPS
& \setlow{62.54}\setup{155.00}\tcellFO{42.23}
& \cellcolor{onered!35}62.54
& \cellcolor{baselinegreen!35}155.00
& \tcellFO{121.07} & \tcellFO{129.05} & \tcellFO{144.61}
& \tcellFO{114.88} & \tcellFO{136.17} & \tcellFO{140.99}
& \tcellFO{120.58} & \tcellFO{131.59} & \tcellFO{145.36}
& \tcellFO{118.51} & \tcellFO{132.50} & \tcellFO{143.85}
& \tcellFO{119.23} & \tcellFO{127.32} & \tcellFO{143.91}
& \tcellFO{120.61} & \tcellFO{128.32} & \tcellFO{141.98} \\

\midrule

4k & Thpt
& \setlow{7.09}\setup{18.03}\tcellFO{OOM}
& \cellcolor{onered!35}7.09
& \cellcolor{baselinegreen!35}18.03
& \tcellFO{14.09} & \tcellFO{14.95} & \tcellFO{16.74}
& \tcellFO{13.50} & \tcellFO{15.98} & \tcellFO{16.43}
& \tcellFO{14.12} & \tcellFO{15.37} & \tcellFO{16.84}
& \tcellFO{13.89} & \tcellFO{15.43} & \tcellFO{16.65}
& \tcellFO{13.89} & \tcellFO{14.75} & \tcellFO{16.65}
& \tcellFO{14.03} & \tcellFO{14.97} & \tcellFO{16.49} \\

& TOPS
& \setlow{60.41}\setup{153.54}\tcellFO{OOM}
& \cellcolor{onered!35}60.41
& \cellcolor{baselinegreen!35}153.54
& \tcellFO{120.09} & \tcellFO{127.37} & \tcellFO{142.70}
& \tcellFO{115.00} & \tcellFO{136.10} & \tcellFO{139.94}
& \tcellFO{120.31} & \tcellFO{130.89} & \tcellFO{143.35}
& \tcellFO{118.25} & \tcellFO{131.40} & \tcellFO{141.80}
& \tcellFO{118.37} & \tcellFO{125.64} & \tcellFO{141.89}
& \tcellFO{119.51} & \tcellFO{127.56} & \tcellFO{140.44} \\

\bottomrule
\end{tabular}
\end{adjustbox}
\vspace{-10pt}
\captionof{table}{
Throughput (Thpt, K tokens/s) and TOPS on \textbf{Qwen 2.5 7B} across sequence lengths.
}
\label{tab:efficiency_qwen2.5}
\vspace{-10pt}
\end{minipage}
\end{table*}

\newsavebox{\TableThirteenBox}
\begin{lrbox}{\TableThirteenBox}
\begin{minipage}{\textwidth}
\centering
\scriptsize
\setlength{\tabcolsep}{1.8pt}
\renewcommand{\arraystretch}{0.9}

\begin{adjustbox}{width=\textwidth}
\begin{tabular}{
c @{\vsep}  
c @{\vsep}  
c @{\vsep}  
c @{\vsep}  
c @{\vsep}  
c c c @{\vsep}          
c c c @{\vsep}          
c c c @{\vsep}          
c c c @{\vsep}          
c c c @{\vsep}          
c c c                     
}

\toprule
\multirow{2}{*}{\rotatebox{90}{\textbf{Len}}} &
\multirow{2}{*}{\rotatebox{90}{\scalebox{0.60}[1]{\textbf{Metric}}}} &
\multirow{2}{*}{\rotatebox{90}{\scalebox{0.68}[1]{\textbf{Torch}}}} &
\multirow{2}{*}{\rotatebox{90}{\scalebox{0.68}[1]{\textbf{Flash}}}} &
\multirow{2}{*}{\rotatebox{90}{\textbf{One}}} &
\multicolumn{3}{c}{\textbf{AlignSparse}} &
\multicolumn{3}{c}{\textbf{Band}} &
\multicolumn{3}{c}{\textbf{BigBird}} &
\multicolumn{3}{c}{\textbf{Global}} &
\multicolumn{3}{c}{\textbf{RowRand}} &
\multicolumn{3}{c}{\textbf{SpTrans}} \\

\cmidrule(lr){6-8}
\cmidrule(lr){9-11}
\cmidrule(lr){12-14}
\cmidrule(lr){15-17}
\cmidrule(lr){18-20}
\cmidrule(lr){21-23}

& & & &
& \textbf{25} & \textbf{50} & \textbf{75}
& \textbf{25} & \textbf{50} & \textbf{75}
& \textbf{25} & \textbf{50} & \textbf{75}
& \textbf{25} & \textbf{50} & \textbf{75}
& \textbf{25} & \textbf{50} & \textbf{75}
& \textbf{25} & \textbf{50} & \textbf{75} \\

\midrule

1k & Thpt
& \setlow{8.21}\setup{19.78}\tcellFO{7.16}
& \cellcolor{onered!35}8.21
& \cellcolor{baselinegreen!35}19.78
& \tcellFO{16.20} & \tcellFO{19.82} & \tcellFO{18.04}
& \tcellFO{16.12} & \tcellFO{17.93} & \tcellFO{17.95}
& \tcellFO{16.23} & \tcellFO{18.09} & \tcellFO{18.07}
& \tcellFO{16.20} & \tcellFO{18.04} & \tcellFO{20.56}
& \tcellFO{16.25} & \tcellFO{18.02} & \tcellFO{17.98}
& \tcellFO{16.25} & \tcellFO{20.26} & \tcellFO{20.14} \\

& TOPS
& \setlow{64.81}\setup{156.18}\tcellFO{56.51}
& \cellcolor{onered!35}64.81
& \cellcolor{baselinegreen!35}156.18
& \tcellFO{127.94} & \tcellFO{156.43} & \tcellFO{142.38}
& \tcellFO{127.26} & \tcellFO{141.56} & \tcellFO{141.74}
& \tcellFO{128.19} & \tcellFO{142.87} & \tcellFO{142.69}
& \tcellFO{127.95} & \tcellFO{142.38} & \tcellFO{162.36}
& \tcellFO{128.23} & \tcellFO{142.28} & \tcellFO{141.99}
& \tcellFO{128.34} & \tcellFO{159.91} & \tcellFO{159.07} \\

\midrule

2k & Thpt
& \setlow{7.89}\setup{19.56}\tcellFO{5.35}
& \cellcolor{onered!35}7.89
& \cellcolor{baselinegreen!35}19.56
& \tcellFO{15.99} & \tcellFO{19.56} & \tcellFO{17.77}
& \tcellFO{15.89} & \tcellFO{18.52} & \tcellFO{19.29}
& \tcellFO{16.04} & \tcellFO{17.86} & \tcellFO{17.78}
& \tcellFO{16.02} & \tcellFO{17.78} & \tcellFO{20.29}
& \tcellFO{16.01} & \tcellFO{17.77} & \tcellFO{17.77}
& \tcellFO{16.01} & \tcellFO{19.86} & \tcellFO{19.76} \\

& TOPS
& \setlow{63.93}\setup{158.52}\tcellFO{43.34}
& \cellcolor{onered!35}63.93
& \cellcolor{baselinegreen!35}158.52
& \tcellFO{129.50} & \tcellFO{158.47} & \tcellFO{143.92}
& \tcellFO{128.76} & \tcellFO{150.08} & \tcellFO{156.33}
& \tcellFO{129.92} & \tcellFO{144.62} & \tcellFO{144.07}
& \tcellFO{129.79} & \tcellFO{144.14} & \tcellFO{164.38}
& \tcellFO{129.65} & \tcellFO{144.00} & \tcellFO{144.00}
& \tcellFO{129.76} & \tcellFO{160.98} & \tcellFO{160.15} \\

\midrule

4k & Thpt
& \setlow{7.22}\setup{18.49}\tcellFO{OOM}
& \cellcolor{onered!35}7.22
& \cellcolor{baselinegreen!35}18.49
& \tcellFO{15.17} & \tcellFO{16.80} & \tcellFO{16.80}
& \tcellFO{15.17} & \tcellFO{17.62} & \tcellFO{18.31}
& \tcellFO{15.26} & \tcellFO{16.91} & \tcellFO{16.80}
& \tcellFO{15.25} & \tcellFO{16.91} & \tcellFO{19.23}
& \tcellFO{15.22} & \tcellFO{16.85} & \tcellFO{16.82}
& \tcellFO{15.07} & \tcellFO{18.76} & \tcellFO{18.66} \\

& TOPS
& \setlow{61.48}\setup{157.51}\tcellFO{OOM}
& \cellcolor{onered!35}61.48
& \cellcolor{baselinegreen!35}157.51
& \tcellFO{129.19} & \tcellFO{143.03} & \tcellFO{143.12}
& \tcellFO{129.21} & \tcellFO{150.04} & \tcellFO{156.04}
& \tcellFO{130.00} & \tcellFO{144.09} & \tcellFO{143.19}
& \tcellFO{129.94} & \tcellFO{144.05} & \tcellFO{163.77}
& \tcellFO{129.58} & \tcellFO{143.53} & \tcellFO{143.21}
& \tcellFO{128.32} & \tcellFO{159.69} & \tcellFO{158.85} \\

\bottomrule
\end{tabular}
\end{adjustbox}
\captionof{table}{
Throughput (Thpt, K tokens/s) and TOPS on \textbf{Qwen 2 7B} across sequence lengths.
}
\label{tab:efficiency_qwen2}
\end{minipage}
\end{lrbox}

\begin{table*}[p]
\centering
\usebox{\TableTenBox}
\vspace{8pt}
\usebox{\TableThirteenBox}
\vspace{8pt}
\begin{minipage}{\textwidth}
\centering
\scriptsize
\setlength{\tabcolsep}{1.8pt}
\renewcommand{\arraystretch}{0.9}
\begin{adjustbox}{width=\textwidth}
\begin{tabular}{
c @{\vsep}  
c @{\vsep}  
c @{\vsep}  
c @{\vsep}  
c @{\vsep}  
c c c @{\vsep}          
c c c @{\vsep}          
c c c @{\vsep}          
c c c @{\vsep}          
c c c @{\vsep}          
c c c                     
}

\toprule
\multirow{2}{*}{\rotatebox{90}{\textbf{Len}}} &
\multirow{2}{*}{\rotatebox{90}{\scalebox{0.60}[1]{\textbf{Metric}}}} &
\multirow{2}{*}{\rotatebox{90}{\scalebox{0.68}[1]{\textbf{Torch}}}} &
\multirow{2}{*}{\rotatebox{90}{\scalebox{0.68}[1]{\textbf{Flash}}}} &
\multirow{2}{*}{\rotatebox{90}{\textbf{One}}} &
\multicolumn{3}{c}{\textbf{AlignSparse}} &
\multicolumn{3}{c}{\textbf{Band}} &
\multicolumn{3}{c}{\textbf{BigBird}} &
\multicolumn{3}{c}{\textbf{Global}} &
\multicolumn{3}{c}{\textbf{RowRand}} &
\multicolumn{3}{c}{\textbf{SpTrans}} \\

\cmidrule(lr){6-8}
\cmidrule(lr){9-11}
\cmidrule(lr){12-14}
\cmidrule(lr){15-17}
\cmidrule(lr){18-20}
\cmidrule(lr){21-23}

& & & &
& \textbf{25} & \textbf{50} & \textbf{75}
& \textbf{25} & \textbf{50} & \textbf{75}
& \textbf{25} & \textbf{50} & \textbf{75}
& \textbf{25} & \textbf{50} & \textbf{75}
& \textbf{25} & \textbf{50} & \textbf{75}
& \textbf{25} & \textbf{50} & \textbf{75} \\

\midrule

1k & Thpt
& \setlow{8.40}\setup{18.13}\tcellFO{6.52}
& \cellcolor{onered!35}8.40
& \cellcolor{baselinegreen!35}18.13
& \tcellFO{15.54} & \tcellFO{16.71} & \tcellFO{18.02}
& \tcellFO{15.38} & \tcellFO{16.71} & \tcellFO{18.02}
& \tcellFO{15.32} & \tcellFO{16.48} & \tcellFO{17.88}
& \tcellFO{15.23} & \tcellFO{16.38} & \tcellFO{18.36}
& \tcellFO{15.08} & \tcellFO{16.54} & \tcellFO{17.83}
& \tcellFO{15.01} & \tcellFO{16.15} & \tcellFO{17.52} \\

& TOPS
& \setlow{66.30}\setup{124.65}\tcellFO{44.86}
& \cellcolor{onered!35}66.30
& \cellcolor{baselinegreen!35}124.65
& \tcellFO{106.85} & \tcellFO{114.97} & \tcellFO{123.98}
& \tcellFO{105.80} & \tcellFO{114.97} & \tcellFO{123.98}
& \tcellFO{105.39} & \tcellFO{113.33} & \tcellFO{122.93}
& \tcellFO{104.72} & \tcellFO{112.67} & \tcellFO{126.36}
& \tcellFO{103.68} & \tcellFO{113.78} & \tcellFO{122.71}
& \tcellFO{103.28} & \tcellFO{111.07} & \tcellFO{120.47} \\

\midrule

2k & Thpt
& \setlow{8.07}\setup{17.35}\tcellFO{4.60}
& \cellcolor{onered!35}8.07
& \cellcolor{baselinegreen!35}17.35
& \tcellFO{14.91} & \tcellFO{16.05} & \tcellFO{17.30}
& \tcellFO{14.76} & \tcellFO{16.05} & \tcellFO{17.30}
& \tcellFO{14.66} & \tcellFO{15.88} & \tcellFO{17.11}
& \tcellFO{14.61} & \tcellFO{15.73} & \tcellFO{17.64}
& \tcellFO{14.46} & \tcellFO{15.89} & \tcellFO{17.13}
& \tcellFO{14.37} & \tcellFO{15.56} & \tcellFO{16.77} \\

& TOPS
& \setlow{65.40}\setup{124.02}\tcellFO{32.88}
& \cellcolor{onered!35}65.40
& \cellcolor{baselinegreen!35}124.02
& \tcellFO{106.64} & \tcellFO{114.82} & \tcellFO{123.67}
& \tcellFO{105.58} & \tcellFO{114.82} & \tcellFO{123.67}
& \tcellFO{104.81} & \tcellFO{113.53} & \tcellFO{122.37}
& \tcellFO{104.51} & \tcellFO{112.53} & \tcellFO{126.05}
& \tcellFO{103.47} & \tcellFO{113.64} & \tcellFO{122.40}
& \tcellFO{102.72} & \tcellFO{111.26} & \tcellFO{119.93} \\

\midrule

4k & Thpt
& \setlow{7.39}\setup{15.80}\tcellFO{OOM}
& \cellcolor{onered!35}7.39
& \cellcolor{baselinegreen!35}15.80
& \tcellFO{13.78} & \tcellFO{14.80} & \tcellFO{15.92}
& \tcellFO{13.64} & \tcellFO{14.80} & \tcellFO{15.92}
& \tcellFO{13.51} & \tcellFO{14.67} & \tcellFO{15.66}
& \tcellFO{13.51} & \tcellFO{14.51} & \tcellFO{16.22}
& \tcellFO{13.37} & \tcellFO{14.65} & \tcellFO{15.75}
& \tcellFO{13.24} & \tcellFO{14.38} & \tcellFO{15.35} \\

& TOPS
& \setlow{62.89}\setup{121.58}\tcellFO{OOM}
& \cellcolor{onered!35}62.89
& \cellcolor{baselinegreen!35}121.58
& \tcellFO{106.03} & \tcellFO{113.93} & \tcellFO{122.50}
& \tcellFO{104.98} & \tcellFO{113.93} & \tcellFO{122.50}
& \tcellFO{103.93} & \tcellFO{112.87} & \tcellFO{120.48}
& \tcellFO{103.91} & \tcellFO{111.65} & \tcellFO{124.85}
& \tcellFO{102.88} & \tcellFO{112.76} & \tcellFO{121.24}
& \tcellFO{101.85} & \tcellFO{110.62} & \tcellFO{118.07} \\

\bottomrule
\end{tabular}
\end{adjustbox}
\vspace{-5pt}
\captionof{table}{
Throughput (Thpt, K tokens/s) and TOPS on \textbf{Vicuna 7B} across sequence lengths.
}
\label{tab:efficiency_vicuna}
\end{minipage}
\end{table*}

Across models and lengths, Torch provides the lowest throughput and exhibits earlier OOM behavior, while FlashAttention serves as a strong baseline when it fits in memory.
\textbf{One} provides the uniform INT8 reference, while mixed-routing layouts form nearby empirical operating points whose ordering depends on complete-pipeline execution.
Across models, higher INT8 coverage generally increases throughput and TOPS, and several mixed-routing configurations match or exceed \textbf{One}.
Differences among layouts reflect complete-pipeline dispatch, rescaling, memory-access, and scheduling behavior under different spatial routing arrangements; all layouts preserve the same dense legal connectivity.
Overall, the results suggest that \modelname provides a practical accuracy-efficiency knob: higher INT8 ratios increase low-precision execution, while more conservative ratios preserve more FP16-routed tile groups for layouts or tasks that are more accuracy-sensitive.

\newsavebox{\TableFifteenBox}
\begin{lrbox}{\TableFifteenBox}
\begin{minipage}{\textwidth}
\centering
\small
\setlength{\tabcolsep}{4pt}
\renewcommand{\arraystretch}{1.05}

\begin{tabular}{c @{\vsep}
                c @{\vsep}
                c @{\vsep}
                c @{\vsep}
                c @{\vsep}
                c @{\vsep}
                c}
\toprule
\textbf{Seq Len} &
\textbf{\small{Torch Value}} &
\textbf{Flash} &
\textbf{100\% INT8} &
\textbf{75\% INT8} &
\textbf{50\% INT8} &
\textbf{25\% INT8} \\
\midrule
1024 & $2.85\!\times\!10^{0}$ & $1.57\!\times\!10^{-2}$ & $5.40\!\times\!10^{-2}$ & $3.54\!\times\!10^{-2}$ & $1.68\!\times\!10^{-2}$ & $1.95\!\times\!10^{-3}$ \\
2048 & $2.82\!\times\!10^{0}$ & $1.42\!\times\!10^{-2}$ & $6.49\!\times\!10^{-2}$ & $4.25\!\times\!10^{-2}$ & $2.20\!\times\!10^{-2}$ & $1.95\!\times\!10^{-3}$ \\
4096 & $2.98\!\times\!10^{0}$ & $1.12\!\times\!10^{-2}$ & $5.86\!\times\!10^{-2}$ & $4.00\!\times\!10^{-2}$ & $2.00\!\times\!10^{-2}$ & $1.95\!\times\!10^{-3}$ \\
8192 & $3.46\!\times\!10^{0}$ & $1.22\!\times\!10^{-2}$ & $6.25\!\times\!10^{-2}$ & $4.10\!\times\!10^{-2}$ & $2.10\!\times\!10^{-2}$ & $1.95\!\times\!10^{-3}$ \\
\bottomrule
\end{tabular}
\captionof{table}{
Single-layer model on random inputs.
Torch Value reports the maximum absolute logit magnitude under the fixed Torch FP16 reference; all other columns report maximum absolute deviation from this reference.
}
\label{tab:num_single_layer_logits}
\end{minipage}
\end{lrbox}

\newsavebox{\TableSixteenBox}
\begin{lrbox}{\TableSixteenBox}
\begin{minipage}{\textwidth}
\centering
\small
\setlength{\tabcolsep}{4pt}
\renewcommand{\arraystretch}{1.05}

\begin{tabular}{c @{\vsep}
                c @{\vsep}
                c @{\vsep}
                c @{\vsep}
                c @{\vsep}
                c @{\vsep}
                c}
\toprule
\textbf{Seq Len} &
\textbf{\small{Torch Value}} &
\textbf{Flash} &
\textbf{100\% INT8} &
\textbf{75\% INT8} &
\textbf{50\% INT8} &
\textbf{25\% INT8} \\
\midrule
1024 & $2.91\!\times\!10^{0}$ & $1.48\!\times\!10^{-1}$ & $2.03\!\times\!10^{-1}$ & $1.34\!\times\!10^{-1}$ & $6.93\!\times\!10^{-2}$ & $3.91\!\times\!10^{-3}$ \\
2048 & $2.79\!\times\!10^{0}$ & $1.40\!\times\!10^{-1}$ & $1.75\!\times\!10^{-1}$ & $1.23\!\times\!10^{-1}$ & $6.40\!\times\!10^{-2}$ & $3.17\!\times\!10^{-3}$ \\
4096 & $3.05\!\times\!10^{0}$ & $1.47\!\times\!10^{-1}$ & $2.19\!\times\!10^{-1}$ & $1.46\!\times\!10^{-1}$ & $7.42\!\times\!10^{-2}$ & $3.42\!\times\!10^{-3}$ \\
8192 & $3.22\!\times\!10^{0}$ & $1.52\!\times\!10^{-1}$ & $2.22\!\times\!10^{-1}$ & $1.48\!\times\!10^{-1}$ & $7.25\!\times\!10^{-2}$ & $3.91\!\times\!10^{-3}$ \\
\bottomrule
\end{tabular}
\captionof{table}{
Numerical behavior of a 12-layer attention model on random inputs.
Torch Value reports the maximum absolute logit magnitude under the selected Torch FP16 reference; all other columns report maximum absolute deviation from this reference.
}
\label{tab:num_12layer_logits}
\end{minipage}
\end{lrbox}

\begin{table*}[t]
\centering
\usebox{\TableFifteenBox}
\vspace{8pt}
\usebox{\TableSixteenBox}
\vspace{8pt}
\begin{minipage}{\textwidth}
\centering
\small
\setlength{\tabcolsep}{4pt}
\renewcommand{\arraystretch}{1.05}

\begin{tabular}{c @{\vsep}
                c @{\vsep}
                c @{\vsep}
                c @{\vsep}
                c @{\vsep}
                c @{\vsep}
                c}
\toprule
\textbf{Seq Len} &
\textbf{\small{Torch Value}} &
\textbf{FlashAttn} &
\textbf{100\% INT8} &
\textbf{75\% INT8} &
\textbf{50\% INT8} &
\textbf{25\% INT8} \\
\midrule
1024 & $3.02\!\times\!10^{0}$ & $3.55\!\times\!10^{-1}$ & $3.56\!\times\!10^{-1}$ & $2.22\!\times\!10^{-1}$ & $1.14\!\times\!10^{-1}$ & $5.86\!\times\!10^{-3}$ \\
2048 & $2.90\!\times\!10^{0}$ & $3.18\!\times\!10^{-1}$ & $4.48\!\times\!10^{-1}$ & $2.96\!\times\!10^{-1}$ & $1.37\!\times\!10^{-1}$ & $5.00\!\times\!10^{-3}$ \\
4096 & $3.13\!\times\!10^{0}$ & $3.79\!\times\!10^{-1}$ & $3.69\!\times\!10^{-1}$ & $2.51\!\times\!10^{-1}$ & $1.28\!\times\!10^{-1}$ & $5.86\!\times\!10^{-3}$ \\
8192 & $3.21\!\times\!10^{0}$ & $3.31\!\times\!10^{-1}$ & $3.98\!\times\!10^{-1}$ & $2.60\!\times\!10^{-1}$ & $1.34\!\times\!10^{-1}$ & $5.86\!\times\!10^{-3}$ \\
\bottomrule
\end{tabular}
\captionof{table}{
Numerical behavior of a 32-layer attention model on random inputs.
Torch Value reports the maximum absolute logit magnitude under the selected Torch FP16 reference; all other columns report maximum absolute deviation from this reference.
}
\label{tab:num_32layer_logits}
\end{minipage}
\end{table*}

\newcommand{\TableEighteenBlock}{%
\begin{minipage}{\textwidth}
\centering
\small
\setlength{\tabcolsep}{11pt}
\renewcommand{\arraystretch}{1.05}
\begin{tabular}{lcc @{\vsep} lcc}
\toprule
\textbf{Config} & \textbf{MaxDiff} & \textbf{MeanDiff} &
\textbf{Config} & \textbf{MaxDiff} & \textbf{MeanDiff} \\
\midrule
Zero (0\% INT8) & $1.96{\times}10^{-4}$ & $5.25{\times}10^{-6}$ &
BigBird75 & $1.96{\times}10^{-4}$ & $8.45{\times}10^{-6}$ \\
One (100\% INT8) & $1.96{\times}10^{-4}$ & $5.25{\times}10^{-6}$ &
Global0 & $1.96{\times}10^{-4}$ & $5.25{\times}10^{-6}$ \\
AlignSparse0 & $1.96{\times}10^{-4}$ & $5.25{\times}10^{-6}$ &
Global25 & $1.96{\times}10^{-4}$ & $6.75{\times}10^{-6}$ \\
AlignSparse25 & $1.96{\times}10^{-4}$ & $7.01{\times}10^{-6}$ &
Global50 & $1.96{\times}10^{-4}$ & $8.01{\times}10^{-6}$ \\
AlignSparse50 & $1.96{\times}10^{-4}$ & $7.23{\times}10^{-6}$ &
Global75 & $1.96{\times}10^{-4}$ & $8.41{\times}10^{-6}$ \\
AlignSparse75 & $1.96{\times}10^{-4}$ & $7.93{\times}10^{-6}$ &
RowRand0 & $1.96{\times}10^{-4}$ & $5.25{\times}10^{-6}$ \\
Band0 & $1.96{\times}10^{-4}$ & $5.25{\times}10^{-6}$ &
RowRand25 & $1.96{\times}10^{-4}$ & $6.62{\times}10^{-6}$ \\
Band25 & $1.96{\times}10^{-4}$ & $6.52{\times}10^{-6}$ &
RowRand50 & $1.96{\times}10^{-4}$ & $8.07{\times}10^{-6}$ \\
Band50 & $1.96{\times}10^{-4}$ & $7.67{\times}10^{-6}$ &
RowRand75 & $1.96{\times}10^{-4}$ & $8.45{\times}10^{-6}$ \\
Band75 & $4.84{\times}10^{-4}$ & $8.88{\times}10^{-6}$ &
SpTrans0 & $1.96{\times}10^{-4}$ & $5.25{\times}10^{-6}$ \\
BigBird0 & $1.96{\times}10^{-4}$ & $5.25{\times}10^{-6}$ &
SpTrans25 & $1.96{\times}10^{-4}$ & $6.33{\times}10^{-6}$ \\
BigBird25 & $1.96{\times}10^{-4}$ & $6.62{\times}10^{-6}$ &
SpTrans50 & $1.96{\times}10^{-4}$ & $6.84{\times}10^{-6}$ \\
BigBird50 & $1.96{\times}10^{-4}$ & $8.07{\times}10^{-6}$ &
SpTrans75 & $4.84{\times}10^{-4}$ & $8.68{\times}10^{-6}$ \\
\bottomrule
\end{tabular}
\captionof{table}{
Direct numerical difference between \modelname and FlashAttention under different precision layouts.
A nonzero gap exists even at 0\% INT8 due to fused-kernel implementation differences.
}
\label{tab:tilemix_vs_flash_numerical}
\end{minipage}%
}

\begin{table*}[t]
\centering
\TableEighteenBlock
\vspace{12pt}
\begin{minipage}{\textwidth}
\centering
\scriptsize
\setlength{\tabcolsep}{10pt}
\renewcommand{\arraystretch}{1.05}
\begin{tabular}{lcc @{\vsep} lcc}
\toprule
\textbf{Config} & \textbf{MaxAbsDiff} & \textbf{MeanAbsDiff} &
\textbf{Config} & \textbf{MaxAbsDiff} & \textbf{MeanAbsDiff} \\
\midrule
One (fp16acc) & $6.59{\times}10^{-1}$ & $6.99{\times}10^{-3}$ &
Global25 (fp16acc) & $1.24{\times}10^{-1}$ & $2.44{\times}10^{-3}$ \\
One (fp32acc) & $6.59{\times}10^{-1}$ & $6.99{\times}10^{-3}$ &
Global25 (fp32acc) & $1.24{\times}10^{-1}$ & $2.44{\times}10^{-3}$ \\
FP16 (fp16acc) & $2.93{\times}10^{-4}$ & $1.02{\times}10^{-5}$ &
Global50 (fp16acc) & $1.80{\times}10^{-1}$ & $4.29{\times}10^{-3}$ \\
FP16 (fp32acc) & $0$ & $0$ &
Global50 (fp32acc) & $1.80{\times}10^{-1}$ & $4.28{\times}10^{-3}$ \\
AlignSparse25 (fp16acc) & $2.38{\times}10^{-1}$ & $2.71{\times}10^{-3}$ &
Global75 (fp16acc) & $1.80{\times}10^{-1}$ & $4.64{\times}10^{-3}$ \\
AlignSparse25 (fp32acc) & $2.38{\times}10^{-1}$ & $2.71{\times}10^{-3}$ &
Global75 (fp32acc) & $1.80{\times}10^{-1}$ & $4.64{\times}10^{-3}$ \\
AlignSparse50 (fp16acc) & $2.38{\times}10^{-1}$ & $2.92{\times}10^{-3}$ &
RowRand25 (fp16acc) & $1.56{\times}10^{-1}$ & $2.39{\times}10^{-3}$ \\
AlignSparse50 (fp32acc) & $2.38{\times}10^{-1}$ & $2.91{\times}10^{-3}$ &
RowRand25 (fp32acc) & $1.56{\times}10^{-1}$ & $2.38{\times}10^{-3}$ \\
AlignSparse75 (fp16acc) & $2.38{\times}10^{-1}$ & $3.82{\times}10^{-3}$ &
RowRand50 (fp16acc) & $2.27{\times}10^{-1}$ & $4.34{\times}10^{-3}$ \\
AlignSparse75 (fp32acc) & $2.38{\times}10^{-1}$ & $3.81{\times}10^{-3}$ &
RowRand50 (fp32acc) & $2.27{\times}10^{-1}$ & $4.34{\times}10^{-3}$ \\
Band25 (fp16acc) & $1.10{\times}10^{-1}$ & $1.91{\times}10^{-3}$ &
RowRand75 (fp16acc) & $2.12{\times}10^{-1}$ & $4.60{\times}10^{-3}$ \\
Band25 (fp32acc) & $1.10{\times}10^{-1}$ & $1.90{\times}10^{-3}$ &
RowRand75 (fp32acc) & $2.12{\times}10^{-1}$ & $4.60{\times}10^{-3}$ \\
Band50 (fp16acc) & $2.38{\times}10^{-1}$ & $3.62{\times}10^{-3}$ &
SpTrans25 (fp16acc) & $1.40{\times}10^{-1}$ & $1.73{\times}10^{-3}$ \\
Band50 (fp32acc) & $2.38{\times}10^{-1}$ & $3.61{\times}10^{-3}$ &
SpTrans25 (fp32acc) & $1.41{\times}10^{-1}$ & $1.72{\times}10^{-3}$ \\
Band75 (fp16acc) & $6.59{\times}10^{-1}$ & $5.21{\times}10^{-3}$ &
SpTrans50 (fp16acc) & $1.40{\times}10^{-1}$ & $2.50{\times}10^{-3}$ \\
Band75 (fp32acc) & $6.59{\times}10^{-1}$ & $5.21{\times}10^{-3}$ &
SpTrans50 (fp32acc) & $1.41{\times}10^{-1}$ & $2.49{\times}10^{-3}$ \\
BigBird25 (fp16acc) & $1.56{\times}10^{-1}$ & $2.39{\times}10^{-3}$ &
SpTrans75 (fp16acc) & $6.59{\times}10^{-1}$ & $5.09{\times}10^{-3}$ \\
BigBird25 (fp32acc) & $1.56{\times}10^{-1}$ & $2.38{\times}10^{-3}$ &
SpTrans75 (fp32acc) & $6.59{\times}10^{-1}$ & $5.09{\times}10^{-3}$ \\
BigBird50 (fp16acc) & $2.27{\times}10^{-1}$ & $4.34{\times}10^{-3}$ &
BigBird75 (fp16acc) & $2.12{\times}10^{-1}$ & $4.60{\times}10^{-3}$ \\
BigBird50 (fp32acc) & $2.27{\times}10^{-1}$ & $4.34{\times}10^{-3}$ &
BigBird75 (fp32acc) & $2.12{\times}10^{-1}$ & $4.60{\times}10^{-3}$ \\
\bottomrule
\end{tabular}
\captionof{table}{
Numerical differences compared to full FP16 with FP32 accumulation under different precision layouts and mixing ratios.
Each configuration is evaluated with FP16 accumulation and FP32 accumulation.
}
\label{tab:pattern_accumulation_numerical}
\end{minipage}
\end{table*}

\newcommand{\TableTwentyBlock}{%
\begin{minipage}{\textwidth}
\centering
\small
\setlength{\tabcolsep}{8pt}
\renewcommand{\arraystretch}{1.05}
\begin{tabular}{lcc @{\vsep} lcc}
\toprule
\textbf{Config} & \textbf{MaxAbsDiff} & \textbf{MeanAbsDiff} &
\textbf{Config} & \textbf{MaxAbsDiff} & \textbf{MeanAbsDiff} \\
\midrule
One & $3.91{\times}10^{-4}$ & $1.51{\times}10^{-5}$ &
Global25 & $2.93{\times}10^{-4}$ & $1.21{\times}10^{-5}$ \\
FP16 & $2.93{\times}10^{-4}$ & $1.02{\times}10^{-5}$ &
Global50 & $3.91{\times}10^{-4}$ & $1.36{\times}10^{-5}$ \\
AlignSparse25 & $3.91{\times}10^{-4}$ & $1.24{\times}10^{-5}$ &
Global75 & $3.91{\times}10^{-4}$ & $1.44{\times}10^{-5}$ \\
AlignSparse50 & $3.91{\times}10^{-4}$ & $1.27{\times}10^{-5}$ &
RowRand25 & $3.91{\times}10^{-4}$ & $1.17{\times}10^{-5}$ \\
AlignSparse75 & $2.93{\times}10^{-4}$ & $1.37{\times}10^{-5}$ &
RowRand50 & $3.91{\times}10^{-4}$ & $1.37{\times}10^{-5}$ \\
Band25 & $3.91{\times}10^{-4}$ & $1.19{\times}10^{-5}$ &
RowRand75 & $3.91{\times}10^{-4}$ & $1.46{\times}10^{-5}$ \\
Band50 & $3.91{\times}10^{-4}$ & $1.32{\times}10^{-5}$ &
SpTrans25 & $3.91{\times}10^{-4}$ & $1.15{\times}10^{-5}$ \\
Band75 & $3.42{\times}10^{-4}$ & $1.41{\times}10^{-5}$ &
SpTrans50 & $4.39{\times}10^{-4}$ & $1.21{\times}10^{-5}$ \\
BigBird25 & $3.91{\times}10^{-4}$ & $1.17{\times}10^{-5}$ &
SpTrans75 & $4.39{\times}10^{-4}$ & $1.37{\times}10^{-5}$ \\
BigBird50 & $3.91{\times}10^{-4}$ & $1.37{\times}10^{-5}$ &
BigBird75 & $3.91{\times}10^{-4}$ & $1.46{\times}10^{-5}$ \\
\bottomrule
\end{tabular}
\captionof{table}{
Direct comparison between FP16 and FP32 accumulation under different precision layouts and mixing ratios.
}
\label{tab:fp16_fp32_accum_direct}
\end{minipage}%
}

\begin{table*}[t]
\centering
\captionsetup{skip=2pt}
\TableTwentyBlock
\vspace{-2pt}
\begin{minipage}{\textwidth}
\centering
\scriptsize
\setlength{\tabcolsep}{4pt}
\renewcommand{\arraystretch}{0.95}
\begin{adjustbox}{width=\textwidth}
\begin{tabular}{c c c @{\vsep} c c c @{\vsep} c c c @{\vsep} c c c}
\toprule
\textbf{L} & \textbf{Max} & \textbf{Mean} &
\textbf{L} & \textbf{Max} & \textbf{Mean} &
\textbf{L} & \textbf{Max} & \textbf{Mean} &
\textbf{L} & \textbf{Max} & \textbf{Mean} \\
\midrule
0  & $0.00$ & $0.00$ & 8  & $1.09{\times}10^{-2}$ & $8.17{\times}10^{-4}$ & 16 & $1.10{\times}10^{-2}$ & $1.47{\times}10^{-3}$ & 24 & $1.11{\times}10^{-2}$ & $3.52{\times}10^{-3}$ \\
1  & $1.08{\times}10^{-5}$ & $1.75{\times}10^{-6}$ & 9  & $1.09{\times}10^{-2}$ & $9.81{\times}10^{-4}$ & 17 & $1.10{\times}10^{-2}$ & $1.62{\times}10^{-3}$ & 25 & $1.11{\times}10^{-2}$ & $3.96{\times}10^{-3}$ \\
2  & $1.08{\times}10^{-5}$ & $3.91{\times}10^{-6}$ & 10 & $1.10{\times}10^{-2}$ & $1.06{\times}10^{-3}$ & 18 & $1.10{\times}10^{-2}$ & $1.81{\times}10^{-3}$ & 26 & $1.11{\times}10^{-2}$ & $4.39{\times}10^{-4}$ \\
3  & $1.09{\times}10^{-5}$ & $5.58{\times}10^{-6}$ & 11 & $1.10{\times}10^{-2}$ & $1.06{\times}10^{-3}$ & 19 & $1.11{\times}10^{-2}$ & $2.11{\times}10^{-3}$ & 27 & $1.11{\times}10^{-2}$ & $4.98{\times}10^{-4}$ \\
4  & $1.09{\times}10^{-5}$ & $6.42{\times}10^{-6}$ & 12 & $1.10{\times}10^{-2}$ & $1.10{\times}10^{-3}$ & 20 & $1.11{\times}10^{-2}$ & $2.31{\times}10^{-3}$ & 28 & $1.11{\times}10^{-2}$ & $5.54{\times}10^{-4}$ \\
5  & $1.09{\times}10^{-5}$ & $6.95{\times}10^{-6}$ & 13 & $1.10{\times}10^{-2}$ & $1.11{\times}10^{-3}$ & 21 & $1.11{\times}10^{-2}$ & $2.60{\times}10^{-3}$ & 29 & $1.11{\times}10^{-2}$ & $6.65{\times}10^{-4}$ \\
6  & $1.09{\times}10^{-5}$ & $7.20{\times}10^{-6}$ & 14 & $1.10{\times}10^{-2}$ & $1.20{\times}10^{-3}$ & 22 & $1.11{\times}10^{-2}$ & $2.82{\times}10^{-3}$ & 30 & $1.16{\times}10^{-2}$ & $8.42{\times}10^{-4}$ \\
7  & $1.09{\times}10^{-5}$ & $7.51{\times}10^{-6}$ & 15 & $1.11{\times}10^{-2}$ & $1.35{\times}10^{-3}$ & 23 & $1.11{\times}10^{-2}$ & $3.14{\times}10^{-3}$ & 31 & $1.94{\times}10^{-2}$ & $1.27{\times}10^{-5}$ \\
\bottomrule
\end{tabular}
\end{adjustbox}
\captionof{table}{
Layer-wise numerical differences between FP16 and FP32 accumulation on \textbf{LLaMA 3.1 8B}.
Max and Mean report the maximum and mean absolute differences at each layer.
}
\label{tab:accum_llama31_8b}
\end{minipage}
\vspace{-2pt}

\begin{minipage}{\textwidth}
\centering
\scriptsize
\setlength{\tabcolsep}{4pt}
\renewcommand{\arraystretch}{0.95}
\begin{adjustbox}{width=\textwidth}
\begin{tabular}{c c c @{\vsep} c c c @{\vsep} c c c @{\vsep} c c c}
\toprule
\textbf{L} & \textbf{Max} & \textbf{Mean} &
\textbf{L} & \textbf{Max} & \textbf{Mean} &
\textbf{L} & \textbf{Max} & \textbf{Mean} &
\textbf{L} & \textbf{Max} & \textbf{Mean} \\
\midrule
0  & $0.00$ & $0.00$ & 12 & $5.63{\times}10^{-2}$ & $1.07{\times}10^{-4}$ & 24 & $5.82{\times}10^{-2}$ & $1.18{\times}10^{-4}$ & 36 & $5.97{\times}10^{-2}$ & $1.29{\times}10^{-4}$ \\
1  & $3.23{\times}10^{-3}$ & $2.30{\times}10^{-5}$ & 13 & $5.63{\times}10^{-2}$ & $1.08{\times}10^{-4}$ & 25 & $5.82{\times}10^{-2}$ & $1.19{\times}10^{-4}$ & 37 & $5.95{\times}10^{-2}$ & $1.31{\times}10^{-4}$ \\
2  & $6.38{\times}10^{-3}$ & $3.50{\times}10^{-5}$ & 14 & $5.62{\times}10^{-2}$ & $1.08{\times}10^{-4}$ & 26 & $5.88{\times}10^{-2}$ & $1.19{\times}10^{-4}$ & 38 & $5.92{\times}10^{-2}$ & $1.34{\times}10^{-4}$ \\
3  & $9.78{\times}10^{-3}$ & $5.82{\times}10^{-5}$ & 15 & $5.62{\times}10^{-2}$ & $1.09{\times}10^{-4}$ & 27 & $5.90{\times}10^{-2}$ & $1.20{\times}10^{-4}$ & 39 & $5.88{\times}10^{-2}$ & $1.35{\times}10^{-4}$ \\
4  & $5.10{\times}10^{-2}$ & $7.92{\times}10^{-5}$ & 16 & $5.62{\times}10^{-2}$ & $1.09{\times}10^{-4}$ & 28 & $5.94{\times}10^{-2}$ & $1.21{\times}10^{-4}$ & 40 & $5.82{\times}10^{-2}$ & $1.37{\times}10^{-4}$ \\
5  & $5.43{\times}10^{-2}$ & $8.40{\times}10^{-5}$ & 17 & $5.63{\times}10^{-2}$ & $1.10{\times}10^{-4}$ & 29 & $5.96{\times}10^{-2}$ & $1.21{\times}10^{-4}$ & 41 & $5.76{\times}10^{-2}$ & $1.39{\times}10^{-4}$ \\
6  & $5.64{\times}10^{-2}$ & $8.95{\times}10^{-5}$ & 18 & $5.62{\times}10^{-2}$ & $1.10{\times}10^{-4}$ & 30 & $5.96{\times}10^{-2}$ & $1.22{\times}10^{-4}$ & 42 & $5.74{\times}10^{-2}$ & $1.41{\times}10^{-4}$ \\
7  & $5.65{\times}10^{-2}$ & $9.94{\times}10^{-5}$ & 19 & $5.62{\times}10^{-2}$ & $1.11{\times}10^{-4}$ & 31 & $5.98{\times}10^{-2}$ & $1.23{\times}10^{-4}$ & 43 & $5.71{\times}10^{-2}$ & $1.44{\times}10^{-4}$ \\
8  & $5.64{\times}10^{-2}$ & $1.02{\times}10^{-4}$ & 20 & $5.62{\times}10^{-2}$ & $1.12{\times}10^{-4}$ & 32 & $5.98{\times}10^{-2}$ & $1.24{\times}10^{-4}$ & 44 & $3.65{\times}10^{-2}$ & $1.37{\times}10^{-4}$ \\
9  & $5.64{\times}10^{-2}$ & $1.05{\times}10^{-4}$ & 21 & $5.62{\times}10^{-2}$ & $1.12{\times}10^{-4}$ & 33 & $5.98{\times}10^{-2}$ & $1.25{\times}10^{-4}$ & 45 & $2.96{\times}10^{-2}$ & $1.39{\times}10^{-4}$ \\
10 & $5.64{\times}10^{-2}$ & $1.06{\times}10^{-4}$ & 22 & $5.78{\times}10^{-2}$ & $1.14{\times}10^{-4}$ & 34 & $5.98{\times}10^{-2}$ & $1.26{\times}10^{-4}$ & 46 & $2.98{\times}10^{-2}$ & $1.76{\times}10^{-4}$ \\
11 & $5.64{\times}10^{-2}$ & $1.07{\times}10^{-4}$ & 23 & $5.79{\times}10^{-2}$ & $1.17{\times}10^{-4}$ & 35 & $5.98{\times}10^{-2}$ & $1.27{\times}10^{-4}$ & 47 & $8.93{\times}10^{-4}$ & $1.83{\times}10^{-5}$ \\
\bottomrule
\end{tabular}
\end{adjustbox}
\captionof{table}{
Layer-wise numerical differences between FP16 and FP32 accumulation on \textbf{Qwen 2.5 14B}.
Max and Mean report the maximum and mean absolute differences at each layer.
}
\label{tab:accum_qwen25_14b}
\end{minipage}
\vspace{-2pt}
\makebox[\textwidth][l]{%
\begin{minipage}{0.48\textwidth}
\raggedright
\small
\setlength{\tabcolsep}{6pt}
\renewcommand{\arraystretch}{1.05}
\begin{tabular}{lcccc}
\toprule
\textbf{Config} & \textbf{Top5} & \textbf{Top10} & \textbf{Top20} & \textbf{Top30} \\
\midrule
SpTrans25 & 8.57\%  & 8.63\%  & 8.63\%  & 8.48\% \\
SpTrans50 & 19.19\% & 19.15\% & 18.63\% & 17.76\% \\
SpTrans75 & 21.55\% & 21.66\% & 21.60\% & 21.15\% \\
\bottomrule
\end{tabular}
\captionof{table}{
Weighted INT8 exposure of heavy-hitter importance under SpTrans precision layouts.
Lower values indicate that less high-importance attention mass is routed to INT8.
}
\label{tab:heavy_hitter_int8_exposure}
\end{minipage}
}
\end{table*}

\section{Numerical Analysis}
\label{app:numerical-analysis}

The fixed Torch FP16 implementation provides a common reference for comparing kernel schedules and routed arithmetic paths.
Reported deviations measure implementation-level output agreement with this reference across differences in quantization, accumulation order, rounding behavior, and reduction schedule.

This appendix complements the main numerical analysis in Section~\ref{sec:numerical_analysis}.
We examine numerical behavior from six perspectives:
(i) kernel-level output deviation on random attention inputs,
(ii) model-depth and sequence-length effects,
(iii) INT8 coverage ratio,
(iv) direct comparison with FlashAttention,
(v) accumulation precision and larger-model accumulation checks,
and (vi) static-routing exposure of high-mass attention interactions.

\subsection{Depth and Sequence-Length Effects}

Tables~\ref{tab:num_single_layer_logits}, \ref{tab:num_12layer_logits}, and \ref{tab:num_32layer_logits} report results for a single-layer model, a 12-layer model, and a 32-layer model, respectively.
All experiments use identical random inputs and shared weights across attention implementations.
For each sequence length, the \emph{Torch Value} column reports the maximum absolute logit magnitude produced by the selected Torch FP16 reference, serving as a scale anchor.
All other entries report maximum absolute deviation of model logits relative to this reference.

Across model depths, deviations generally increase as depth increases, reflecting accumulation of implementation-level differences across layers.
Uniform INT8 attention produces larger deviations than mixed-routing configurations.
Reducing the fraction of INT8-routed tile groups generally reduces deviation from the fixed Torch FP16 reference.
Within this controlled setup, model depth and INT8 coverage produce larger changes in logit deviation than sequence length over the evaluated range.

\subsection{Direct Comparison with FlashAttention}

Table~\ref{tab:tilemix_vs_flash_numerical} reports direct output differences between \modelname and FlashAttention under different precision layouts and INT8 ratios.

A nonzero difference exists at 0\% INT8 because \modelname and FlashAttention use different fused-kernel schedules.
Within most structured layouts, MeanDiff increases with INT8 coverage; repeated MaxDiff or MeanDiff entries indicate equality at the reported numerical precision, not bitwise-identical outputs.

\subsection{Pattern and Accumulation Effects}

Table~\ref{tab:pattern_accumulation_numerical} separates the effects of precision layout, INT8 coverage, and accumulation precision.

The results show that numerical behavior depends mainly on the precision layout and INT8 coverage ratio, while FP16 and FP32 accumulation produce similar deviations in this setting.

Table~\ref{tab:fp16_fp32_accum_direct} directly compares FP16 and FP32 accumulation under matched precision layouts.

The small differences across configurations indicate that routing layout and INT8 coverage produce much larger numerical effects than accumulation precision in the evaluated setting.

\subsection{Larger-Model Accumulation Stability}

We further compare \modelname with FP16 accumulation against \modelname with FP32 accumulation on LLaMA3.1-8B and Qwen2.5-14B. Tables~\ref{tab:accum_llama31_8b} and~\ref{tab:accum_qwen25_14b} show that FP16 accumulation remains close to FP32 accumulation across larger models.

On LLaMA 3.1 8B, the maximum layer-wise difference remains on the order of $10^{-2}$, while most mean differences stay around $10^{-3}$ or lower. On Qwen 2.5 14B, maximum differences are larger but remain stable across layers, and mean differences stay around $10^{-4}$. Together, these larger-model checks show that FP16-to-FP32 accumulation differences remain smaller than the routed-precision effects characterized above.

\raggedbottom
\subsection{Positional Routing and Heavy-Hitter Exposure}

We quantify how static SpTrans layouts distribute INT8 execution over high-importance interactions while retaining constant-time routing lookup inside the kernel.

For each layer-head attention map, we define first-order query-key importance as
\[
I_{qk}
=
\left|
P_{qk}
\frac{\partial\mathcal{L}}{\partial P_{qk}}
\right|
\]
For each query $q$, let $\mathcal{H}_q^\tau$ be the smallest key set whose cumulative importance reaches $\tau\%$, where $\tau \in \{5,10,20,30\}$.
Let $D_{qk}=1$ indicate that position $(q,k)$ is routed to INT8.
We define the weighted INT8 exposure of heavy hitters as
\[
E^\tau =
\frac{
\sum_q \sum_{k \in \mathcal{H}_q^\tau} I_{qk} D_{qk}
}{
\sum_q \sum_{k \in \mathcal{H}_q^\tau} I_{qk}
}.
\]
A lower $E^\tau$ indicates stronger protection of high-importance attention mass from INT8 routing.

Table~\ref{tab:heavy_hitter_int8_exposure} shows that SpTrans25 routes only about 8.5\% of the selected high-importance mass to INT8, substantially below its nominal 25\% tile-group coverage.

Across all three coverage levels, structured spatial routing retains a larger share of high-importance interactions in FP16 without online detection.

\end{document}